\documentclass[sigconf, 9pt]{acmart}
\usepackage{amsmath}
\usepackage{array}
\usepackage{booktabs}
\usepackage{multirow}
\usepackage{tabularx}
\usepackage{comment}
\usepackage{algorithm}
\usepackage{algpseudocode}  
\usepackage{enumitem}
\usepackage{soul}
\usepackage{listings}
\usepackage{comment}

\begin{document}

% -----------------------------------------------------------
% TITLE & AUTHORS
% -----------------------------------------------------------
\title{Q-LocalAdam: Memory-Efficient Client-Side Adaptive Optimization for Edge Federated Learning}

% The 'anonymous' class option will automatically suppress this author block in the compiled PDF to comply with the double-blind policy.
\author{Vedant Waykole}
\affiliation{%
  \institution{IISER Bhopal}
  \country{India}
}
\email{waykole22@iiserb.ac.in}

\author{Haroon R Lone}
\affiliation{%
  \institution{IISER Bhopal}
  \country{India}
}
\email{haroon@iiserb.ac.in}

% ACM documents require the abstract to be placed BEFORE \maketitle
\begin{abstract}
On-device federated learning (FL) on edge devices is limited by the memory required by client-side adaptive optimizers. Maintaining optimizer states locally increases memory usage and can restrict the size of models that can be deployed on resource-constrained edge devices. State quantization can reduce this memory footprint, but uniform compression methods may perform poorly under skewed, non-IID data distributions because they can lose small-variance values that are important for stable optimization.

We present \textbf{Q-LocalAdam}, a distribution-aware optimizer for memory-constrained edge devices. It applies block-wise linear quantization to the bounded momentum state and log-space quantization to the heavy-tailed variance state. A post-dequantization variance floor is used to maintain numerical stability. Q-LocalAdam reduces the persistent optimizer-state memory footprint while achieving convergence close to full-precision training.

We evaluate Q-LocalAdam on physical edge devices and GPU, measuring persistent optimizer-state storage, process-level memory, and the latency introduced by quantization across human activity, vision, and language workloads. Our results show reduced peak resident memory with modest latency overhead, indicating that distribution-aware state compression can make adaptive optimization practical for edge deployments.
\end{abstract}

\maketitle
\begin{comment}
    
\section{To Do}
\begin{enumerate}
    \item Finalize abstract - keep sensys in mind - Vedant - then- Haroon
    \item Finalize introduction - Haroon
    \item Try to evaluate on some other Sensor dataset - vedant
    \item Convert to Sensys Template
    \item Read Methods - Haroon
    \item Results 
\end{enumerate}
\hl{haroon- align it to the sensys theme}
\end{comment}

\section{Introduction}
\label{sec:intro}

Cross-device federated learning (FL) enables collaborative model training across a large number of resource-constrained client edge devices (e.g., smartphones, Raspberry Pi, etc.) without centralizing their private data, but heterogeneous non-IID client distributions make optimization difficult and often slow convergence \cite{mcmahan2017fedavg,pfitzner2021federated,imteaj2021survey}. On resource-constrained edge devices, this challenge is compounded by a memory bottleneck: client-side Adam optimizer\cite{kingma2014adam} maintains two additional state tensors first and second moments whose combined storage can equal that of the model parameters when all three are stored in the same precision. Because clients may perform multi-epoch local training, operate in peer-to-peer networks, or participate in multi-task objectives where local optimizer state is needed to track client-specific data distributions \cite{kundroo2023efficient,tao2023preconditioned,qu2022blockchain,pacheco2024securing,mills2021multi}, this optimizer-state footprint directly constrains which models and batch sizes can be trained on device.

Existing efficiency work in federated learning has largely targeted communication overhead (gradient quantization, sparsification, low-rank factorization) or model-size reduction (pruning, distillation, split learning), leaving the client-side optimizer-state footprint comparatively unexplored \cite{alistarh2017qsgd,basu2019qsparse,mills2021communication,li2019fedmd}. Low-bit optimizer-state quantization has been studied in centralized settings, including 4-bit schemes that address a zero-point problem in second-moment quantization \cite{li2023memory,dettmers2021bit}. However, these methods do not address the combination of state-specific distributions, client-side non-IID training, numerical stability under skewed local data, and measured memory behavior on constrained edge hardware the setting this paper targets.

We observe empirically that momentum and variance buffers exhibit distinct
distributions during federated training: momentum is approximately symmetric and
tightly bounded, while variance is strictly positive, highly skewed, and spans
several orders of magnitude. This motivates \textbf{Q-LocalAdam}, a client-side
optimizer that applies block-wise linear 8-bit quantization to momentum and
block-wise log-space 8-bit quantization to variance, while keeping model
parameters in full precision. Log-space coding bounds the reconstruction error
multiplicatively across the encoded range of the variance,and a post-dequantization variance floor $\tau$ ensures that reconstructed values, which could otherwise be driven to zero or below by quantization error, remain strictly positive like the raw variance. Crucially, our sensitivity study (Section~\ref{sec:floor}) shows that this floor is an important design choice rather than an implementation detail: without it, Q-LocalAdam fails to train reliably across seeds, while full-precision Adam also benefits from an appropriately chosen value of $\tau$. Under extreme heterogeneity,
Q-LocalAdam matches full-precision Adam when both use the same floor, while
reducing the optimizer state that must remain resident on the client between
consecutive local training steps within each federated round by $3.37\times$.

We evaluate Q-LocalAdam as a federated-systems method, not only as a quantization scheme. Across ResNet-18 \cite{shafiq2022deep}, ViT-Tiny \cite{sharir2021image}, and a text transformer under several levels of non-IID heterogeneity, the method consistently reduces persistent optimizer-state memory by $3.37\times$ and matches or exceeds full-precision Adam under matched variance floors. We further extend the design to 4-bit states, achieving a $6.40\times$ memory reduction with accuracy within $0.90$~pp of the 8-bit variant across all evaluated heterogeneity levels, and identify linear 4-bit momentum as the main remaining fragility. To validate the systems claim end-to-end, we evaluate Q-LocalAdam on a Raspberry Pi~5 with CPU-only execution and chunked state reconstruction, as well as on Jetson and GPU platform. On the Raspberry Pi~5, persistent optimizer-state storage falls by $3.56\times$, and process peak RSS falls by $19.9\%$ on a 25.2M-parameter model, at a $13.8\%$ increase in measured iteration time.

We make the following contributions in this work.

\begin{itemize}
    \item \textbf{Federated-systems characterization:} We identify the client-side optimizer-state footprint as a distinct memory bottleneck in cross-device FL, and show that naive uniform quantization of Adam's states fails under non-IID data because it destroys precision in heavy-tailed variance values. We further show that a variance floor is required for stability, and that full-precision Adam benefits from the same floor.
    \item \textbf{Method:} We introduce Q-LocalAdam, a client-side optimizer combining block-wise linear 8-bit momentum quantization, log-space 8-bit variance quantization, and a post-dequantization variance floor, while preserving full-precision model parameters. We extend the design to 4-bit states with a $6.40\times$ memory reduction.
    \item \textbf{Edge deployment and evaluation:} We evaluate Q-LocalAdam across ResNet, ViT, and text architectures under non-IID heterogeneity. We validate it on a Raspberry Pi~5 with CPU-only execution, Jetson and GPU platforms, measuring persistent-state storage, process-level peak RSS, and the latency cost of quantization..
\end{itemize}

%%==============================================================================
%% 2. RELATED WORK
%%==============================================================================
\section{Related Work}

Federated learning often relies on adaptive optimizers to counter non-IID data degradation~\cite{mcmahan2017fedavg,reddi2021adaptive,karimireddy2020scaffold}. Recent client-side adaptive approaches, such as FedCAda~\cite{tang2024fedcada}, maintain optimizer states locally to better track client-specific distributions. However, storing momentum and variance in full precision roughly triples the training memory footprint, severely limiting deployment on edge devices. Our work directly addresses this bottleneck through distribution-aware state quantization.

Low-precision training, such as 8-bit Adam~\cite{dettmers2021bit}, is effective in centralized or IID settings where gradient statistics are relatively stable. However, in cross-device federated learning, local updates are computed on highly heterogeneous data~\cite{li2020federated}. Under these conditions, naive uniform quantization of momentum and variance destroys essential numerical precision and severely degrades training. This failure motivates our asymmetric design: linear quantization for bounded momentum and log-space quantization for the heavy-tailed variance.

Broader efficiency efforts in federated learning have largely focused on orthogonal bottlenecks. Prior works extensively explore communication compression via gradient quantization, sparsification, and low-rank factorization~\cite{alistarh2017qsgd,basu2019qsparse,hyeon2021fedpara,jin2020stochastic,zhang2025fedhq,reisizadeh2020fedpaq,oh2022communication}. Other methods reduce the storage or computational burden of the models themselves through pruning, knowledge distillation, and split learning~\cite{mills2021communication,li2019fedmd,tamirisa2024fedselect,thapa2022splitfed,vepakomma2018split,mu2025federated,zhang2024fedsl}. Because these approaches primarily target communication overhead or model weight compression rather than the memory footprint of client-side optimizer states, Q-LocalAdam serves as a natural and complementary extension to existing federated optimization pipelines.
~\cite{li2023memory} propose memory-efficient optimizers that
quantize Adam's first and second moments to 4 bits, using a quantile-based
encoding and a zero-point-free linear scheme for the second moment to avoid
the underflow that plagues naive low-bit quantization. Their method targets
centralized training, where gradient statistics are relatively stable, and
does not address the combination of state-specific distributions, client-side
non-IID heterogeneity, and measured memory behaviour on constrained edge
hardware that motivates our work. We use their recipe as a baseline in our
4-bit extension (Section~8.1) and find that, while their zero-free linear
variance quantizer is a necessary safeguard, it is not sufficient under
extreme non-IID skew: it reaches $51.59\%$ on CIFAR-100 ($\alpha=0.1$) \cite{Krizhevsky2009LearningML},
whereas our log-space variance coding reaches $60.27\%$ at a similar memory
footprint.

\section{Method}

\subsection{The Statistical Geometry of Optimizer States}
\label{sec:client_adaptive}

We consider a standard cross-device federated learning objective where $K$ clients collaboratively minimize a global function $f(\theta) = \sum_{k=1}^{K} p_k f_k(\theta)$. To effectively navigate non-IID local data landscapes, clients execute multiple epochs of Adam optimization locally before aggregating weights at the central server. While this stabilizes convergence, maintaining full-precision (FP32) momentum ($m$) and variance ($v$) buffers requires $8N$ bytes for an $N$-parameter model, tripling the client's memory footprint (Appendix \ref{app:fedadam}).

To resolve this memory bottleneck, we first analyze the statistical behavior of these buffers during federated training. Empirical profiling (Appendix \ref{app:optimizer_dist}) reveals that the momentum and variance tensors exhibit fundamentally divergent native geometries: momentum values are highly symmetric and tightly bounded (typically occupying $[-0.1, 0.1]$), closely resembling a zero-mean Gaussian distribution while variance values are strictly positive, span up to eight orders of magnitude (from $10^{-12}$ to $10^{-3}$), and follow a heavy-tailed, log-normal distribution (Appendix \ref{app:optimizer_dist}).

This asymmetry explains why uniform linear 8-bit quantization catastrophically fails in non-IID federated settings. Linearly compressing the variance buffer systematically destroys the precision of small magnitude values. Because the Adam update step divides by $\sqrt{v}$, these tiny variance values are mathematically critical for scaling up gradient steps on rare features. Erasing them causes the optimizer to diverge. Consequently, state compression must be distribution-aware.

\subsection{Distribution-Aware Block-wise Quantization}

Motivated by this observation, Q-LocalAdam compresses the optimizer buffers asymmetrically. Both tensors are partitioned into contiguous blocks of size $B$ (default $B=64$), and each block is quantized independently. Crucially, the model parameters $\theta$ remain in full precision to preserve gradients.

\paragraph{Block-wise quantization.}
Each optimizer-state tensor is flattened and split into contiguous blocks of
$B{=}64$ values. We write $m^{(b)}$ and $v^{(b)}$ for the vectors of momentum and
variance values in block $b$. All operations below are applied separately to each
block, and $\min(\cdot)$ and $\max(\cdot)$ are taken over the values inside that
block. Because each block uses its own scale, a few large values in one block do
not reduce the precision of the other blocks. The symbol $\lfloor\cdot\rceil$
means rounding to the nearest integer. Every value is stored as an 8-bit integer
$q \in \{0,\dots,255\}$, together with two FP32 numbers per block (the block
minimum and range) that are needed to decode it.

\paragraph{Linear quantization for momentum.}
Momentum values are roughly centred around zero and lie in a narrow range, so
evenly spaced levels represent them well. For block $b$, let
$r^{(b)} = \max(m^{(b)}) - \min(m^{(b)})$ be the spread of the block. We map each value to an integer code $q^{(b)} \in \{0,\dots,255\}$, giving 256
quantization levels and 255 evenly spaced intervals between the block minimum
and maximum:
\begin{equation}
q^{(b)} = \left\lfloor \frac{m^{(b)} - \min(m^{(b)})}{r^{(b)}} \cdot 255 \right\rceil .
\label{eq:lin_quantize}
\end{equation}
When the momentum is needed for an update, it is recovered as
\begin{equation}
\tilde{m}^{(b)} = \frac{q^{(b)}}{255} \cdot r^{(b)} + \min(m^{(b)}),
\label{eq:lin_dequantize}
\end{equation}
where $\tilde{m}^{(b)}$ is the reconstructed (approximate) momentum. The $256$
codes define $255$ evenly spaced intervals, so the level spacing is
$r^{(b)}/255$. Since each value is rounded to the nearest level, the
reconstruction error is at most $r^{(b)}/510$.
spacing between levels.

\paragraph{Log-space quantization for variance.}
Variance values are positive and span many orders of magnitude. Evenly spaced
levels (Eq.~\ref{eq:lin_quantize}) would spend almost all $256$ levels on the
largest values and round the many tiny values to zero (Appendix~\ref{app:quant_analysis}).
We therefore quantize the \emph{logarithm} of the variance instead. For block $b$,
we first compute
$\ell^{(b)} = \log\!\big(v^{(b)} + \epsilon_{\log}\big)$,
where $\epsilon_{\log}$ is a small constant that avoids $\log(0)$. Let
$r_{\ell}^{(b)} = \max(\ell^{(b)}) - \min(\ell^{(b)})$ be the spread of the
log-values in the block. We then apply the same linear mapping, now in log space:
\begin{equation}
q_{\ell}^{(b)} = \left\lfloor \frac{\ell^{(b)} - \min(\ell^{(b)})}{r_{\ell}^{(b)}} \cdot 255 \right\rceil .
\label{eq:log_quantize}
\end{equation}
The variance is reconstructed by undoing the steps in reverse order, decoding the
log-value and then exponentiating:
\begin{equation}
\tilde{v}^{(b)} = \exp\!\left(\frac{q_{\ell}^{(b)}}{255} \cdot r_{\ell}^{(b)} + \min(\ell^{(b)})\right) - \epsilon_{\log}.
\label{eq:log_dequantize}
\end{equation}
Equal steps in log space correspond to equal \emph{ratios} in the original space,
so a small variance such as $10^{-12}$ and a large one such as $10^{-3}$ are each
stored with about the same relative precision. This is why small variance values
survive quantization.

\paragraph{Memory Layout.} Each compressed block stores $B$ INT8 values (1 byte each) and two FP32 metadata scalars (the minimum and the range, totaling 8 bytes). For $B = 64$, a block requires 72 bytes, compared to 256 bytes for a standard FP32 block. Accounting for tensor padding and memory alignment, this yields a consistent empirical client-side optimizer memory reduction of $3.37\times$.

\subsection{The Q-LocalAdam Algorithm}
\label{sec:algorithm}

Q-LocalAdam reduces the persistent memory of client-side Adam by storing the first and second moments as 8-bit blocks between optimizer steps. The key design choice is to separate the \emph{storage} representation from the \emph{working} precision: moments are kept quantized, but every update is computed in FP32 on the dequantized values. Quantization is performed entirely on the client, so the server-side aggregation is unchanged. Algorithm~\ref{alg:qlocaladam} gives the procedure.

 The key design choice is to separate the representation used for storage from the precision used for optimization: optimizer states are stored as quantized blocks between local training steps, but reconstructed in FP32 whenever they are needed for an update. Algorithm~\ref{alg:qlocaladam} summarizes the complete procedure.

\begin{algorithm}[t]
\caption{Q-LocalAdam: client-side training in one round}
\label{alg:qlocaladam}
\small
\begin{algorithmic}[1]
\Require Global model $\theta_t$, learning rate $\eta$, decay rates $\beta_1,\beta_2$, local epochs $E$, block size $B$, variance floor $\tau$, constant $\epsilon$
\State $\theta \leftarrow \theta_t$; \quad $s \leftarrow 0$
\State For every weight tensor $i$: $\mathcal{S}_i \leftarrow \varnothing$ \Comment{$\mathcal{S}_i$: stored 8-bit state (momentum and variance blocks)} \Comment{no stored state}
\For{each local epoch $1,\dots,E$}
  \For{each minibatch $\mathcal{B} \subseteq \mathcal{D}_k$}
    \State $g \leftarrow \nabla_\theta L(\theta;\mathcal{B})$ \Comment{FP32 gradient of all tensors}
    \State $s \leftarrow s+1$
    \For{each weight tensor $i$}
      \If{$\mathcal{S}_i = \varnothing$}
        \State $m \leftarrow 0,\; v \leftarrow 0$ \Comment{first update of the round}
      \Else
        \State $m \leftarrow \textsc{DequantLinear}(\mathcal{S}_i^{m})$ \Comment{Eq.~\ref{eq:lin_dequantize}}
        \State $v \leftarrow \max\!\big(\textsc{DequantLog}(\mathcal{S}_i^{v}),\,\tau\big)$ \Comment{Eq.~\ref{eq:log_dequantize}, variance floor}
      \EndIf
      \State $m \leftarrow \beta_1 m + (1-\beta_1)\, g_i$ \Comment{update momentum}
      \State $v \leftarrow \beta_2 v + (1-\beta_2)\, g_i^{2}$ \Comment{update variance}
      \State $\mathcal{S}_i^{m} \leftarrow \textsc{QuantLinear}(m)$ \Comment{Eq.~\ref{eq:lin_quantize}}
      \State $\mathcal{S}_i^{v} \leftarrow \textsc{QuantLog}(v)$ \Comment{Eq.~\ref{eq:log_quantize}}
      \State $\hat m \leftarrow m/(1-\beta_1^{s})$; \quad $\hat v \leftarrow v/(1-\beta_2^{s})$ \Comment{bias correction using $s$}
      \State $\theta_i \leftarrow \theta_i - \eta\, \hat m/(\sqrt{\hat v}+\epsilon)$ \Comment{weight update}
    \EndFor
  \EndFor
\EndFor
\State \Return $\theta$ \Comment{only weights are sent to the server}
\end{algorithmic}
\end{algorithm}

\paragraph{Initialization.}
At the start of each round the client copies the global model $\theta_t$, creates a fresh optimizer, and sets the step counter $s$ to zero. The step counter $s$ tracks how many updates have been performed in the
current round. It is used for bias correction: because the momentum and
variance estimates start at zero, their early values are biased toward zero.
Dividing by $(1-\beta_1^s)$ and $(1-\beta_2^s)$, as in the weight-update
step, corrects this bias, so the effective learning rate is stable from the
first update onward. No optimizer state is stored yet, so the first update of every round starts from $m=0$ and $v=0$ and nothing is dequantized. Quantization begins after this first update, when the new moments are written back as 8-bit blocks.

\paragraph{One local update.}
For each minibatch the client computes the FP32 gradient of the whole model, increments $s$, and then updates the weight tensors one at a time. For each weight tensor $i$, the optimizer performs four operations: it loads
the previous moments (dequantizing them and applying the variance floor),
updates the moments in FP32 from the current gradient, requantizes and stores
the updated moments, and then applies the bias-corrected weight update. These
four steps are marked with comments in Algorithm~\ref{alg:qlocaladam}.
\begin{enumerate}
  \item \textbf{Load the previous moments.} The stored momentum and variance blocks are dequantized to FP32. The reconstructed variance is clamped from below by $\tau$ so that it is strictly positive before the Adam update.
  \item \textbf{Update the moments.} The exponential moving averages are computed in FP32 from the current gradient $g_i$.
  \item \textbf{Quantize and store.} The updated momentum is stored with block-wise linear 8-bit quantization and the updated variance with block-wise log-space 8-bit quantization, replacing the previous blocks.
  \item \textbf{Update the weights.} Bias correction uses the step counter $s$, and the weights are updated with the FP32 moments computed in step~2.
\end{enumerate}
The FP32 moments of the current step determine the current weight update; only their quantized copies are stored. Quantization error therefore enters later updates through state reconstruction, not the current update directly.

\paragraph{Variance floor.}
Quantization error can make a reconstructed variance entry zero or slightly negative. The floor $\tau$ (default $10^{-8}$) is applied when the stored variance is read back, which removes this failure mode; Section~\ref{sec:floor} studies its effect on both Q-LocalAdam and full-precision Adam.

\paragraph{Communication.}
After $E$ local epochs the client returns only its weights $\theta$. Optimizer states are never transmitted and are discarded at the end of the round, so the server aggregates client models with the standard data-size-weighted FedAvg rule.

\paragraph{Execution on constrained devices.}
The algorithm above updates one weight tensor at a time. On the Raspberry Pi, we additionally split each tensor into chunks aligned with the quantization blocks, bounding the temporary FP32 workspace to a single chunk (Section~\ref{sec:rpi}).
%%==============================================================================
%% 4. EXPERIMENTAL SETUP
%%==============================================================================
\section{Experimental Setup}   
\label{sec:experiments}

% ---------------------------------------------------------------------------
% Section 4 (rewritten): Experimental Setup
% Requires: booktabs, xcolor (for \todo), graphicx
% \newcommand{\todo}[1]{\textcolor{red}{[TODO: #1]}}   % put in preamble
% All \todo{...} items need author input and must be removed before submission.
% ---------------------------------------------------------------------------

%Our evaluation follows the way Q-LocalAdam would be developed and deployed.
We first deploy Q-LocalAdam on a Raspberry~Pi~5 to measure persistent
optimizer-state storage, process-level memory, and the latency cost of
quantization. We then validate it on an NVIDIA Jetson Orin Nano. Next, we
study convergence under controlled statistical heterogeneity in a GPU-based
federated simulation. Finally, we run a naturally partitioned
human-activity-recognition workload on a Pi~5 under an
operating-system-enforced memory limit that emulates a 512\,MiB-class client.
Together these isolate the two failure modes common in edge FL: convergence
collapse under severe data heterogeneity, and optimizer-memory exhaustion
under realistic hardware constraints..

%---------------PLATFORMS-----------------

\begin{table*}[]
\centering
\small
\setlength{\tabcolsep}{6pt}
\begin{tabular}{lll}
\toprule
\textbf{Regime} & \textbf{Platform/Hardware} & \textbf{Purpose} \\
\midrule
Controlled FL simulation & NVIDIA GeForce RTX~5080, 16\,GB GDDR7 & Accuracy under Dirichlet non-IID skew; ablations \\
Accelerator edge board & Jetson Orin Nano, 8\,GB LPDDR5 (unified memory) & In-situ GPU edge validation \\
CPU-only edge board & Raspberry Pi~5 Model~B, 4\,GB, aarch64 & Persistent state, peak RSS, latency, chunked reconstruction \\
Emulated 512\,MiB-class client & Pi~5 with a 472\,MiB enforced memory cap & Whether a client trains or is killed \\
\bottomrule
\end{tabular}
\caption{Execution regimes used in the evaluation.}
\label{tab:platforms}
\end{table*}

\subsection{Platforms and Execution Regimes}
\label{sec:platforms}

We evaluate Q-LocalAdam on three classes of platforms, each chosen to answer a
different question. A server-grade GPU is used for controlled federated-learning
simulations, where we need to isolate the effect of data heterogeneity on
convergence and run ablations across many configurations. Two edge boards, an
NVIDIA Jetson Orin Nano and a Raspberry Pi~5, are then used to validate the
method under realistic hardware constraints. The Jetson provides an
accelerator-based edge setting with a shared CPU--GPU memory pool, while the
Raspberry Pi provides a CPU-only setting in which we can measure persistent
optimizer-state storage, process-level memory, latency, and behavior under an
enforced memory cap.

\paragraph{GPU platform.}
Controlled federated simulations and ablations run on a single NVIDIA GeForce
RTX~5080 with 16\,GB of GDDR7 memory, driver version 595.99.02 and CUDA
version 13.2. This platform provides sufficient compute and memory to run many
federated configurations (multiple heterogeneity levels, model architectures,
and optimizer variants) under identical settings, making it suitable for
isolating the effect of non-IID skew on optimizer-state quantization rather
than measuring deployment behavior.

\paragraph{Edge platforms.}
The edge evaluation uses two boards with complementary constraints. The Jetson
Orin Nano provides a GPU-accelerated edge setting with 8\,GB of LPDDR5 unified
memory shared between CPU and GPU, allowing us to validate Q-LocalAdam in situ
on accelerator hardware. The Raspberry Pi~5 \cite{} provides a CPU-only setting with
4\,GB of memory, where optimizer-state compression has a direct effect on
process-level memory and where we can additionally enforce a hard per-client
memory cap to test feasibility under constrained budgets.
All Pi experiments run on a Raspberry Pi~5 Model~B (4\,GB) with an
aarch64, CPU-only PyTorch build, using four intra-op threads and one
inter-op thread.

\noindent{\bf Execution regimes:} Table~\ref{tab:platforms} defines different execution regimes depending upon the platform used for the specific experiment. All regimes
simulate federated clients sequentially on a single machine; no wireless
communication is measured.

%\paragraph{Raspberry Pi~5 platform.}

% The entire local training pipeline (forward and backward
% passes, state reconstruction, moment updates, requantization, and parameter
% updates) runs on the CPU to keep measurements independent, every (model, optimizer, seed) run is launched as a separate process, and runs are executed one after another on the board. Because the Pi's speed can change as it heats up, we alternate which optimizer runs first from seed to seed, so neither consistently benefits from a cooler device. A background monitor
% samples CPU temperature, frequency, process memory, available system memory,
% and throttling flags every two seconds. Before each run the driver attempts
% to cool the device to $55^\circ$C within a bounded wait, so initial
% temperatures are not guaranteed to be identical across runs.

\subsection{Datasets and Federated Partitioning}
\label{sec:datasets}

We evaluate Q-LocalAdam on four datasets that probe complementary aspects of
federated learning on constrained devices. CIFAR-10 and CIFAR-100 \cite{Krizhevsky2009LearningML} provide
controlled image-classification workloads in which we vary the degree of
non-IID client skew using Dirichlet partitions, allowing us to isolate the
effect of heterogeneity on optimizer-state quantization. AG News \cite{zhang2015character} extends this
evaluation to federated text classification. HHAR ~\cite{heterogeneity_activity_recognition_344} provides a deployment-style
sensor workload in which clients correspond to physical devices rather than
synthetic partitions, and is used to test whether training completes under an
enforced memory limit. The Raspberry Pi experiments additionally use two
CIFAR-100 configurations: a small fixed subset for measuring storage, memory,
and latency, and the full dataset with Dirichlet partitioning for multi-round
federated simulation.

\paragraph{HHAR: naturally partitioned sensor data.} The Heterogeneity Human Activity Recognition (HHAR)
dataset~\cite{heterogeneity_activity_recognition_344} is a standard benchmark
for mobile and edge sensor analytics. It contains accelerometer and gyroscope
time series recorded while users perform physical activities carrying
different smartphones and smartwatches. We use the three-axis accelerometer
signal, cut into 128-sample windows, and retain raw sensor noise and
per-device sampling-rate variation, so window duration varies with the device
sampling rate. The six classes are bike, sit, stand, walk, stairs-up, and
stairs-down.

The HHAR dataset contains recordings from 9 participants, each carrying two devices: a
smartphone and a smartwatch. Rather than treating each participant as a single client, we
treat each individual sensor device as a separate client. This yields 18 clients in total:
one phone client and one watch client for each of the 9 participants. For example,
participant ``a'' contributes two clients, \texttt{a\_phone} and \texttt{a\_watch}, each
with its own accelerometer stream, sampling rate, and sensor noise characteristics. This
partition reflects a realistic edge deployment, in which each device trains locally on the
data it senses, and it preserves both user-level behavioral variation and device-level
hardware variation. No synthetic Dirichlet partitioning is applied. This natural partition
preserves two sources of edge heterogeneity: behavioral variation across users and hardware
variation between phones and watches, including different sensor noise and sampling rates.

For each user--device--activity combination, we cut the accelerometer
readings into non-overlapping windows. Neighbouring windows
from the same activity are highly similar, so we group every $16$ consecutive
windows into a block and assign whole blocks to the training set ($80\%$) and test set ($20\%$) to reduce leakage between the two sets. Each device is
treated as one federated client, giving $18$ clients in total. To keep
per-client compute and memory small enough for our memory-constrained setup,
we retain at most $1{,}500$ training and $300$ test windows per client. All
windows are standardized using the mean and standard deviation of the pooled
training windows. The test set pools the test windows of all clients, so it
measures accuracy on the same users and devices seen during training. This
workload is used only to determine whether training completes under a fixed
memory limit, not as an HHAR accuracy benchmark.

\paragraph{CIFAR-10/100 with Dirichlet partitions.}
We evaluate on CIFAR-10 and CIFAR-100 (50{,}000 training / 10{,}000 test
images, $32{\times}32$), standard benchmarks for non-IID federated
learning~\cite{mcmahan2017fedavg,reddi2021adaptive}. These datasets provide
controlled image-classification workloads with well-understood architectures
and heterogeneity settings, allowing us to isolate the effect of non-IID client
skew on optimizer-state quantization before evaluating deployment behavior on
edge hardware. Training data is partitioned across clients via a Dirichlet
distribution with concentration $\alpha \in \{0.1, 0.5, 1.0, \mathrm{IID}\}$;
lower $\alpha$ induces stronger skew. At $\alpha{=}0.1$, the dominant class
comprises $65.7\%$ of a client's CIFAR-10 data on average (some clients
exceed $96\%$), creating the minority-class variance-underflow conditions
that motivate our design. Tables and details with 
heterogeneity levels and per-client distributions are in Appendix \ref{app:data_distribution}.

% The Raspberry Pi evaluation uses the full CIFAR-100 training dataset, comprising 50,000 images across all 100 classes, with the same ResNet-18 architecture and 100-output classification head as in the federated experiments. The training data are partitioned across five clients using a Dirichlet distribution with $\alpha=0.1$.

To bound the computational workload on the resource-constrained Raspberry Pi board, local training is limited to a fixed maximum number of mini-batches per client per round. The training data are partitioned across five clients using a Dirichlet distribution with $\alpha=0.1$. The batch size, local epochs, and number of training rounds are controlled across experiments. We measure optimizer-state storage, peak process memory, forward/backward latency, and optimizer-step latency. The complete training and test datasets are loaded before resetting the process peak-memory counter, allowing the reported peak-memory increase to characterize additional memory consumption during model training rather than initial dataset loading. Classification accuracy is also evaluated using the test dataset.

\paragraph{AG News.}
For a non-vision workload we use AG News~\cite{zhang2015character}, tokenized
with \texttt{bert-base-uncased} and truncated to 64 tokens, under IID and
$\alpha{=}0.1$ partitions (Section~\ref{sec:nlp}).

%------------MODELS-----------------

\subsection{Models}
\label{sec:models}

Table~\ref{tab:models} lists every model used in the paper and their parameters. We deliberately use models at different scales and architectures to answer complementary questions: whether
Q-LocalAdam preserves accuracy across standard federated workloads, whether its
benefits extend beyond convolutional networks, and whether reducing persistent
optimizer state produces measurable system-level benefits on edge hardware.
Model sizes are selected so that Adam's two FP32 moment buffers, which require
$8N$ bytes for an $N$-parameter model, occupy a substantial fraction of client
memory on the target device.

\begin{table*}[t]
\centering
\small
\setlength{\tabcolsep}{6pt}
\begin{tabular}{lllrl}
\toprule
\textbf{Workload} & {\bf Platform} & \textbf{Model} & \textbf{Params.} & \textbf{Input} \\
\midrule
CIFAR-10 & GPU simulation & ResNet-18 (width-adapted) & $\sim$6.3M & $224{\times}224$ \\
CIFAR-100 & GPU simulation & ResNet-18 (width-adapted) & $\sim$11M & $224{\times}224$ \\
CIFAR-10 & GPU simulation & ViT-Tiny (patch 16, $d{=}192$, depth 12, 3 heads) & 5.49M & $64{\times}64$ \\
AG News & GPU simulation & TextTransformer (2 encoder layers, CLS head) &   &64 tokens \\
CIFAR-100 & Jetson Orin Nano & Slim ResNet-18 & 2.8M & $32{\times}32$ \\
\midrule
CIFAR-100 & Pi~5 benchmark & ResNet-18 (width 64) & 11.2M & $32{\times}32$ \\
CIFAR-100 & Pi~5 benchmark & Transformer ($d{=}384$, 8 layers, 6 heads, patch 4) & 14.3M & $32{\times}32$ \\
CIFAR-100 & Pi~5 benchmark & ResNet-34 (width 64) & 21.3M & $32{\times}32$ \\
CIFAR-100 & Pi~5 benchmark & ResNet-18 (width 96) & 25.2M & $32{\times}32$ \\
CIFAR-100 & Pi~5 fed. simulation & ResNet-18 (width 16) & 0.713M & $32{\times}32$ \\
HHAR & Pi~5, 472\,MiB cap & KWT-style transformer encoder~\cite{berg2021keyword} (patch 8, $d{=}512$, 4 layers, 8 heads, FFN 2048) & 12.64M & 128-sample, 3-axis \\
\bottomrule
\end{tabular}
\caption{Models and workloads. Convolutional Pi models use a $3{\times}3$ stem
and native $32{\times}32$ inputs. The model set spans convolutional, vision
transformer, text transformer, and sensing-transformer architectures to test
whether Q-LocalAdam's benefits depend on a particular architecture or workload.}
\label{tab:models}
\end{table*}

\paragraph{Jetson Orin Nano model.}
The Jetson Orin Nano experiments use a compact CIFAR-style ResNet-18 with base width 32. The network has a \(3 \times 3\), stride-1 convolutional stem without max-pooling, followed by four stages of two BasicBlocks each. The stages have 32, 64, 128, and 256 channels, with stride-2 downsampling at the start of stages 2--4. It ends in global average pooling and a 100-way linear classifier. The model has approximately 2.8M parameters, so Adam's two FP32 moment buffers require \(8N \approx 22.6\)~MB. It takes native \(32 \times 32\) CIFAR-100 inputs, with no upsampling, which keeps activation memory low. We chose this configuration to fit the 8~GB of LPDDR5 memory that the CPU and GPU share on the Jetson, alongside the operating system and telemetry tools. The Jetson workload is otherwise the same as the main GPU simulation: it uses five clients with full participation, 120 rounds of FedAvg, and two local epochs. Data are partitioned with a Dirichlet distribution (\(\alpha = 0.1\)), and clients use batch size 8, learning rate \(10^{-3}\), and seed 42.

\paragraph{GPU-simulation models.}
We use width-adapted ResNet-18 models on CIFAR-10 and CIFAR-100 as the primary
controlled federated-learning workloads. ResNet-18 is a standard, widely
reproduced backbone, and varying its width lets us keep the architecture fixed
while changing the optimizer-state footprint. Inputs are upsampled from
$32{\times}32$ to $224{\times}224$ to create a memory-intensive setting
representative of backbone-based vision pipelines, where optimizer buffers
compete with large activation maps. ViT-Tiny~\cite{dosovitskiy2020image} is
included to test whether the benefits of distribution-aware state quantization
extend beyond convolutional models to self-attention architectures. The text transformer on AG News tests modality generality beyond image
classification.

\paragraph{Pi benchmark models.}
The four Raspberry Pi benchmark models span 11--25M parameters and are chosen
to make optimizer-state memory a measurable share of process memory on a 4\,GB
board: under FP32, Adam's moments occupy roughly 90--200\,MB. ResNet-18 and
ResNet-34 provide standard convolutional backbones; the wider ResNet-18
increases model size while holding the architecture family fixed, isolating the
effect of parameter count; and the transformer tests whether the memory and
latency behavior extends beyond convolutional models. Together, these models
allow us to measure how persistent-state compression, process RSS, and
quantization latency scale with model size and architecture.

\paragraph{Pi federated-simulation model.}
The multi-round Pi federated simulation uses a much smaller width-16 ResNet-18
(0.713M parameters). This model is intentionally small so that five clients and many federated rounds can complete within a practical time budget
on one board. Its purpose is therefore to validate multi-round operation and
held-out accuracy under non-IID partitioning, not to demonstrate peak memory
savings; the larger Pi benchmark models serve that purpose.

\paragraph{HHAR model.}
The HHAR experiment uses a 12.64M-parameter KWT-style transformer
encoder~\cite{berg2021keyword} operating on 1-D patches of three-axis
accelerometer signals. We select a transformer-based sensing model because it
is representative of modern on-device sequence models and because its size
makes FP32 Adam state a dominant component of client memory: its FP32 weights
($\approx$50.6\,MB), gradients ($\approx$50.6\,MB), and Adam state
($8N\approx101$\,MB) already approach the budget of a 512\,MiB-class device
before activations and runtime overhead are counted. This makes HHAR the
appropriate workload for the enforced 472\,MiB memory-limit experiment, where
the question is whether a client can complete federated training at all.

\subsection{Training Protocol}
\label{sec:protocol}

\paragraph{GPU simulation.}
Training runs for 120 federated rounds (FedAvg; 100 for ViT-Tiny and AG News),
with 5 clients per round, 2 local epochs, and batch size 64, using
$\beta_1{=}0.9$, $\beta_2{=}0.999$, Adam $\epsilon{=}10^{-8}$, and
$\eta{=}10^{-3}$ ($3{\times}10^{-4}$ for ViT-Tiny, $5{\times}10^{-4}$ for AG
News). Optimizer states $(m,v)$ are reinitialized at the start of each round,
following standard cross-device FL
convention~\cite{mcmahan2017fedavg,reddi2021adaptive}. Primary experiments
use seed 42; multi-seed validation (seeds 42, 123, 456) is in
Table~\ref{tab:multiseed}.

\paragraph{Pi~5 local-training benchmark.}
Each configuration performs 100 updates per client per round (2 local epochs, capped at 50 minibatches each) with batch size 8,
$\eta{=}10^{-3}$, $\beta_1{=}0.9$, $\beta_2{=}0.999$, sampling minibatches with
replacement, over three seeds.

\paragraph{Pi~5 federated simulation.}
Five clients run sequentially on one board, all participating in every
round, on CIFAR-100 with $\alpha{=}0.1$. In each round every client starts from
the global model, resets its optimizer state, and performs two local passes,
each capped at 50 minibatches of 8 examples (100 updates per client per round).
The server averages models weighted by partition size, and accuracy uses the
full 10{,}000-image test set. We run 40 rounds and 120 rounds (seeds 1, 2, 3).

\paragraph{HHAR under a memory cap.}
120 FedAvg rounds, 4 of 18 clients sampled per round, one local epoch, batch
size 32, $\eta{=}3{\times}10^{-4}$, $\beta{=}(0.9,0.999)$,
$\epsilon{=}10^{-8}$, four CPU threads, seed 42. Optimizer state
is reset every round. Each client trains in its own freshly created process
under a kernel-enforced 472\,MiB limit (Section~\ref{sec:hardcap}).

\subsection{Quantization Details}
\label{sec:quantdetails}

Q-LocalAdam uses block size $B{=}64$: a tensor with $N$ elements is split into
$\lceil N/B\rceil$ blocks, each storing $B$ INT8 values and two FP32 scalars
(the block minimum and range, or their log-space equivalents), totaling
$(B{+}8)$ bytes per block. For $B{=}64$ this is 72 bytes versus 256 bytes in
FP32, a theoretical $3.56{\times}$ compression. Tensors not divisible by $B$
are zero-padded and padding indices are ignored during dequantization. All
reported memory figures include metadata and padding overhead. The Pi
implementation realizes the full $3.56{\times}$, whereas the GPU reference
implementation measures $3.37{\times}$ ; 3.37 corresponds to $\approx$76 bytes per block, i.e.\ $\approx$12 B
of metadata rather than 8 B. The 4-bit variant packs two codes per byte,
giving 40 bytes per block and $6.40{\times}$ (Section~\ref{sec:fourbit}). A
stability constant $\epsilon_{\log}{=}10^{-8}$ in the log-space quantizer
prevents $\log(0)$, and a post-dequantization variance floor $\tau{=}10^{-8}$
keeps reconstructed variance strictly positive (Section~\ref{sec:client_adaptive}).

\begin{table*}[!t]
\centering

\setlength{\tabcolsep}{4pt}
\begin{tabular}{lrrrrrrr}
\toprule
 & & \multicolumn{2}{c}{State (MB)} & \multicolumn{2}{c}{Optimizer step (ms)} & \multicolumn{2}{c}{Iteration (ms)} \\
\cmidrule(lr){3-4}\cmidrule(lr){5-6}\cmidrule(lr){7-8}
Model & Params & FP32 & Q-LocalAdam & FP32 & Q-LocalAdam & FP32 & Q-LocalAdam \\
\midrule
ResNet-18 (w64) & 11.2M & 89.761 & 25.245 & $174.7{\pm}1.3$ & $472.2{\pm}8.6$ & $596.0{\pm}6.2$ & $886.1{\pm}12.3$ \\
Transformer (d384) & 14.3M & 114.233 & 32.128 & $219.7{\pm}1.8$ & $588.3{\pm}10.0$ & $624.0{\pm}11.0$ & $992.6{\pm}12.3$ \\
ResNet-34 (w64) & 21.3M & 170.626 & 47.989 & $330.1{\pm}4.7$ & $907.0{\pm}10.9$ & $1134.2{\pm}26.1$ & $1702.4{\pm}18.8$ \\
ResNet-18 (w96) & 25.2M & 201.586 & 56.697 & $391.0{\pm}1.9$ & $1078.9{\pm}10.9$ & $1245.1{\pm}14.8$ & $1934.5{\pm}19.5$ \\
\bottomrule
\end{tabular}
\caption{Raspberry Pi~5 local-training benchmark (mean $\pm$ std over 3 seeds; Q-LocalAdam uses a maximum chunk size of $2^{20}$ elements). Persistent state shrinks by $\approx\!3.56\times$; the optimizer step is $2.68$--$2.76\times$ slower.}
\label{tab:rpi_main}
\end{table*}

\subsection{Metrics}
\label{sec:metrics}

\paragraph{Accuracy.}
We report peak test accuracy (maximum over rounds), final accuracy (last round),
and, where rounds are noisy, the mean over the last 10 rounds.

\paragraph{Persistent optimizer-state storage.}
Computed from the live optimizer-state tensors, including values, metadata, and
padding. GPU experiments measure it on client~0 after round~1. Pi experiments
report decimal MB ($10^6$ bytes), excluding Python object overhead, allocator
reservations, reconstructed FP32 intermediates, and other training allocations.

\paragraph{Process-level memory.}
We report the peak resident set size (RSS), the
maximum amount of physical memory the process used at any point during
the run, read from Linux /proc (VmHWM) and reported in MiB ($2^{20}$
bytes). It includes parameters,
gradients, optimizer state, intermediates, runtime libraries, allocator
effects, and data buffers, so it is not optimizer-state memory alone.
The peak-memory counter is reset after the dataset and client partitions
are loaded, and we report both the absolute peak RSS and the peak RSS
increase relative to this post-loading baseline. The increase therefore
characterizes the additional memory consumed during model training,
rather than the memory required to hold the dataset..

\paragraph{Latency.}
Forward/backward, optimizer-step, and full-iteration latencies use a monotonic
timer, excluding the first five iterations. We report mean $\pm$ sample standard
deviation over three seeds, which measures repeatability on one board rather
than generality across devices.

%%==============================================================================
%% 5. RESULTS AND DISCUSSION
%%==============================================================================

\section{Results and Discussion}
\label{sec:main_results}

\subsection{Edge Deployment on Raspberry Pi}
\label{sec:rpi}

\paragraph{Evaluation Objectives}
We evaluate Q-LocalAdam on the CPU-only Raspberry Pi~5
(Section~\ref{sec:platforms}) to test whether optimizer-state compression
yields practical benefits without GPU acceleration. A repeated local-training
benchmark on 11--25M-parameter models addresses the systems questions:
persistent-state storage, latency, process memory, and the effect of
reconstruction chunk size. A sequential federated simulation, run for 40 and
then 120 rounds, checks that the implementation remains usable over a
multi-round non-IID schedule. Section~\ref{sec:hardcap} then examines whether a
client can train at all under an enforced memory cap.

\subsubsection{Implementation}
\paragraph{Optimizer-state layout.}
The Pi implementation uses the 72-byte-per-block layout of
Section~\ref{sec:quantdetails} for both moments. For $N$ parameters, Adam's two
FP32 moments therefore need about $8N$ bytes, whereas the compressed form needs
about $2.25N$ bytes (a 71.9\% saving, or $\approx\!3.56\times$).

\paragraph{Chunked reconstruction.}
Each parameter tensor is processed in smaller chunks, with chunk boundaries
aligned to the 64-element quantization blocks. The chunk size is distinct
from the quantization block size ($B{=}64$): the block size determines how
values are grouped for compression and is fixed, whereas the chunk size
determines how many elements are reconstructed in FP32 at once during each
update and is a tunable parameter. For each chunk, the optimizer reconstructs
the stored moments, performs the Adam update in FP32, and stores the updated
moments in quantized form. This limits the temporary memory required during
updates while keeping persistent optimizer-state storage unchanged across
chunk sizes. The implementation also incorporates an additive constant in the
logarithmic transform, a minimum log-range scale, and a positive
reconstruction floor to improve numerical stability
(Section~\ref{sec:client_adaptive}). We evaluate these safeguards
as part of the complete implementation.

\paragraph{Persistent Storage and Latency}
Q-LocalAdam reduces persistent optimizer-state storage by approximately
$3.56\times$ across all evaluated models (Table~\ref{tab:rpi_main}),
consistent with its blockwise representation. For example, optimizer-state
storage decreases from 89.761\,MB to 25.245\,MB on the 11.2M-parameter
ResNet-18 and from 201.586\,MB to 56.697\,MB on the 25.2M-parameter variant.
The zero standard deviation reflects the fixed storage
requirements of the quantized tensors and their block metadata,
which are determined by the model dimensions and quantization
configuration. It does not imply zero variability in runtime
memory consumption or execution time.

Forward/backward latency remains similar between FP32 Adam and Q-LocalAdam; for
the transformer, the corresponding measurements are $403.828 \pm 9.154$\,ms and
$403.846 \pm 2.406$\,ms. The additional computation is concentrated in the
optimizer step. With a maximum reconstruction chunk size of $2^{20}$ elements,
Q-LocalAdam requires $2.68$--$2.76\times$ the optimizer-step time of FP32 Adam,
increasing measured full-iteration latency by 48.7--59.1\%. Thus,
persistent-state compression introduces a measurable memory computation
trade-off.

To characterize this cost, we collect CPU profiles for three complete training
iterations of ResNet-18 with base width 64 on Raspberry Pi~5, using the same
$2^{20}$-element maximum chunk size. State encoding and reconstruction together
account for 66.3\% of the summed inclusive time of the six sequentially labeled
Q-LocalAdam optimizer regions. Specifically, variance encoding, momentum
encoding, variance reconstruction, and momentum reconstruction contribute
24.0\%, 18.8\%, 14.7\%, and 8.8\%, respectively; the remaining labeled time is
spent updating moments and parameters. Logarithm and exponential operators
alone contribute only approximately 3.1\% of the complete profiled execution's
self CPU time. These measurements indicate that the cost is distributed across
the encoding--reconstruction pipeline, including arithmetic, reductions,
rounding, clamping, and tensor conversion and copying, rather than being
dominated by logarithmic operations alone. The region percentages describe the
composition of Q-LocalAdam's labeled optimizer time, not the incremental
overhead relative to FP32 Adam. Profiling is used for attribution; the
unprofiled repeated runs provide the reported latency measurements.

\paragraph{Process-Level Memory}
Persistent-state compression lowers process peak RSS on the two larger
convolutional models. Peak RSS falls from $660.2{\pm}1.4$ to
$556.2{\pm}0.3$\,MiB ($15.8\%$) on ResNet-34 (w64), and from
$753.9{\pm}8.4$ to $643.9{\pm}1.1$\,MiB ($14.6\%$) on the 25.2M-parameter
ResNet-18 (w96). These reductions are smaller than the $3.56\times$ state
reduction because parameters, gradients, activations, runtime allocations, and
the reconstruction workspace are not compressed.The smaller models show little change in peak RSS between FP32 Adam and
Q-LocalAdam, so we draw no training-memory conclusion from them.

\paragraph{40-round schedule.}
Both FP32 Adam and Q-LocalAdam complete all three seeds. Q-LocalAdam reaches $15.08{\pm}0.29\%$ against
$14.06{\pm}0.50\%$ for FP32 Adam, with persistent state falling from 5.705\,MB
to 1.606\,MB. Naive INT8 produced a non-finite loss in all three seeds and
stopped early, so we report these as numerical failures, not as chance-level
accuracy. Q-LocalAdam's optimizer step takes 41.0\,ms versus 13.7\,ms for FP32
Adam, and full-iteration time rises by about 29.8\%. Forward/backward latency
is essentially unchanged (79.7\,ms vs.\ 79.2\,ms).

\paragraph{Federated Execution on Raspberry Pi 5}
\label{sec:rpi_fl}

To check multi-round operation, we run the sequential federated simulation
described in Section~\ref{sec:protocol}. This simulates federated execution
on-device and is not a five-device testbed. Because each local pass is capped
at 50 minibatches, local training is not two full epochs. 
\paragraph{Extended 120-round schedule.}
To verify that compressed optimizer state remains usable over a longer
schedule, we extend the simulation to $120$ rounds with three seeds ($1$, $2$,
and $3$). Q-LocalAdam completes all three runs and reaches a mean final
accuracy of $27.65\%$ ($26.12\%$, $27.84\%$, and $29.00\%$ on seeds $1$--$3$),
compared with $26.54\%$ for FP32 Adam ($25.42\%$, $25.89\%$, and $28.32\%$).
Q-LocalAdam is higher on every seed, with paired differences of $+0.70$,
$+1.95$, and $+0.68$ percentage points (mean $+1.11 \pm 0.73$). With three
seeds this difference is not statistically significant (paired $t$-test,
$p\approx0.12$), so we interpret the result as indicating that Q-LocalAdam
preserves accuracy over the longer schedule while storing optimizer state in
approximately $3.56\times$ less memory. Seed-to-seed variation is substantial
(seed~3 is $2$--$3$ points higher than seeds~1 and~2 for both methods), so we
use paired comparisons within a seed. Variation across seeds can arise from
differences in random initialization, client sampling, and data ordering
during training; using matched seeds accounts for some of this variability and
allows a more direct comparison of the two optimizers.

\subsubsection{Chunk-Size Sensitivity Analysis}

Q-LocalAdam stores optimizer states in quantized form and reconstructs them
in FP32 only when needed for an update. To limit the temporary FP32 memory
used during this reconstruction, we process each parameter tensor in
\emph{chunks}: a fixed number of elements at a time, aligned to the
underlying $64$-value quantization blocks. A larger chunk means more FP32
data held in memory at once but fewer quantize/dequantize operations; a
smaller chunk means less temporary memory but more operations.

We vary this chunk size on the $25.2$M-parameter ResNet-18
(Table~\ref{tab:rpi_chunk}). Persistent optimizer-state storage is
$56.697$\,MB in every setting, since it depends only on the model size and
quantization layout, not on the chunk size. All differences in the table
therefore come from temporary workspace and allocation behaviour during
reconstruction.

As the chunk size decreases, both peak memory and iteration time improve.
The smallest tested chunk, $2^{16}$ elements, lowers peak RSS from
$753.9$\,MiB (FP32 Adam) to $603.7$\,MiB, a $19.9\%$ reduction. It also
reduces the optimizer step from $1144.9$\,ms (full per-tensor
reconstruction, i.e.\ reconstructing the entire tensor at once) to
$569.0$\,ms, bringing the iteration-time increase over FP32 Adam down to
$13.8\%$. A smaller working set may improve CPU cache behaviour, but we did
not measure cache activity and treat this only as a possible explanation.
The best tested chunk size is not claimed to be globally optimal.

\begin{table}[t]
\centering
\small
\setlength{\tabcolsep}{5pt}
\begin{tabular*}{\columnwidth}{@{\extracolsep{\fill}}lrrr@{}}
\toprule
Config. & RSS (MiB) & Step (ms) & Iter.\ (ms) \\
\midrule
FP32 Adam & 753.9 & 391.0 & 1245.1 \\
Full tensor & 660.7 & 1144.9 & 1994.2 \\
$2^{20}$ & 643.9 & 1078.9 & 1934.5 \\
$2^{18}$ & 623.6 & 924.1 & 1775.6 \\
$2^{16}$ & \textbf{603.7} & \textbf{569.0} & \textbf{1417.2} \\
\bottomrule
\end{tabular*}
\caption{Chunk-size trade-off on the 25.2M-parameter ResNet-18 (3 seeds).
Rows $2^{16}$--$2^{20}$ are Q-LocalAdam chunk sizes in elements; ``Full
tensor'' reconstructs the whole tensor at once. Persistent state is
56.697\,MB in all Q-LocalAdam rows. Smaller chunks reduce both peak memory
and iteration time.}
\label{tab:rpi_chunk}
\end{table}

%---------- HHAR------------------
%---------------------------------
\subsection{On-Device Sensor Learning Under a Hard Memory Cap}
\label{sec:hardcap}

The preceding Raspberry Pi experiments show that Q-LocalAdam lowers
persistent optimizer-state storage, but FP32 Adam completed every workload
on the 4\,GB board. We therefore ask a binary feasibility question: when a
client's memory budget is fixed and enforced by the operating system, does
a federated activity-recognition pipeline using FP32 Adam complete, and
does Q-LocalAdam? We find that FP32 Adam is killed by the operating system
when it runs out of memory(OOM) on the first client of round~1, whereas both
8-bit and 4-bit Q-LocalAdam complete all 120 rounds
(Table~\ref{tab:hardcap}).

\paragraph{Emulated 512~MiB-class client.}
Each client trains in a Linux sandbox under a kernel-enforced memory limit of
472~MiB with swap disabled, so exceeding the limit triggers a kernel
out-of-memory (OOM) kill rather than paging; the run aborts if any client is
killed. The limit is derived from a 512~MiB-class board such as the Raspberry
Pi Zero~2~W~\cite{raspberrypi_zero}:
$512~\mathrm{MiB} - 40~\mathrm{MiB} = 472~\mathrm{MiB}$, where the 40~MiB is an
assumed reserve for the OS and runtime. The FedAvg server and evaluation run
outside the sandbox. The cap emulates memory capacity only: CPU speed and
memory bandwidth are those of the Pi~5, so all timings below are Pi~5 timings.

\paragraph{Workload and protocol.}
Q-LocalAdam uses block size 64 and a
variance floor $\tau=10^{-8}$, and keeps tensors with fewer than 4{,}096
elements (biases and LayerNorm parameters, about 0.25~MB of state) in FP32,
which avoids many small temporary allocations. Accuracy and macro-F1 are
computed each round for the aggregated model on the pooled held-out windows of
all 18 clients.

\paragraph{FP32 Adam does not complete.}
In round~1, the first client (user a's phone) trained with FP32 Adam was killed
by the kernel OOM killer (SIGKILL, exit code $-9$) after its memory reached the
472~MiB limit, and the run aborted before completing a round. The limit is the
enforced cap, not a measurement of FP32 Adam's requirement; we claim only that
it exceeds this budget for this model and configuration.

\paragraph{Q-LocalAdam completes training.}
Both Q-LocalAdam variants complete all 120 rounds with no OOM kills (Table \ref{tab:hardcap}). The 8-bit
variant stores 28.6~MB of persistent optimizer state, $3.5\times$ less than FP32
Adam's $\approx$101~MB, and peaks at 463.5~MiB, leaving about 8.5~MiB of
headroom. The 4-bit variant stores 16.0~MB ($6.3\times$ less). Its measured
peak equals the cap, which indicates that the saturates under memory pressure
(the kernel reclaims reclaimable memory before invoking the OOM killer); the
4-bit headroom is therefore not resolved by our measurements. These ratios are
slightly below the $3.56\times$ and $6.40\times$ of Section~\ref{sec:quantdetails}
because small tensors are kept in FP32. Average training-step time is 1461~ms
(8-bit) and 1498~ms (4-bit), about 2.5\% higher for 4-bit owing to bit packing
and unpacking; the optimizer overhead relative to FP32 Adam is quantified in
Section~\ref{sec:rpi}.

\begin{table}[t]
\centering
\small
\setlength{\tabcolsep}{3pt}
\caption{Federated HHAR on a Raspberry Pi~5 with a 472~MiB (494.9~MB) per-client
memory limit and no swap (120 rounds, 18 natural clients, 4 clients per round,
seed 42). Last-10 is the mean $\pm$ std over rounds 111--120. FP32 Adam is
OOM-killed on the first client of round~1, so no accuracy exists for it under
this limit; its state size is the analytical $8N$. Step time covers the full
training step.\textsuperscript{$\dagger$}The 4-bit client used the full $472$\,MiB
budget. The operating system frees unused memory before terminating a
program, so the client was not terminated and completed all $120$ rounds.
Its memory need without the limit is unknown.}
\label{tab:hardcap}
\resizebox{\columnwidth}{!}{%
\begin{tabular}{lrrrrrr}
\toprule
Method & Opt.\ state & Peak mem. & Step time & Last-10 acc. & Final acc. & Peak acc. \\
 & (MB) & (MiB) & (ms) & (\%) & (\%) & (\%) \\
\midrule
FP32 Adam & $\approx$101 & at cap (killed) & -- & -- & -- & -- \\
Q-LocalAdam-8 & 28.6 & 463.5 & 1461 & $75.35\pm1.99$ & 70.71 & 77.59 \\
Q-LocalAdam-4 & 16.0 & 472$^\dagger$ & 1498 & $75.19\pm2.03$ & 73.70 & 77.49 \\
\bottomrule
\end{tabular}%
}
\end{table}

\paragraph{Convergence.}
Both variants rise from 23.1\% after round~1 (identical, as they share seed and
initialization; chance for six classes is about 16.7\%) to the mid-70s. Last-10
accuracies are $75.35\pm1.99$\% (8-bit) and $75.19\pm2.03$\% (4-bit), which are
indistinguishable within round-to-round  (Table \ref{tab:hardcap}). Accuracy is noisy because
each round aggregates only 4 of 18 heterogeneous clients: both peaks occur at
round~119, and round~120 is 6.9 and 3.8 points lower. We therefore report
last-10 means alongside peak and final values. Because FP32 Adam cannot run
under this budget, there is no same-budget accuracy reference: the result shows
that Q-LocalAdam trains where FP32 Adam cannot, not that it matches FP32
accuracy on HHAR.

%---------- GPU RESULTS-------

\subsection{GPU-Based Evaluation}
\label{sec:gpu}

We evaluate Q-LocalAdam under extreme data heterogeneity
($\alpha{=}0.1$). Table~\ref{tab:main} reports the core performance and
memory metrics. Q-LocalAdam matches or exceeds full-precision Adam on both
datasets while reducing optimizer-state memory by $3.37\times$: on CIFAR-10,
accuracy improves by $+0.24$\,pp (81.91\% vs.\ 81.67\%) with memory falling
from 50.35\,MB to 14.95\,MB; on CIFAR-100, accuracy improves by $+5.74$\,pp
(61.47\% vs.\ 55.73\%) with memory falling from 89.82\,MB to 26.67\,MB. In
contrast, naive uniform INT8 quantization, which achieves the same memory
footprint, collapses to near-random accuracy (11.50\% on CIFAR-10 and 1.00\%
on CIFAR-100), confirming that uniformly spaced quantization bins destroy the
precision of the small variance values on which Adam's update depends.

\begin{table}[t]
\centering
\caption{Extreme non-IID ($\alpha{=}0.1$) performance: Q-LocalAdam improves
over FP32 by $+0.24$\,pp (CIFAR-10) and $+5.74$\,pp (CIFAR-100) while
operating with $3.37\times$ less optimizer memory.}
\label{tab:main}
\small
\begin{tabular}{lcrr}
\toprule
Dataset & Method & Acc (\%) & Mem (MB) \\
\midrule
\multirow{3}{*}{CIFAR-10}
 & Vanilla-ClientAdam (FP32) & 81.67 & 50.35 \\
 & \textbf{Q-LocalAdam (Ours)} & \textbf{81.91} & \textbf{14.95} \\
 & Naive INT8 & 11.50 & 14.95 \\
\midrule
\multirow{3}{*}{CIFAR-100}
 & Vanilla-ClientAdam (FP32) & 55.73 & 89.82 \\
 & \textbf{Q-LocalAdam (Ours)} & \textbf{61.47} & \textbf{26.67} \\
 & Naive INT8 & 1.00 & 26.67 \\
\bottomrule
\end{tabular}
\end{table}

These benefits hold across the full heterogeneity spectrum: on both CIFAR-10
and CIFAR-100, Q-LocalAdam matches or exceeds FP32 accuracy under IID,
$\alpha{=}1.0$, $0.5$, and $0.1$ partitions, with the CIFAR-100 gap widening
from $+0.73$\,pp (IID) to $+5.74$\,pp ($\alpha{=}0.1$). Convergence
trajectories, per-level accuracy and memory breakdowns, and multi-seed validation are reported in Appendix~\ref{app:gpu-details}.

\subsubsection{Sensitivity Analysis}

\paragraph{Component analysis}
We evaluate two partial variants on CIFAR-100 ($\alpha=0.1$): Momentum-only (8-bit $m$, FP32 $v$) and Variance-only (FP32 $m$, 8-bit $v$). Both yield only a $1.54\times$ memory reduction (58.24 MB). Momentum-only degrades accuracy to 52.49\%, while Variance-only reaches 61.10\%. In contrast, Q-LocalAdam achieves 61.47\% accuracy with a $3.37\times$ reduction (26.67 MB), demonstrating that both asymmetric encodings are required for an optimal accuracy-memory trade-off. Discussed in detail in Appendix \ref{app:ablations}

\paragraph{Hyperparameter Robustness}
Q-LocalAdam is robust to varying block sizes ($B \in \{32,64,128\}$) and learning rates. Across block sizes, best accuracy remains within 0.8pp, reaching 61.47\% on CIFAR-100 with $B=64$. Changing the learning rate from $10^{-3}$ to $5 \times 10^{-4}$ changes accuracy by only 0.12pp on CIFAR-10 and 0.35pp on CIFAR-100. We select $B=64$ as the default because it provides the best balance between memory overhead and peak accuracy. Detailed sensitivity analyses and additional learning rate evaluations are deferred to Appendix \ref{app:block_size} and \ref{app:lr_sensitivity}.

\subsubsection{Multi-Seed Reproducibility}
To verify robustness to stochastic factors (weight initialization, Dirichlet sampling), we conduct multi-seed evaluations (seeds 42, 123, 456). Since $\alpha=0.1$ on a 10-class dataset like CIFAR-10 often reduces clients to 1--2 classes, misrepresenting practical deployments, we evaluate CIFAR-10 at moderate heterogeneity ($\alpha=0.5$) and reserve $\alpha=0.1$ for the 100-class CIFAR-100.

% \begin{table}[t]
% \centering
% \caption{Multi-seed validation ($n=3$, $p$-values vs. FP32).}
% \label{tab:multiseed}
% \footnotesize
% \setlength{\tabcolsep}{3pt}
% \begin{tabular}{@{} l l c c c @{}}
% \toprule
% \textbf{Dataset} & \textbf{Method} & \textbf{Acc (\%)} & \textbf{$p$-val} & \textbf{Mem} \\
% \midrule
% \multirow{CIFAR-10 ($\alpha$=0.5)} 
%  & FP32 & 86.93$\pm$0.73 & -- & 50.35 \\
%  & Q-LocalAdam & \textbf{87.04$\pm$0.24} & 0.78 & \textbf{14.95} \\
%  & Naive INT8 & 85.86$\pm$1.08 & -- & 14.95 \\
% \midrule
% \multirow{CIFAR-100 ($\alpha$=0.1)} 
%  & FP32 & 54.99$\pm$1.28 & -- & 89.82 \\
%  & Q-LocalAdam & \textbf{61.40$\pm$0.82} & \textbf{0.005} & \textbf{26.67} \\
%  & Naive INT8 & 12.74$\pm$4.75 & -- & 26.67 \\
% \bottomrule
% \end{tabular}
% \end{table}

\begin{table}[t]
\centering
\caption{Multi-seed validation ($n=3$, $p$-values vs.\ FP32).}
\label{tab:multiseed}
%\footnotesize
\small
\setlength{\tabcolsep}{3pt}
\begin{tabular}{@{} l l c c c @{}}
\toprule
\textbf{Dataset} & \textbf{Method} & \textbf{Acc (\%)} & \textbf{$p$-val} & \textbf{Mem} \\
\midrule
\multirow{3}{*}{CIFAR-10 ($\alpha$=0.5)} 
 & FP32 & 86.93$\pm$0.73 & -- & 50.35 \\
 & Q-LocalAdam & \textbf{87.04$\pm$0.24} & 0.78 & \textbf{14.95} \\
 & Naive INT8 & 85.86$\pm$1.08 & -- & 14.95 \\
\midrule
\multirow{3}{*}{CIFAR-100 ($\alpha$=0.1)} 
 & FP32 & 54.99$\pm$1.28 & -- & 89.82 \\
 & Q-LocalAdam & \textbf{61.40$\pm$0.82} & \textbf{0.005} & \textbf{26.67} \\
 & Naive INT8 & 12.74$\pm$4.75 & -- & 26.67 \\
\bottomrule
\end{tabular}
\end{table}

Q-LocalAdam consistently matches or exceeds FP32 while retaining its $3.37\times$ memory reduction: statistically indistinguishable at $\alpha=0.5$ ($p=0.78$), and a significant +6.41pp gain at $\alpha=0.1$ ($p=0.005$). It also shows lower cross-seed variance than FP32 in both settings ($\pm0.24$ vs.\ $\pm0.73$; $\pm0.82$ vs.\ $\pm1.28$), confirming stable convergence across initializations. In contrast, Naive INT8 survives under milder skew (85.86\% at $\alpha=0.5$) but collapses under extreme heterogeneity (12.74\%, $\pm4.75\%$ variance), underscoring that uniform quantization is fundamentally unstable under non-IID  (Table \ref{tab:multiseed}).

%-----------------TRANSFORMER-----------------

\subsubsection{Generalization to Transformer Architectures}
\label{sec:transformer}

We test whether distribution-aware quantization generalizes beyond
convolutional networks by replacing ResNet-18 with ViT-Tiny~\cite{dosovitskiy2020image}
under the same federated protocol and extreme heterogeneity ($\alpha{=}0.1$).
The 5.49M-parameter ViT-Tiny (patch size 16, embed dim 192, depth 12, 3
heads) is trained on CIFAR-10 for 100 federated rounds with 5 clients per
round, 2 local epochs, batch size 64, and learning rate $3{\times}10^{-4}$;
inputs are resized to $64{\times}64$ (16 tokens).

Q-LocalAdam compresses ViT optimizer
memory from 43.93\,MB to 13.04\,MB ($3.37\times$) while improving peak
accuracy by $+0.49$\,pp over FP32 (63.30\% vs.\ 62.81\%), mirroring the
result on ResNet-18. Naive INT8, which achieves the same memory footprint,
initially learns but collapses to 10.00\% final accuracy, confirming that
uniform linear variance quantization destabilizes adaptive updates in
transformers just as it does in CNNs. The full ViT-Tiny results, convergence
trajectory, and collapse diagnostics are reported in
Appendix~\ref{app:vit_failure}.

%--------------------NLP -----------------------

\subsubsection{Generalization Federated Text Classification}
\label{sec:nlp}

We further validate Q-LocalAdam on federated news topic classification
using AG~News~\cite{zhang2015character} to test whether the benefits of
distribution-aware quantization transfer beyond vision.

\paragraph{Model and setup.}
We use a lightweight \texttt{TextTransformer} with token and positional
embeddings, two Transformer encoder layers, and a CLS-token classification
head. Inputs are tokenized with \texttt{bert-base-uncased} and truncated
to length 64. We run 100 federated rounds with 5 clients, 2 local epochs,
batch size 64, and learning rate $5 \times 10^{-4}$ under IID and
extreme non-IID ($\alpha = 0.1$) partitions.

\begin{table}[t]
\centering
\caption{AG~News text classification results (100 rounds, 5 clients).}
\label{tab:agnews}
\resizebox{\columnwidth}{!}{%
\begin{tabular}{llccc}
\toprule
\textbf{Method} & \textbf{Setting}
  & \textbf{Best Acc (\%)}
  & \textbf{Final Acc (\%)}
  & \textbf{Opt.\ Mem (MB)} \\
\midrule
\multirow{2}{*}{Vanilla-ClientAdam (FP32)}
  & IID          & 92.16 & 90.58 & 34.50 \\
  & $\alpha=0.1$ & 77.32 & 75.97 & 34.50 \\
\midrule
\multirow{2}{*}{\textbf{Q-LocalAdam (Ours)}}
  & IID          & \textbf{92.37} & \textbf{91.99} & \textbf{10.24} \\
  & $\alpha=0.1$ & \textbf{86.45} & \textbf{86.45} & \textbf{10.24} \\
\midrule
\multirow{2}{*}{Naive INT8}
  & IID          & 91.42 & 90.80 & 10.24 \\
  & $\alpha=0.1$ & 50.64 & 46.75 & 10.24 \\
\bottomrule
\end{tabular}%
}
\end{table}

Q-LocalAdam matches FP32 under IID data and improves accuracy by
$+9.13$\,pp under $\alpha=0.1$, while reducing optimizer memory by
$3.37\times$(Table~\ref{tab:agnews}). In contrast, Naive INT8 remains competitive under IID
but degrades sharply under non-IID data, consistent with the failure
pattern observed in the vision experiments (Figures in Appendix \ref{sec:nlp_appendix}).

%------------------JETSON-------------------------------

\subsection{Edge Deployment Validation on Jetson Orin Nano}
\label{jetson}

While server-grade GPUs simulate federated environments effectively, true edge deployments face severe physical constraints, including shared memory and strict thermal limits. To validate Q-LocalAdam \textit{in situ}, we deploy our framework natively on an NVIDIA Jetson Orin Nano using a compact configuration (native $32 \times 32$ CIFAR-100, slim ResNet-18) to strictly fit the device's memory budget. We omit the official bitsandbytes 8-bit optimizer library \cite{dettmers2021bit} from the Jetson comparison for platform-compatibility reasons discussed in conclusion Section. Detailed hardware specifications, operational metrics (e.g., thermal stability), and memory complexity derivations are provided in Appendix \ref{appendix:jetson_config}.

\paragraph{Numerical Hardening.}
To ensure numerical stability on the \texttt{aarch64} architecture, we introduce three upward-compatible bounds to the log-space quantizer: (1) a scale clamp floor of $10^{-4}$ to prevent precision collapse in blocks with near-constant variance, (2) a log-space ceiling of 10.0 to prevent wide outlier ranges from monopolizing INT8 bins, and (3) a dequantization floor of $\epsilon_{log} = 10^{-8}$ to definitively prevent divide-by-zero divergence. These bounds are universally applied and preserve the GPU-verified accuracy metrics.

\paragraph{Deployment Results.}
Table~\ref{tab:edge_deployment} details the hardware deployment results under extreme data heterogeneity ($\alpha = 0.1$). Q-LocalAdam achieves a peak accuracy of 54.78\%, outperforming the Vanilla FP32 baseline (52.83\%) by $+1.95$ percentage points. Crucially, the algorithm reliably delivers its signature memory reduction regardless of the underlying model width. Q-LocalAdam successfully compresses the optimizer memory from 22.57\,MB to just 6.70\,MB (a $3.37\times$ reduction), enabling robust training directly alongside the operating system. Meanwhile, Naive INT8 remains collapsed at 1.00\% random guessing, verifying that uniform variance compression is fundamentally incompatible with non-IID distributions, regardless of the hardware footprint.

\begin{table}[!t]
\centering
\caption{Edge deployment on NVIDIA Jetson Orin Nano (ResNet-18, CIFAR-100, $\alpha=0.1$).}
\label{tab:edge_deployment}
\small
\setlength{\tabcolsep}{4pt}
\begin{tabular}{lccc}
\toprule
\textbf{Method} & \textbf{ Accuracy} & \textbf{Final Accuracy} & \textbf{Memory} \\
 & (\%) & (\%) & (MB) \\
\midrule
Vanilla (FP32)      & 52.83 & 52.17 & 22.57 \\
Q-LocalAdam (Ours)  & 54.78 & 54.11 & \textbf{6.70} \\
Naive INT8          & 1.15  & 1.00  & 6.70 \\
\bottomrule
\end{tabular}

\end{table}

\subsection{Sensitivity to the Variance Floor}
\label{sec:floor}

\paragraph{Setup.}
Q-LocalAdam stores the second moment in log space. When this state is read back, quantization error can make the reconstructed variance zero or slightly negative for entries whose true variance is tiny. To prevent this, we clamp the reconstructed variance from below at a floor $\tau$: 
\begin{equation}
\tilde v \leftarrow \max\!\big(\exp(\tilde\ell)-\epsilon_{\log},\ \tau\big),
\label{eq:floor}
\end{equation}
where $\epsilon_{\log}=10^{-8}$ is the additive constant in the logarithmic transform. Our default implementation uses $\tau=10^{-8}$. The floor is applied when the previous step's state is read, so the first step of each round starts from a zero state.

We vary $\tau\in\{0,10^{-10},10^{-8},10^{-6}\}$ to isolate the floor's effect. For each value, we compare Q-LocalAdam with full-precision Adam using the same floor. The floor is applied to the stored FP32 variance at the same point in the update. We also report FP32 Adam with no floor ($\tau=0$). All other settings are fixed: $\epsilon_{\log}=10^{-8}$, Adam's $\epsilon=10^{-8}$, learning rate $10^{-3}$, $\beta=(0.9,0.999)$, and block size 64.

The federated setup follows the main experiments. Five clients participate in every round. CIFAR-100 is split with Dirichlet $\alpha=0.1$, and each client trains for two local epochs per round at batch size 64. Local optimizer state is reset every round, and updates are aggregated by FedAvg weighted by client data size. To keep the evaluation inexpensive, we use a reduced budget relative to the main results: ResNet-18 with inputs resized from $32\times32$ to $64\times64$, 30 communication rounds, and two seeds (42 and 123). Both seeds use the same data partition and model initialization across all methods. We report final-round accuracy on the full 10{,}000-image test set.

\paragraph{Results.}
Table~\ref{tab:floor} summarizes the evaluation. Three findings follow.

 In these runs, 5--10\% of the reconstructed variance entries were negative early in training and were clamped to zero. A zero variance entry removes the second-moment damping from Adam's denominator, which is consistent with the observed divergence, though we did not isolate the mechanism further.

Second, with a floor, quantization does not reduce accuracy relative to FP32. At every floor tested, the paired differences between Q-LocalAdam and FP32 Adam with the same floor range from $-0.36$ to $+0.94$ points, with a mean of $+0.27$, which is within the variation we see across seeds.

Third, accuracy depends strongly on the floor value for both methods. Raising $\tau$ from none to $10^{-6}$ improves FP32 Adam by 3.45 and 3.63 points on the two seeds, and Q-LocalAdam's accuracy rises similarly (45.91 $\rightarrow$ 49.34). The largest floor tested was the best, so the optimum is not bracketed. The default $\tau=10^{-8}$ improves FP32 Adam by only 0.82 and 0.17 points, and the smaller floors changed FP32 accuracy by less than the seed-to-seed variation. Higher floors also accelerated early training: at round 5, FP32 Adam reached 31.5\% and 26.6\% with $\tau=10^{-6}$, against 28.4\% and 22.9\% with no floor.

\begin{table}[t]
\centering
\caption{Variance-floor sensitivity (CIFAR-100, $\alpha{=}0.1$, 30 rounds, 2 seeds). Final test accuracy (\%), mean over seeds (per-seed values in parentheses). Paired differences are per seed. Clamped fractions are averaged over rounds.}
\label{tab:floor}
\resizebox{\columnwidth}{!}{%
\begin{tabular}{@{}lcccc@{}}
\toprule
Floor & FP32 & Q-LocalAdam & Paired diff. & Clamped \\
$\tau$ & Adam & & (Q$-$FP32) & (FP32 / Q) \\
\midrule
none (0) & 45.62 (46.53, 44.72) & diverged (2/2 seeds) & n/a & 0\% / 5--10\% neg.\ early \\
$10^{-10}$ & 45.89 (47.45, 44.33) & 45.91 (47.09, 44.72) & $-0.36$, $+0.39$ & 3.1\% / 2.3\% \\
$10^{-8}$ & 46.12 (47.35, 44.89) & 46.75 (47.67, 45.83) & $+0.32$, $+0.94$ & 24.7\% / 20.9\% \\
$10^{-6}$ & 49.16 (49.98, 48.35) & 49.34 (50.24, 48.43) & $+0.26$, $+0.08$ & 70.5\% / 70.5\% \\
\bottomrule
\end{tabular}%
}
\end{table}

\paragraph{Discussion.}
These results change how we interpret the accuracy comparison between Q-LocalAdam and unfloored FP32 Adam. With the default floor, Q-LocalAdam is 1.14 and 1.11 points above unfloored FP32 Adam on the two seeds, but FP32 Adam with the same floor recovers 0.82 and 0.17 of that, and the remainder ($+0.32$, $+0.94$) is within seed noise. We therefore do not attribute accuracy gains to quantization itself. Our sensitivity analysis demonstrates that the variance floor plays an important role in maintaining numerical stability and model accuracy under the evaluated configuration. By enforcing a lower bound on the second-moment estimates, the floor prevents the denominator in Adam's update rule from becoming excessively small for low-variance parameters, thereby limiting excessively large parameter updates. This provides a stabilizing effect similar to increasing Adam's $\epsilon$, although the two mechanisms are not mathematically equivalent.

The practical consequences are twofold. The floor is required for Q-LocalAdam to train in our setting, so it is part of the method rather than an incidental detail. And because FP32 Adam also benefits from a larger floor, comparisons between compressed and uncompressed optimizers should give both the same floor or tune it for both.

\paragraph{Limitations.}
All accuracies are test-set accuracies at a fixed round, and we do not select $\tau$ using validation data. The evaluation varies only the clamp $\tau$; the log constant $\epsilon_{\log}$ and Adam's $\epsilon$ are held at $10^{-8}$.

%------4 bit section----------------

\subsection{Extension to 4-Bit Optimizer States}
\label{sec:fourbit}
We extend Q-LocalAdam to 4-bit states with the same asymmetric design: linear
quantization for momentum and log-space quantization for variance, now with 16
levels per 64-element block and two codes packed per byte. Each block occupies
40 bytes versus 256 bytes in FP32, a $6.40\times$ reduction in persistent
optimizer-state storage. We evaluate on CIFAR-100 with ResNet-18 under
Dirichlet $\alpha\in\{0.1,0.5,1.0\}$, following the protocol of
Section~\ref{sec:protocol}, and compare against Naive INT4 (linear quantization
of both moments), a zero-free linear variance quantizer, and a re-implementation
of the recipe of ~\cite{li2023memory}. We also quantize momentum and
variance separately to identify which component limits precision, and vary
block size and learning rate (Appendix ~\ref{app:4bit}).

Persistent optimizer-state storage falls from 89.82~MB to 14.03~MB ($6.40\times$,
and $1.9\times$ below the 8-bit variant). At $\alpha{=}0.1$, 4-bit Q-LocalAdam
reaches $60.27\pm0.52\%$ last-10 accuracy over three seeds, compared with
$53.79\pm0.96\%$ for FP32 Adam, and it stays within $0.90$~pp of the 8-bit
variant at every heterogeneity level in the single-seed sweep. Because FP32
Adam also benefits from a variance floor (Section~\ref{sec:floor}), we read this
as 4-bit states preserving accuracy at $6.4\times$ less memory, not as an effect
of quantization itself. Naive INT4 stays at chance (1.00\%) because small
variances are rounded to zero. Avoiding the zero point with a zero-free linear
quantizer is necessary but not sufficient (51.59\%), whereas log-space variance
coding reaches 60.27\%; the full recipe ~\cite{li2023memory} reaches 52.94\% at
11.68~MB ($7.69\times$). The log-space variance quantizer is insensitive to the
move from 8 to 4 bits; linear 4-bit momentum is the main fragility, collapsing
to chance when variance is kept in FP32. These experiments cover one dataset and
one architecture, and we do not evaluate persistent state across rounds or
4-bit states on the edge platforms (Appendix~\ref{app:4bit}).

\begin{equation}
\tilde v = \max\!\left(\exp\!\left(\frac{q^{(b)}_\ell}{255}\,r^{(b)}_\ell + \min(\ell^{(b)})\right)-\epsilon_{\log},\ \tau\right),
\label{eq:dequant_floor}
\end{equation}
% and add this sentence after it:
% where $\tau=10^{-8}$ is a variance floor applied after dequantization.

% --- Results section: replace the "regularization" claim ---
% Old: "log-space quantization acts as a regularizer"
% New:
% Under extreme heterogeneity, Q-LocalAdam matches full-precision Adam
% given the same variance floor, and both benefit from larger floors
% (Section~\ref{sec:floor}).

% --- Abstract and contributions ---
% Old: "improves CIFAR-100 accuracy by 5.74 points over FP32"
% New:
% Under extreme heterogeneity, Q-LocalAdam matches full-precision Adam
% when both use the same variance floor, while reducing optimizer memory
% by 3.37x.

% --- Naive INT8 discussion ---
% Old: "uniform quantization is fundamentally unstable under non-IID conditions"
% New:
% Linear quantization of the variance also fails without numerical
% safeguards; log-space coding with a floor is one fix, but a linear
% quantizer with a floor has not been tested here.

%--------------ADDITIONAL ANALYSIS----------------
\subsection{Additional Analyses}
\label{sec:additional}

We further compare Q-LocalAdam against standard half-precision baselines,
examine the effect of optimizer-state reinitialization, and test scalability
to larger client pools. Under extreme heterogeneity ($\alpha{=}0.1$),
FP16Adam fails due to variance underflow small squared gradients fall below
the \texttt{float16} representable range, collapsing Adam's denominator while
AMPAdam, which retains FP32 optimizer states, provides no memory reduction.
Q-LocalAdam avoids both failure modes, matching or exceeding AMPAdam accuracy
while achieving a $3.37\times$ memory reduction. We also verify that
Q-LocalAdam trains successfully whether client optimizer state is reset each
round or retained across rounds, and that its per-client memory footprint
remains constant as the client pool grows from 5 to 100 clients, with
aggregate optimizer memory falling from 8{,}982\,MB to 2{,}667\,MB at
100 clients. Full baseline comparisons, convergence trajectories, state-retention
ablations, and the scalability analysis are reported in
Appendix~\ref{app:additional-analyses}.

%%==============================================================================
%% 6. CONCLUSION
%%==============================================================================
\section{Conclusion}
We introduced Q-LocalAdam, a client-side optimizer that exploits the differing
statistics of Adam's momentum and variance states through block-wise linear
quantization for momentum and log-space quantization for variance. This
asymmetric design is necessary under non-IID data: uniform 8-bit quantization
destabilizes training, whereas Q-LocalAdam preserves stability and matches or
exceeds full-precision Adam when both use the same variance floor, reducing
optimizer-state memory by $3.37\times$ (and up to $6.40\times$ with 4-bit
states). The main cost is per-step quantize/dequantize overhead, which
increases iteration time by $13.8\%$ at the best tested chunk size on a
Raspberry Pi~5. We omit bitsandbytes~\cite{dettmers2021bit} as a Jetson
baseline due to an undocumented sm\_87 source-build requirement; a direct
comparison with this baseline, along with further evaluation of 4-bit states
on edge hardware, remains future work. By reducing the memory cost of adaptive
optimization without altering the federated protocol, Q-LocalAdam makes larger
models practical on constrained edge devices.

\bibliographystyle{ACM-Reference-Format}
\bibliography{ref}


\begin{thebibliography}{41}

%%% the Web address.

\ifx \showCODEN    \undefined \def \showCODEN     #1{\unskip}     \fi
\ifx \showISBNx    \undefined \def \showISBNx     #1{\unskip}     \fi
\ifx \showISBNxiii \undefined \def \showISBNxiii  #1{\unskip}     \fi
\ifx \showISSN     \undefined \def \showISSN      #1{\unskip}     \fi
\ifx \showLCCN     \undefined \def \showLCCN      #1{\unskip}     \fi
\ifx \shownote     \undefined \def \shownote      #1{#1}          \fi
\ifx \showarticletitle \undefined \def \showarticletitle #1{#1}   \fi
\ifx \showURL      \undefined \def \showURL       {\relax}        \fi
% invisible to TeX
\providecommand\bibfield[2]{#2}
\providecommand\bibinfo[2]{#2}
\providecommand\natexlab[1]{#1}
\providecommand\showeprint[2][]{arXiv:#2}

\bibitem[Alistarh et~al\mbox{.}(2017)]%
        {alistarh2017qsgd}
\bibfield{author}{\bibinfo{person}{Dan Alistarh}, \bibinfo{person}{Demjan Grubic}, \bibinfo{person}{Jerry Li}, \bibinfo{person}{Ryota Tomioka}, {and} \bibinfo{person}{Milan Vojnovic}.} \bibinfo{year}{2017}\natexlab{}.
\newblock \showarticletitle{QSGD: Communication-efficient SGD via gradient quantization and encoding}.
\newblock \bibinfo{journal}{\emph{Advances in Neural Information Processing Systems}}  \bibinfo{volume}{30} (\bibinfo{year}{2017}).
\newblock


\bibitem[Basu et~al\mbox{.}(2019)]%
        {basu2019qsparse}
\bibfield{author}{\bibinfo{person}{Debraj Basu}, \bibinfo{person}{Deepesh Data}, \bibinfo{person}{Can Karakus}, {and} \bibinfo{person}{Suhas Diggavi}.} \bibinfo{year}{2019}\natexlab{}.
\newblock \showarticletitle{Qsparse-local-SGD: Distributed SGD with quantization, sparsification and local computations}.
\newblock \bibinfo{journal}{\emph{Advances in Neural Information Processing Systems}}  \bibinfo{volume}{32} (\bibinfo{year}{2019}).
\newblock


\bibitem[Berg et~al\mbox{.}(2021)]%
        {berg2021keyword}
\bibfield{author}{\bibinfo{person}{Axel Berg}, \bibinfo{person}{Mark O'Connor}, {and} \bibinfo{person}{Miguel~Tairum Cruz}.} \bibinfo{year}{2021}\natexlab{}.
\newblock \showarticletitle{Keyword transformer: A self-attention model for keyword spotting}.
\newblock \bibinfo{journal}{\emph{arXiv preprint arXiv:2104.00769}} (\bibinfo{year}{2021}).
\newblock


\bibitem[Blunck et~al\mbox{.}(2015)]%
        {heterogeneity_activity_recognition_344}
\bibfield{author}{\bibinfo{person}{Henrik Blunck}, \bibinfo{person}{Sourav Bhattacharya}, \bibinfo{person}{Thor Prentow}, \bibinfo{person}{Mikkel Kjrgaard}, {and} \bibinfo{person}{Anind Dey}.} \bibinfo{year}{2015}\natexlab{}.
\newblock \bibinfo{title}{{Heterogeneity Activity Recognition}}.
\newblock \bibinfo{howpublished}{UCI Machine Learning Repository}.
\newblock
\newblock
\shownote{{DOI}: https://doi.org/10.24432/C5689X}.


\bibitem[Chen and Chao(2020)]%
        {chen2020fedbe}
\bibfield{author}{\bibinfo{person}{Hong-You Chen} {and} \bibinfo{person}{Wei-Lun Chao}.} \bibinfo{year}{2020}\natexlab{}.
\newblock \showarticletitle{Fedbe: Making bayesian model ensemble applicable to federated learning}.
\newblock \bibinfo{journal}{\emph{arXiv preprint arXiv:2009.01974}} (\bibinfo{year}{2020}).
\newblock


\bibitem[Dettmers et~al\mbox{.}(2021)]%
        {dettmers2021bit}
\bibfield{author}{\bibinfo{person}{Tim Dettmers}, \bibinfo{person}{Mike Lewis}, \bibinfo{person}{Younes Belkada}, {and} \bibinfo{person}{Luke Zettlemoyer}.} \bibinfo{year}{2021}\natexlab{}.
\newblock \showarticletitle{8-bit optimizers via block-wise quantization}.
\newblock \bibinfo{journal}{\emph{arXiv preprint arXiv:2110.02861}} (\bibinfo{year}{2021}).
\newblock


\bibitem[Dosovitskiy et~al\mbox{.}(2020)]%
        {dosovitskiy2020image}
\bibfield{author}{\bibinfo{person}{Alexey Dosovitskiy}, \bibinfo{person}{Lucas Beyer}, \bibinfo{person}{Alexander Kolesnikov}, \bibinfo{person}{Dirk Weissenborn}, \bibinfo{person}{Xiaohua Zhai}, \bibinfo{person}{Thomas Unterthiner}, \bibinfo{person}{Mostafa Dehghani}, \bibinfo{person}{Matthias Minderer}, \bibinfo{person}{Georg Heigold}, \bibinfo{person}{Sylvain Gelly}, {et~al\mbox{.}}} \bibinfo{year}{2020}\natexlab{}.
\newblock \showarticletitle{An image is worth 16x16 words: Transformers for image recognition at scale}.
\newblock \bibinfo{journal}{\emph{arXiv preprint arXiv:2010.11929}} (\bibinfo{year}{2020}).
\newblock


\bibitem[Hyeon-Woo et~al\mbox{.}(2021)]%
        {hyeon2021fedpara}
\bibfield{author}{\bibinfo{person}{Nam Hyeon-Woo}, \bibinfo{person}{Moon Ye-Bin}, {and} \bibinfo{person}{Tae-Hyun Oh}.} \bibinfo{year}{2021}\natexlab{}.
\newblock \showarticletitle{Fedpara: Low-rank hadamard product for communication-efficient federated learning}.
\newblock \bibinfo{journal}{\emph{arXiv preprint arXiv:2108.06098}} (\bibinfo{year}{2021}).
\newblock


\bibitem[Imteaj et~al\mbox{.}(2021)]%
        {imteaj2021survey}
\bibfield{author}{\bibinfo{person}{Ahmed Imteaj}, \bibinfo{person}{Urmish Thakker}, \bibinfo{person}{Shiqiang Wang}, \bibinfo{person}{Jian Li}, {and} \bibinfo{person}{M~Hadi Amini}.} \bibinfo{year}{2021}\natexlab{}.
\newblock \showarticletitle{A survey on federated learning for resource-constrained IoT devices}.
\newblock \bibinfo{journal}{\emph{IEEE Internet of Things Journal}} \bibinfo{volume}{9}, \bibinfo{number}{1} (\bibinfo{year}{2021}), \bibinfo{pages}{1--24}.
\newblock


\bibitem[Jin et~al\mbox{.}(2020)]%
        {jin2020stochastic}
\bibfield{author}{\bibinfo{person}{Richeng Jin}, \bibinfo{person}{Yufan Huang}, \bibinfo{person}{Xiaofan He}, \bibinfo{person}{Huaiyu Dai}, {and} \bibinfo{person}{Tianfu Wu}.} \bibinfo{year}{2020}\natexlab{}.
\newblock \showarticletitle{Stochastic-sign SGD for federated learning with theoretical guarantees}.
\newblock \bibinfo{journal}{\emph{arXiv preprint arXiv:2002.10940}} (\bibinfo{year}{2020}).
\newblock


\bibitem[Karimireddy et~al\mbox{.}(2020)]%
        {karimireddy2020scaffold}
\bibfield{author}{\bibinfo{person}{Sai~Praneeth Karimireddy}, \bibinfo{person}{Satyen Kale}, \bibinfo{person}{Mehryar Mohri}, \bibinfo{person}{Sashank Reddi}, \bibinfo{person}{Sebastian Stich}, {and} \bibinfo{person}{Ananda~Theertha Suresh}.} \bibinfo{year}{2020}\natexlab{}.
\newblock \showarticletitle{SCAFFOLD: Stochastic controlled averaging for federated learning}.
\newblock \bibinfo{journal}{\emph{Proceedings of ICML}} (\bibinfo{year}{2020}), \bibinfo{pages}{5132--5143}.
\newblock


\bibitem[Kingma and Ba(2014)]%
        {kingma2014adam}
\bibfield{author}{\bibinfo{person}{Diederik~P Kingma} {and} \bibinfo{person}{Jimmy Ba}.} \bibinfo{year}{2014}\natexlab{}.
\newblock \showarticletitle{Adam: A method for stochastic optimization}.
\newblock \bibinfo{journal}{\emph{arXiv preprint arXiv:1412.6980}} (\bibinfo{year}{2014}).
\newblock


\bibitem[Krizhevsky(2009)]%
        {Krizhevsky2009LearningML}
\bibfield{author}{\bibinfo{person}{Alex Krizhevsky}.} \bibinfo{year}{2009}\natexlab{}.
\newblock \showarticletitle{Learning Multiple Layers of Features from Tiny Images}.
\newblock
\urldef\tempurl%
\url{https://api.semanticscholar.org/CorpusID:18268744}
\showURL{%
\tempurl}


\bibitem[Kundroo and Kim(2023)]%
        {kundroo2023efficient}
\bibfield{author}{\bibinfo{person}{Majid Kundroo} {and} \bibinfo{person}{Taehong Kim}.} \bibinfo{year}{2023}\natexlab{}.
\newblock \showarticletitle{Efficient federated learning with adaptive client-side hyper-parameter optimization}. In \bibinfo{booktitle}{\emph{2023 IEEE 43rd international conference on distributed computing systems (ICDCS)}}. IEEE, \bibinfo{pages}{973--974}.
\newblock


\bibitem[Li et~al\mbox{.}(2023)]%
        {li2023memory}
\bibfield{author}{\bibinfo{person}{Bingrui Li}, \bibinfo{person}{Jianfei Chen}, {and} \bibinfo{person}{Jun Zhu}.} \bibinfo{year}{2023}\natexlab{}.
\newblock \showarticletitle{Memory efficient optimizers with 4-bit states}.
\newblock \bibinfo{journal}{\emph{Advances in Neural Information Processing Systems}}  \bibinfo{volume}{36} (\bibinfo{year}{2023}), \bibinfo{pages}{15136--15171}.
\newblock


\bibitem[Li and Wang(2019)]%
        {li2019fedmd}
\bibfield{author}{\bibinfo{person}{Daliang Li} {and} \bibinfo{person}{Junpu Wang}.} \bibinfo{year}{2019}\natexlab{}.
\newblock \showarticletitle{Fedmd: Heterogenous federated learning via model distillation}.
\newblock \bibinfo{journal}{\emph{arXiv preprint arXiv:1910.03581}} (\bibinfo{year}{2019}).
\newblock


\bibitem[Li et~al\mbox{.}(2022)]%
        {li2022federated}
\bibfield{author}{\bibinfo{person}{Qinbin Li}, \bibinfo{person}{Yiqun Diao}, \bibinfo{person}{Quan Chen}, {and} \bibinfo{person}{Bingsheng He}.} \bibinfo{year}{2022}\natexlab{}.
\newblock \showarticletitle{Federated learning on non-iid data silos: An experimental study}. In \bibinfo{booktitle}{\emph{2022 IEEE 38th international conference on data engineering (ICDE)}}. IEEE, \bibinfo{pages}{965--978}.
\newblock


\bibitem[Li et~al\mbox{.}(2020)]%
        {li2020federated}
\bibfield{author}{\bibinfo{person}{Tian Li}, \bibinfo{person}{Anit~Kumar Sahu}, \bibinfo{person}{Ameet Talwalkar}, {and} \bibinfo{person}{Virginia Smith}.} \bibinfo{year}{2020}\natexlab{}.
\newblock \showarticletitle{Federated learning: Challenges, methods, and future directions}.
\newblock \bibinfo{journal}{\emph{IEEE signal processing magazine}} \bibinfo{volume}{37}, \bibinfo{number}{3} (\bibinfo{year}{2020}), \bibinfo{pages}{50--60}.
\newblock


\bibitem[McMahan et~al\mbox{.}(2017)]%
        {mcmahan2017fedavg}
\bibfield{author}{\bibinfo{person}{Brendan McMahan}, \bibinfo{person}{Eider Moore}, \bibinfo{person}{Daniel Ramage}, \bibinfo{person}{Seth Hampson}, {and} \bibinfo{person}{Blaise~Aguera y Arcas}.} \bibinfo{year}{2017}\natexlab{}.
\newblock \showarticletitle{Communication-efficient learning of deep networks from decentralized data}.
\newblock \bibinfo{journal}{\emph{Proceedings of AISTATS}} (\bibinfo{year}{2017}).
\newblock


\bibitem[Micikevicius et~al\mbox{.}(2017)]%
        {micikevicius2017mixed}
\bibfield{author}{\bibinfo{person}{Paulius Micikevicius}, \bibinfo{person}{Sharan Narang}, \bibinfo{person}{Jonah Alben}, \bibinfo{person}{Gregory Diamos}, \bibinfo{person}{Erich Elsen}, \bibinfo{person}{David Garcia}, \bibinfo{person}{Boris Ginsburg}, \bibinfo{person}{Michael Houston}, \bibinfo{person}{Oleksii Kuchaiev}, \bibinfo{person}{Ganesh Venkatesh}, {et~al\mbox{.}}} \bibinfo{year}{2017}\natexlab{}.
\newblock \showarticletitle{Mixed precision training}.
\newblock \bibinfo{journal}{\emph{arXiv preprint arXiv:1710.03740}} (\bibinfo{year}{2017}).
\newblock


\bibitem[Mills et~al\mbox{.}(2021a)]%
        {mills2021communication}
\bibfield{author}{\bibinfo{person}{Jeffrey Mills}, \bibinfo{person}{Jia Hu}, {and} \bibinfo{person}{Geyong Min}.} \bibinfo{year}{2021}\natexlab{a}.
\newblock \showarticletitle{Communication-efficient federated learning via knowledge distillation}.
\newblock \bibinfo{journal}{\emph{Nature Communications}} \bibinfo{volume}{12}, \bibinfo{number}{1} (\bibinfo{year}{2021}), \bibinfo{pages}{2032}.
\newblock


\bibitem[Mills et~al\mbox{.}(2021b)]%
        {mills2021multi}
\bibfield{author}{\bibinfo{person}{Jed Mills}, \bibinfo{person}{Jia Hu}, {and} \bibinfo{person}{Geyong Min}.} \bibinfo{year}{2021}\natexlab{b}.
\newblock \showarticletitle{Multi-task federated learning for personalised deep neural networks in edge computing}.
\newblock \bibinfo{journal}{\emph{IEEE Transactions on Parallel and Distributed Systems}} \bibinfo{volume}{33}, \bibinfo{number}{3} (\bibinfo{year}{2021}), \bibinfo{pages}{630--641}.
\newblock


\bibitem[Mu and Shen(2025)]%
        {mu2025federated}
\bibfield{author}{\bibinfo{person}{Yujia Mu} {and} \bibinfo{person}{Cong Shen}.} \bibinfo{year}{2025}\natexlab{}.
\newblock \showarticletitle{Federated split learning with improved communication and storage efficiency}.
\newblock \bibinfo{journal}{\emph{IEEE Transactions on Mobile Computing}} (\bibinfo{year}{2025}).
\newblock


\bibitem[Oh et~al\mbox{.}(2022)]%
        {oh2022communication}
\bibfield{author}{\bibinfo{person}{Yongjeong Oh}, \bibinfo{person}{Namyoon Lee}, \bibinfo{person}{Yo-Seb Jeon}, {and} \bibinfo{person}{H~Vincent Poor}.} \bibinfo{year}{2022}\natexlab{}.
\newblock \showarticletitle{Communication-efficient federated learning via quantized compressed sensing}.
\newblock \bibinfo{journal}{\emph{IEEE Transactions on Wireless Communications}} \bibinfo{volume}{22}, \bibinfo{number}{2} (\bibinfo{year}{2022}), \bibinfo{pages}{1087--1100}.
\newblock


\bibitem[Pacheco et~al\mbox{.}(2024)]%
        {pacheco2024securing}
\bibfield{author}{\bibinfo{person}{Alexandre Pacheco}, \bibinfo{person}{S{\'e}bastien De~Vos}, \bibinfo{person}{Andreagiovanni Reina}, \bibinfo{person}{Marco Dorigo}, {and} \bibinfo{person}{Volker Strobel}.} \bibinfo{year}{2024}\natexlab{}.
\newblock \showarticletitle{Securing Federated Learning in Robot Swarms using Blockchain Technology}. In \bibinfo{booktitle}{\emph{International Symposium on Distributed Autonomous Robotic Systems}}. Springer, \bibinfo{pages}{473--488}.
\newblock


\bibitem[Pfitzner et~al\mbox{.}(2021)]%
        {pfitzner2021federated}
\bibfield{author}{\bibinfo{person}{Bjarne Pfitzner}, \bibinfo{person}{Nico Steckhan}, {and} \bibinfo{person}{Bert Arnrich}.} \bibinfo{year}{2021}\natexlab{}.
\newblock \showarticletitle{Federated learning in a medical context: a systematic literature review}.
\newblock \bibinfo{journal}{\emph{ACM Transactions on Internet Technology (TOIT)}} \bibinfo{volume}{21}, \bibinfo{number}{2} (\bibinfo{year}{2021}), \bibinfo{pages}{1--31}.
\newblock


\bibitem[Qu et~al\mbox{.}(2022)]%
        {qu2022blockchain}
\bibfield{author}{\bibinfo{person}{Youyang Qu}, \bibinfo{person}{Md~Palash Uddin}, \bibinfo{person}{Chenquan Gan}, \bibinfo{person}{Yong Xiang}, \bibinfo{person}{Longxiang Gao}, {and} \bibinfo{person}{John Yearwood}.} \bibinfo{year}{2022}\natexlab{}.
\newblock \showarticletitle{Blockchain-enabled federated learning: A survey}.
\newblock \bibinfo{journal}{\emph{Comput. Surveys}} \bibinfo{volume}{55}, \bibinfo{number}{4} (\bibinfo{year}{2022}), \bibinfo{pages}{1--35}.
\newblock


\bibitem[{Raspberry Pi Ltd.}(2026)]%
        {raspberrypi_zero}
\bibfield{author}{\bibinfo{person}{{Raspberry Pi Ltd.}}} \bibinfo{year}{2026}\natexlab{}.
\newblock \bibinfo{title}{Zero models}.
\newblock \bibinfo{howpublished}{Raspberry Pi Documentation}.
\newblock
\urldef\tempurl%
\url{https://www.raspberrypi.com/documentation/computers/zero.html}
\showURL{%
\tempurl}
\newblock
\shownote{Accessed: 2026-10-31}.


\bibitem[Reddi et~al\mbox{.}(2021)]%
        {reddi2021adaptive}
\bibfield{author}{\bibinfo{person}{Sashank Reddi}, \bibinfo{person}{Zachary Charles}, \bibinfo{person}{Manzil Zaheer}, \bibinfo{person}{Zachary Garrett}, \bibinfo{person}{Keith Rush}, \bibinfo{person}{Jakub Kone{\v{c}}n{\`y}}, \bibinfo{person}{Sanjiv Kumar}, {and} \bibinfo{person}{H~Brendan McMahan}.} \bibinfo{year}{2021}\natexlab{}.
\newblock \showarticletitle{Adaptive federated optimization}.
\newblock \bibinfo{journal}{\emph{Proceedings of ICLR}} (\bibinfo{year}{2021}).
\newblock


\bibitem[Reisizadeh et~al\mbox{.}(2020)]%
        {reisizadeh2020fedpaq}
\bibfield{author}{\bibinfo{person}{Amirhossein Reisizadeh}, \bibinfo{person}{Aryan Mokhtari}, \bibinfo{person}{Hamed Hassani}, \bibinfo{person}{Ali Jadbabaie}, {and} \bibinfo{person}{Ramtin Pedarsani}.} \bibinfo{year}{2020}\natexlab{}.
\newblock \showarticletitle{FedPAQ: A communication-efficient federated learning method with periodic averaging and quantization}. In \bibinfo{booktitle}{\emph{International Conference on Artificial Intelligence and Statistics}}. PMLR, \bibinfo{pages}{2021--2031}.
\newblock


\bibitem[Shafiq and Gu(2022)]%
        {shafiq2022deep}
\bibfield{author}{\bibinfo{person}{Muhammad Shafiq} {and} \bibinfo{person}{Zhaoquan Gu}.} \bibinfo{year}{2022}\natexlab{}.
\newblock \showarticletitle{Deep residual learning for image recognition: A survey}.
\newblock \bibinfo{journal}{\emph{Applied sciences}} \bibinfo{volume}{12}, \bibinfo{number}{18} (\bibinfo{year}{2022}), \bibinfo{pages}{8972}.
\newblock


\bibitem[Sharir et~al\mbox{.}(2021)]%
        {sharir2021image}
\bibfield{author}{\bibinfo{person}{Gilad Sharir}, \bibinfo{person}{Asaf Noy}, {and} \bibinfo{person}{Lihi Zelnik-Manor}.} \bibinfo{year}{2021}\natexlab{}.
\newblock \showarticletitle{An image is worth 16x16 words, what is a video worth?}
\newblock \bibinfo{journal}{\emph{arXiv preprint arXiv:2103.13915}} (\bibinfo{year}{2021}).
\newblock


\bibitem[Tamirisa et~al\mbox{.}(2024)]%
        {tamirisa2024fedselect}
\bibfield{author}{\bibinfo{person}{Rishub Tamirisa}, \bibinfo{person}{Chulin Xie}, \bibinfo{person}{Wenxuan Bao}, \bibinfo{person}{Andy Zhou}, \bibinfo{person}{Ron Arel}, {and} \bibinfo{person}{Aviv Shamsian}.} \bibinfo{year}{2024}\natexlab{}.
\newblock \showarticletitle{Fedselect: Personalized federated learning with customized selection of parameters for fine-tuning}. In \bibinfo{booktitle}{\emph{Proceedings of the IEEE/CVF Conference on Computer Vision and Pattern Recognition}}. \bibinfo{pages}{23985--23994}.
\newblock


\bibitem[Tang et~al\mbox{.}(2024)]%
        {tang2024fedcada}
\bibfield{author}{\bibinfo{person}{Haibo Tang}, \bibinfo{person}{Junyi Yang}, \bibinfo{person}{Sheng Zhou}, \bibinfo{person}{Yuanming Shi}, {and} \bibinfo{person}{Zhisheng Niu}.} \bibinfo{year}{2024}\natexlab{}.
\newblock \showarticletitle{FedCAda: Adaptive Client-Side Optimization for Accelerated and Stable Federated Learning}. In \bibinfo{booktitle}{\emph{IEEE International Conference on Communications}}.
\newblock


\bibitem[Tao et~al\mbox{.}(2023)]%
        {tao2023preconditioned}
\bibfield{author}{\bibinfo{person}{Zeyi Tao}, \bibinfo{person}{Jindi Wu}, {and} \bibinfo{person}{Qun Li}.} \bibinfo{year}{2023}\natexlab{}.
\newblock \showarticletitle{Preconditioned Federated Learning}.
\newblock \bibinfo{journal}{\emph{arXiv preprint arXiv:2309.11378}} (\bibinfo{year}{2023}).
\newblock


\bibitem[Thapa et~al\mbox{.}(2022)]%
        {thapa2022splitfed}
\bibfield{author}{\bibinfo{person}{Chandra Thapa}, \bibinfo{person}{Pathum Chamikara~Mahawaga Arachchige}, \bibinfo{person}{Seyit Camtepe}, {and} \bibinfo{person}{Lichao Sun}.} \bibinfo{year}{2022}\natexlab{}.
\newblock \showarticletitle{Splitfed: When federated learning meets split learning}. In \bibinfo{booktitle}{\emph{Proceedings of the AAAI conference on artificial intelligence}}, Vol.~\bibinfo{volume}{36}. \bibinfo{pages}{8485--8493}.
\newblock


\bibitem[Vepakomma et~al\mbox{.}(2018)]%
        {vepakomma2018split}
\bibfield{author}{\bibinfo{person}{Praneeth Vepakomma}, \bibinfo{person}{Otkrist Gupta}, \bibinfo{person}{Tristan Swedish}, {and} \bibinfo{person}{Ramesh Raskar}.} \bibinfo{year}{2018}\natexlab{}.
\newblock \showarticletitle{Split learning for health: Distributed deep learning without sharing raw patient data}.
\newblock \bibinfo{journal}{\emph{arXiv preprint arXiv:1812.00564}} (\bibinfo{year}{2018}).
\newblock


\bibitem[Wang et~al\mbox{.}(2018)]%
        {wang2018training}
\bibfield{author}{\bibinfo{person}{Naigang Wang}, \bibinfo{person}{Jungwook Choi}, \bibinfo{person}{Daniel Brand}, \bibinfo{person}{Chia-Yu Chen}, {and} \bibinfo{person}{Kailash Gopalakrishnan}.} \bibinfo{year}{2018}\natexlab{}.
\newblock \showarticletitle{Training deep neural networks with 8-bit floating point numbers}.
\newblock \bibinfo{journal}{\emph{Advances in neural information processing systems}}  \bibinfo{volume}{31} (\bibinfo{year}{2018}).
\newblock


\bibitem[Zhang et~al\mbox{.}(2024)]%
        {zhang2024fedsl}
\bibfield{author}{\bibinfo{person}{Weishan Zhang}, \bibinfo{person}{Tao Zhou}, \bibinfo{person}{Qinghua Lu}, \bibinfo{person}{Yong Yuan}, \bibinfo{person}{Amr Tolba}, {and} \bibinfo{person}{Wael Said}.} \bibinfo{year}{2024}\natexlab{}.
\newblock \showarticletitle{FedSL: A communication-efficient federated learning with split layer aggregation}.
\newblock \bibinfo{journal}{\emph{IEEE Internet of Things Journal}} \bibinfo{volume}{11}, \bibinfo{number}{9} (\bibinfo{year}{2024}), \bibinfo{pages}{15587--15601}.
\newblock


\bibitem[Zhang et~al\mbox{.}(2015)]%
        {zhang2015character}
\bibfield{author}{\bibinfo{person}{Xiang Zhang}, \bibinfo{person}{Junbo Zhao}, {and} \bibinfo{person}{Yann LeCun}.} \bibinfo{year}{2015}\natexlab{}.
\newblock \showarticletitle{Character-level convolutional networks for text classification}.
\newblock \bibinfo{journal}{\emph{Advances in neural information processing systems}}  \bibinfo{volume}{28} (\bibinfo{year}{2015}).
\newblock


\bibitem[Zhang et~al\mbox{.}(2025)]%
        {zhang2025fedhq}
\bibfield{author}{\bibinfo{person}{Zhuochen Zhang}, \bibinfo{person}{Xuefei Liu}, \bibinfo{person}{Yuanchun Wang}, {and} \bibinfo{person}{Yunxin Liu}.} \bibinfo{year}{2025}\natexlab{}.
\newblock \showarticletitle{FedHQ: Hybrid Runtime Quantization for Federated Learning}.
\newblock \bibinfo{journal}{\emph{arXiv preprint arXiv:2505.11982}} (\bibinfo{year}{2025}).
\newblock


\end{thebibliography}
%==========================================
% APPENDIX
% ==========================================
\clearpage
\appendix

%\begin{comment}

\section{Appendix}
\label{appendix}

\subsection{Federated Learning and FedAdam Background}
\label{app:fedadam}

We consider the standard federated learning setup with $K$ clients and a central server. Each client $k$ holds a local dataset $\mathcal{D}_k$, and the goal is to minimize the global loss:
\[
\min_{\theta} f(\theta) = \sum_{k=1}^{K} p_k f_k(\theta),
\]
where $f_k(\theta) = \mathbb{E}_{(x,y) \sim \mathcal{D}_k}[L(\theta; x, y)]$ and $p_k = |\mathcal{D}_k| / \sum_j |\mathcal{D}_j|$.

FedAdam~\cite{reddi2021adaptive} maintains server-side momentum $m$ and variance $v$ estimates, updated via:
\begin{align}
\Delta_t &= \frac{1}{K} \sum_{k=1}^{K} (\theta_t^k - \theta_t), \\
m_t &= \beta_1 m_{t-1} + (1 - \beta_1) \Delta_t, \\
v_t &= \beta_2 v_{t-1} + (1 - \beta_2) \Delta_t^2, \\
\theta_{t+1} &= \theta_t - \eta \frac{m_t}{\sqrt{v_t} + \epsilon},
\end{align}
where $\beta_1, \beta_2$ are momentum parameters (typically $0.9, 0.999$), $\eta$ is the learning rate, and $\epsilon$ is a small constant for numerical stability.

For a model with $N$ parameters, storing $m$ and $v$ in FP32 requires $2N \times 4$ bytes, often comparable to the model itself.

\textbf{Client-side adaptation.}
In this work, we apply the Adam update rule \emph{on clients} during local training, rather than maintaining server-side optimizer states as in the original FedAdam formulation. Each client maintains momentum $m_k$ and variance $v_k$ during its local epochs, but these states are reinitialized at the start of each federated round. The server performs only weighted averaging (FedAvg) of client model parameters. This design choice targets client memory constraints the primary bottleneck in resource-limited edge devices while maintaining the adaptive optimization benefits of FedAdam during local training.

\subsection{Empirical Analysis of Optimizer State Distributions}
\label{app:optimizer_dist}

To motivate our quantization design, we characterize the statistical properties of Vanilla-ClientAdam optimizer states. We train ResNet-18 variants on CIFAR-10 (6.3M parameters) and CIFAR-100 (11M parameters) for 50 gradient steps with batch size 64, then extract all momentum and variance values across all layers.

\begin{figure}[t]
    \centering
    \includegraphics[width=\columnwidth]{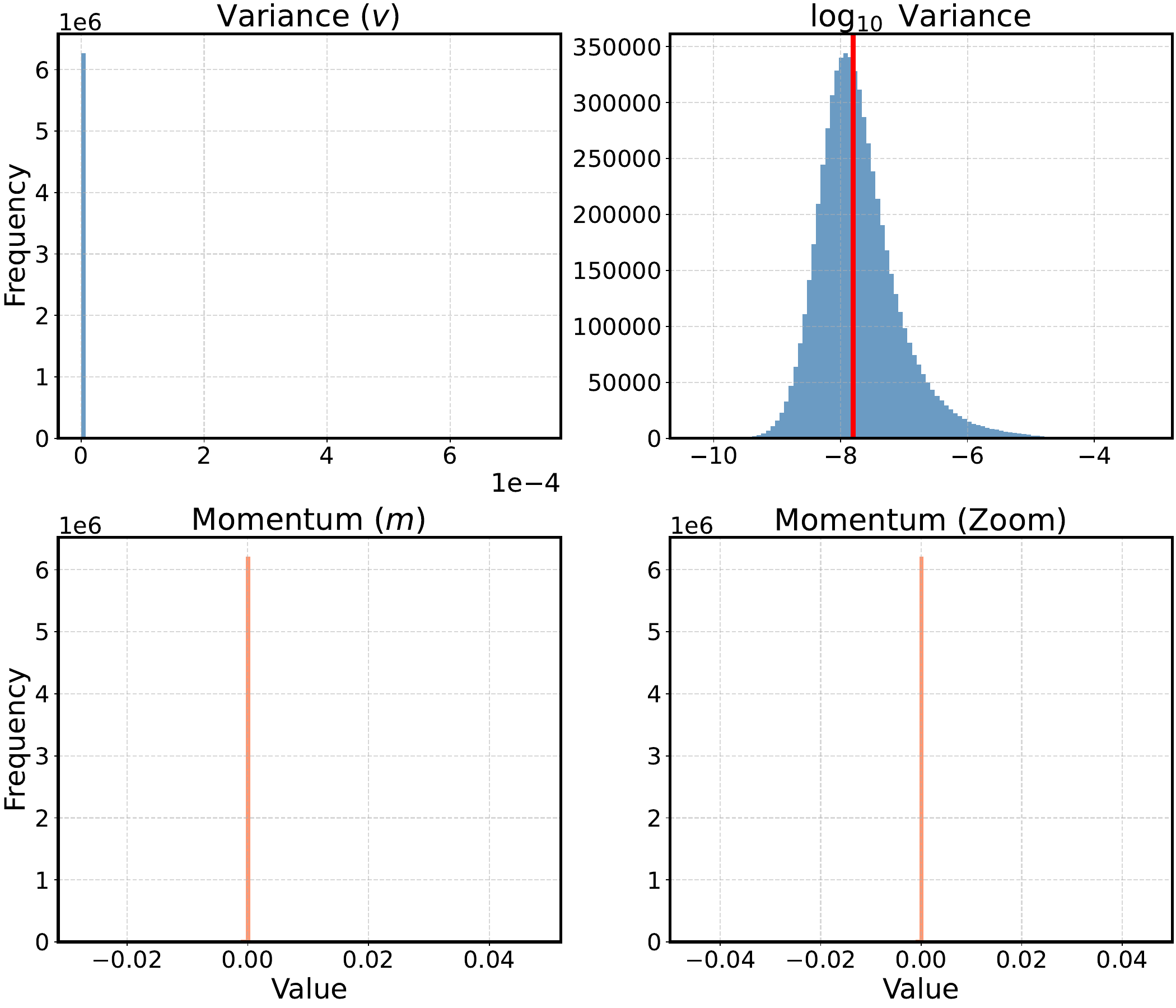}
    \caption{CIFAR-10 optimizer state distributions after 50 training steps.
    \textbf{Top:} Variance ($v$) on linear (left) and log$_{10}$ (right) scales.
    \textbf{Bottom:} Momentum ($m$) on full scale (left).}
    Four-panel histogram showing optimizer state distributions. Top panels show variance with linear and log scales. Bottom panels show momentum distributions with full and zoomed views.
    \label{fig:optimizer_dist_c10}
\end{figure}

\begin{figure}[t]
    \centering
    \includegraphics[width=\columnwidth]{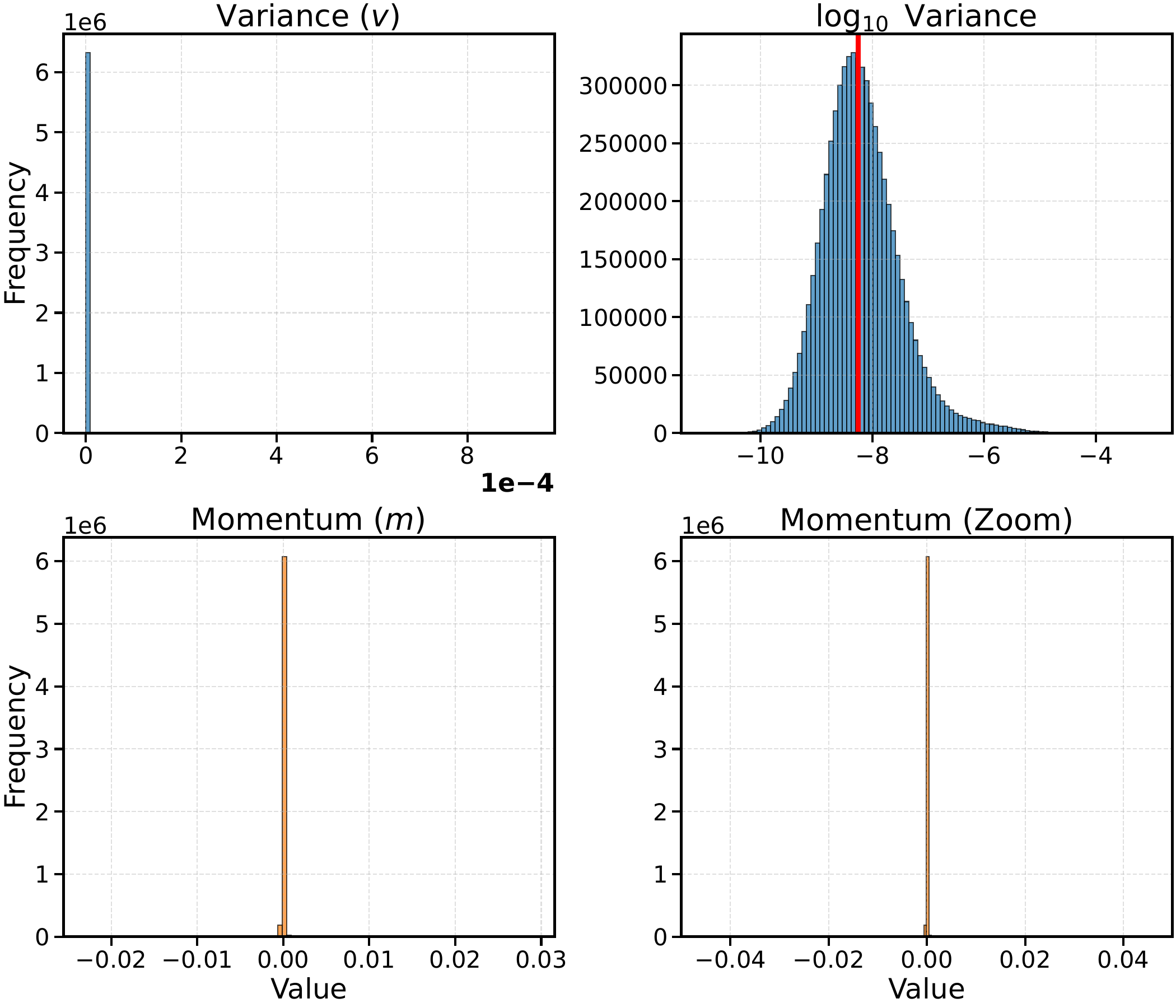}
    \caption{CIFAR-100 optimizer state distributions after 50 training steps.
    Layout identical to Figure~\ref{fig:optimizer_dist_c10}.
    The asymmetry between variance (7--8 orders of magnitude) and momentum ($\pm 0.1$) is consistent across both datasets.}
    Four-panel histogram showing CIFAR-100 optimizer state distributions with same layout as Figure \ref{fig:optimizer_dist_c10}.
    \label{fig:optimizer_dist_c100}
\end{figure}

Figures \ref{fig:optimizer_dist_c10} and \ref{fig:optimizer_dist_c100} show the distributions for CIFAR-10 and CIFAR-100, respectively. On linear scale (top left in each figure), variance distributions are extremely right-skewed, with most values concentrated near zero and a long tail extending to large values. On log scale (top right), the distributions appear approximately Gaussian, spanning roughly 7--8 orders of magnitude from $10^{-12}$ to $10^{-3}$. The minimum-to-maximum ratio is approximately $10^7$ for both datasets, meaning a uniform linear 8-bit quantizer with 256 levels would allocate most of its dynamic range to the sparse tail, crushing small values into a single quantization bin and losing critical precision.

In contrast, momentum values (bottom row) remain within a moderate range and exhibit approximately symmetric, bounded behavior suitable for linear quantization. For both datasets, momentum values lie almost entirely within $[-0.1, +0.1]$, with 95\% of values concentrated within $\pm 0.05$ (bottom right panels show zoomed view). The distributions are approximately Gaussian and centered near zero, indicating that momentum has no skewness or heavy tails.

This stark difference in statistical structure motivates our asymmetric design principle: \emph{optimizer states with fundamentally different distributions should not share the same quantization scheme}. We apply linear quantization to momentum (symmetric, bounded) and log-space quantization to variance (log-normal, spanning orders of magnitude). This design is consistent across both CIFAR-10 and CIFAR-100, despite their different model sizes and class counts.

\subsection{Quantization Precision Analysis}
\label{app:quant_analysis}

A critical challenge in federated learning optimization is the compression of the optimizer's second moment (variance), which typically exhibits a heavy-tailed distribution spanning many orders of magnitude (e.g., $10^{-7}$ to $10^{1}$). To validate our choice of log-space quantization, we compared its representational fidelity against a prefix-based dynamic quantization baseline.

\subsubsection{Simulation Setup}
We generated a synthetic dataset of $N=3000$ values following a log-uniform distribution to mimic the skewed variance states observed in non-IID federated training. The data range was set from $10^{-7}$ to $10^{0}$. We compared two 8-bit quantization schemes:
\begin{itemize}
    \item \textbf{Log-space Quantization (Ours):} Direct mapping of values to a logarithmic grid.
    \item \textbf{Prefix-Based Dynamic Tree (Dettmers):} We simulate a prefix-based dynamic tree quantization scheme that captures the key characteristic of Dettmers-style methods~\cite{dettmers2021bit} namely, magnitude-dependent mantissa allocation due to unary exponent prefixes.
\end{itemize}

\subsubsection{Results and Discussion}
\begin{figure}[t]
    \centering
    \includegraphics[width=\columnwidth]{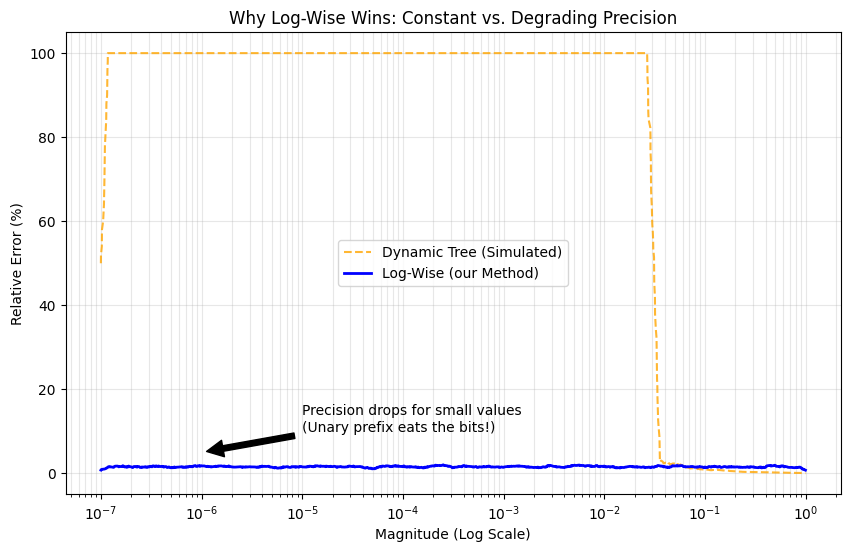}
    \caption{Comparison of relative quantization error between log-space quantization (ours) and a simulated prefix-based dynamic tree baseline. The baseline suffers from bit starvation at small magnitudes ($<10^{-2}$), while our method maintains constant relative error across the full dynamic range.}
    \label{fig:quant_error}
\end{figure}

The empirical results, visualized in Figure \ref{fig:quant_error}, demonstrate a fundamental trade-off in quantization structure. For variance tensors with heavy-tailed distributions, log-space quantization consistently yields lower relative error across magnitudes compared to prefix-based dynamic tree schemes, which suffer from bit starvation at small values.

\begin{itemize}
    \item \textbf{Log-space (Ours):} $1.58\%$
    \item \textbf{Dynamic Tree Baseline:} $78.23\%$
\end{itemize}

A prefix-based encoding forces a trade-off: to represent small numbers, the encoding must use a long unary prefix, drastically reducing the bits available for the mantissa. Our log-space approach decouples scale from resolution, ensuring that relative precision remains invariant to signal magnitude.

\section{Data Distribution Details}
\label{app:data_distribution}

\subsection{Dirichlet Partitioning Methodology}
We employ Dirichlet distribution to simulate realistic non-IID data partitions in federated learning. For a dataset with $K$ classes and $N$ clients, samples of class $k$ are allocated to clients according to proportions $\mathbf{p}_k \sim \text{Dir}(\alpha, \ldots, \alpha)$, where $\alpha$ is the concentration parameter. Lower $\alpha$ values produce more skewed distributions: $\alpha \to 0$ results in each client receiving samples from only one class (complete pathological non-IID), while $\alpha \to \infty$ approaches uniform IID distribution. This partitioning strategy is widely adopted in federated learning literature~\cite{li2022federated,chen2020fedbe} as it captures real-world data heterogeneity where different clients naturally collect data from non-uniform distributions.

\subsection{Heterogeneity Quantification}
Table \ref{tab:data_distribution_summary} summarizes the degree of heterogeneity across different $\alpha$ values using two key metrics:

\textbf{Average Dominant Class Percentage:} Measures data concentration by computing the fraction of samples belonging to the most frequent class in each client, averaged across all clients. For CIFAR-10, this metric ranges from 10.6\% (IID, uniform across 10 classes) to 65.7\% ($\alpha=0.1$, extreme skew). Under $\alpha=0.1$, some clients exhibit over 90\% concentration in a single class (e.g., Client 5 with 96.7\% ships, Client 7 with 93.6\% automobiles), simulating scenarios where edge devices collect highly specialized data analogous to a smartphone primarily photographing pets or a hospital system predominantly imaging specific pathologies. For CIFAR-100, the same $\alpha$ value produces lower percentages (10.4\% at $\alpha=0.1$) due to class dilution across 100 categories, but relative heterogeneity remains significant. We classify heterogeneity based on the ratio to uniform baseline: CIFAR-100's 10.4\% (10$\times$ baseline) is labeled ``Extreme'' similar to CIFAR-10's 65.7\% (6.5$\times$ baseline), as both represent severe class imbalance relative to their respective uniform distributions.

\textbf{Sample Standard Deviation:} Quantifies imbalance in dataset sizes across clients. CIFAR-10 with $\alpha=0.1$ shows extreme variance (std=3,204), with clients holding between 1,040 and 9,826 samples nearly 9.5$\times$ difference. This asymmetry mirrors real federated scenarios where clients have vastly different data volumes (e.g., active vs. inactive users). IID partitioning yields zero variance with exactly 5,000 samples per client, serving as our controlled baseline.

\begin{table}[t]
\centering
\caption{Data heterogeneity summary across clients. \textbf{Ratio} indicates dominant class concentration relative to a uniform distribution ($1/K$).}
\label{tab:data_distribution_summary}

\small
\setlength{\tabcolsep}{3pt}
\renewcommand{\arraystretch}{1.1}

\begin{tabular}{lccccc}
\toprule
\textbf{Dataset} & \textbf{$\alpha$} & \textbf{Dom.\%} & \textbf{Ratio} & \textbf{Std.} & \textbf{Level} \\
\midrule
\multirow{4}{*}{CIFAR-10}
& 0.1 & 65.7\% & 6.6$\times$ & 3204 & Extreme \\
& 0.5 & 42.6\% & 4.3$\times$ & 2101 & High \\
& 1.0 & 26.8\% & 2.7$\times$ & 936  & Moderate \\
& IID & 10.6\% & 1.1$\times$ & 0    & Balanced \\
\midrule
\multirow{4}{*}{CIFAR-100}
& 0.1 & 10.4\% & 10.4$\times$ & 1309 & Extreme \\
& 0.5 & 5.5\%  & 5.5$\times$  & 410  & High \\
& 1.0 & 4.5\%  & 4.5$\times$  & 455$^{*}$ & Moderate \\
& IID & 1.4\%  & 1.4$\times$  & 0    & Balanced \\
\bottomrule
\end{tabular}

\medskip
\footnotesize
\noindent\textit{$^{*}$Sample standard deviations do not necessarily decrease monotonically with $\alpha$ because Dirichlet partitioning is stochastic. With a finite number of clients, small fluctuations (e.g., 455 vs.\ 410) are expected.}

\end{table}

\begin{table}[t]
\centering
\caption{Data heterogeneity under varying Dirichlet concentration parameters ($\alpha$). Avg.\ Dom.\% denotes the average proportion of the dominant class per client. Lower $\alpha$ corresponds to stronger statistical heterogeneity. Sample standard deviations reflect stochastic Dirichlet sampling and are therefore not strictly monotonic across $\alpha$.}
\label{tab:data_distribution}
\resizebox{\columnwidth}{!}{%
\begin{tabular}{lcccc}
\toprule
\textbf{Dataset} & \textbf{$\alpha$} & \textbf{Avg. Dom.\%} & \textbf{Sample Std.} & \textbf{Level} \\
\midrule
CIFAR-10  & 0.1 & 65.7 & 3,204 & Extreme \\
          & 0.5 & 42.6 & 2,101 & High \\
          & 1.0 & 26.8 & 936   & Moderate \\
          & IID & 10.6 & 0     & Balanced \\
\midrule
CIFAR-100 & 0.1 & 10.4 & 1,309 & Extreme \\
          & 0.5 & 5.5  & 410   & High \\
          & 1.0 & 4.5  & 455   & Moderate \\
          & IID & 1.4  & 0     & Balanced \\
\bottomrule
\end{tabular}
}
\end{table}

\subsection{Per-Client Distribution Analysis}
Tables \ref{tab:cifar10_detailed} and \ref{tab:cifar100_detailed} provide exhaustive per-client breakdowns for all $\alpha$ configurations. Key observations:

\paragraph{CIFAR-10 Extreme Non-IID ($\alpha=0.1$):} Client specialization is highly pronounced. Client 5 becomes a near-specialist in ships (96.7\% of its 4,481 samples), while Client 7 focuses on automobiles (93.6\% of 1,590 samples). Clients 3 and 4 both specialize in airplanes but with vastly different data volumes (1,749 vs. 4,467 samples, exhibiting 85.5\% and 77.9\% concentration respectively). Such extreme partitions test whether optimizers can converge when local client objectives diverge significantly from the global optimum a critical challenge in federated learning where local gradients may point in conflicting directions.

\paragraph{CIFAR-100 Extreme Non-IID ($\alpha=0.1$):} Despite 100 classes, heterogeneity manifests differently than CIFAR-10. Dominant class percentages appear lower (avg 10.4\%) due to class dilution, but \emph{relative} concentration is equally extreme 10.4\% represents 10$\times$ oversampling compared to the 1\% uniform baseline, comparable in severity to CIFAR-10's 65.7\% (6.5$\times$ the 10\% baseline).
Client 4 exhibits 18.4\% in a single class (class 29), representing 18$\times$ oversampling compared to the uniform 1\% baseline. This demonstrates that heterogeneity persists even in fine-grained classification tasks, though its impact on optimizer dynamics differs from coarse-grained scenarios.

\paragraph{Sample Imbalance Impact:} At $\alpha=0.1$, small clients (e.g., CIFAR-10 Client 9 with 1,040 samples) perform significantly fewer local SGD updates per round than large clients (e.g., Client 8 with 9,826 samples). With 2 local epochs and batch size 64, Client 9 performs $\approx$33 gradient steps per round while Client 8 performs $\approx$307 steps nearly 9.5$\times$ difference. This asymmetry introduces optimization challenges: quantized optimizers must maintain stable momentum and variance estimates despite drastically different update frequencies across clients.

\paragraph{Client Sampling Statistics:} Over 120 federated rounds with 5 clients randomly selected per round (without replacement), selection frequency remains approximately balanced: mean 60.0 selections per client with standard deviation 6.3 (ranging from 50 to 74 across clients). This ensures all clients contribute equally to global model updates over time, regardless of their local data size or heterogeneity level. Random sampling prevents systematic bias while maintaining statistical efficiency.

% CIFAR-10 Detailed Table
\begin{table}[htbp]
\centering
\caption{Detailed per-client data distribution for CIFAR-10 across all $\alpha$ values. Each row shows a client's total sample count and dominant class statistics. Class labels: 0=Airplane, 1=Automobile, 2=Bird, 3=Cat, 4=Deer, 5=Dog, 6=Frog, 7=Horse, 8=Ship, 9=Truck.}
\label{tab:cifar10_detailed}
\fontsize{9}{11}\selectfont
\setlength{\tabcolsep}{3pt} % Slightly reduces space between columns to ensure it fits
\begin{tabular}{ccrrc}
\toprule
\textbf{$\alpha$} & \textbf{Client} & \textbf{Samples} & \textbf{Dominant Class} & \textbf{Dominant \%} \\
\midrule
\multirow{10}{*}{0.1}
& C0 & 9,348 & 3 (Cat)        & 37.3\% \\
& C1 & 8,295 & 4 (Deer)       & 54.9\% \\
& C2 & 4,053 & 2 (Bird)       & 62.6\% \\
& C3 & 1,749 & 0 (Airplane)   & 85.5\% \\
& C4 & 4,467 & 0 (Airplane)   & 77.9\% \\
& C5 & 4,481 & 8 (Ship)       & 96.7\% \\
& C6 & 5,120 & 5 (Dog)        & 60.6\% \\
& C7 & 1,590 & 1 (Automobile) & 93.6\% \\
& C8 & 9,826 & 1 (Automobile) & 30.9\% \\
& C9 & 1,040 & 5 (Dog)        & 57.2\% \\
\midrule
\multirow{10}{*}{0.5}
& C0 & 3,597 & 9 (Truck)      & 40.6\% \\
& C1 & 5,700 & 4 (Deer)       & 53.7\% \\
& C2 & 2,874 & 1 (Automobile) & 38.7\% \\
& C3 & 2,850 & 0 (Airplane)   & 44.4\% \\
& C4 & 6,509 & 0 (Airplane)   & 33.0\% \\
& C5 & 4,079 & 3 (Cat)        & 58.9\% \\
& C6 & 7,929 & 7 (Horse)      & 42.9\% \\
& C7 & 4,040 & 1 (Automobile) & 57.0\% \\
& C8 & 3,663 & 6 (Frog)       & 30.5\% \\
& C9 & 8,708 & 8 (Ship)       & 25.9\% \\
\midrule
\multirow{10}{*}{1.0}
& C0 & 4,401 & 5 (Dog)        & 34.1\% \\
& C1 & 5,072 & 7 (Horse)      & 25.5\% \\
& C2 & 6,755 & 4 (Deer)       & 31.9\% \\
& C3 & 4,019 & 8 (Ship)       & 24.3\% \\
& C4 & 4,482 & 1 (Automobile) & 25.6\% \\
& C5 & 6,003 & 6 (Frog)       & 26.7\% \\
& C6 & 5,474 & 9 (Truck)      & 21.7\% \\
& C7 & 3,783 & 0 (Airplane)   & 23.7\% \\
& C8 & 4,481 & 0 (Airplane)   & 29.1\% \\
& C9 & 5,483 & 3 (Cat)        & 25.8\% \\
\midrule
\multirow{10}{*}{IID}
& C0 & 5,000 & 2 (Bird)       & 10.5\% \\
& C1 & 5,000 & 9 (Truck)      & 10.8\% \\
& C2 & 5,000 & 5 (Dog)        & 10.6\% \\
& C3 & 5,000 & 9 (Truck)      & 10.4\% \\
& C4 & 5,000 & 1 (Automobile) & 10.6\% \\
& C5 & 5,000 & 4 (Deer)       & 10.6\% \\
& C6 & 5,000 & 6 (Frog)       & 10.6\% \\
& C7 & 5,000 & 4 (Deer)       & 10.7\% \\
& C8 & 5,000 & 6 (Frog)       & 10.6\% \\
& C9 & 5,000 & 7 (Horse)      & 10.7\% \\
\bottomrule
\end{tabular}
\end{table}

% CIFAR-100 Detailed Table
\begin{table}[htbp]
\centering
\caption{Detailed per-client data distribution for CIFAR-100 across all $\alpha$ values. With 100 classes, dominant class percentages are naturally lower than CIFAR-10, yet relative heterogeneity patterns remain consistent. Only dominant class IDs are shown.}
\label{tab:cifar100_detailed}
\fontsize{9}{11}\selectfont
\setlength{\tabcolsep}{3pt}
\begin{tabular}{ccrrc}
\toprule
\textbf{$\alpha$} & \textbf{Client} & \textbf{Samples} & \textbf{Dominant Class} & \textbf{Dominant \%} \\
\midrule
\multirow{10}{*}{0.1}
& C0 & 4,151 & 71 & 11.4\% \\
& C1 & 4,562 & 86 & 10.1\% \\
& C2 & 6,051 & 9  & 8.2\% \\
& C3 & 5,384 & 4  & 9.2\% \\
& C4 & 2,653 & 29 & 18.4\% \\
& C5 & 5,819 & 34 & 8.3\% \\
& C6 & 5,003 & 6  & 9.7\% \\
& C7 & 6,424 & 98 & 7.7\% \\
& C8 & 6,425 & 94 & 7.3\% \\
& C9 & 3,255 & 77 & 13.5\% \\
\midrule
\multirow{10}{*}{0.5}
& C0 & 5,309 & 84 & 6.1\% \\
& C1 & 4,680 & 65 & 5.7\% \\
& C2 & 5,158 & 85 & 5.5\% \\
& C3 & 4,047 & 8  & 6.8\% \\
& C4 & 4,663 & 12 & 5.2\% \\
& C5 & 5,107 & 55 & 5.2\% \\
& C6 & 5,122 & 2  & 4.7\% \\
& C7 & 4,875 & 25 & 6.5\% \\
& C8 & 5,049 & 93 & 4.6\% \\
& C9 & 5,503 & 37 & 4.8\% \\
\midrule
\multirow{10}{*}{1.0}
& C0 & 5,899 & 71 & 3.7\% \\
& C1 & 4,658 & 90 & 4.4\% \\
& C2 & 4,636 & 79 & 3.4\% \\
& C3 & 4,685 & 11 & 4.4\% \\
& C4 & 4,334 & 31 & 5.3\% \\
& C5 & 4,906 & 17 & 5.7\% \\
& C6 & 5,267 & 18 & 6.5\% \\
& C7 & 5,336 & 64 & 3.7\% \\
& C8 & 5,079 & 57 & 2.9\% \\
& C9 & 4,701 & 92 & 5.5\% \\
\midrule
\multirow{10}{*}{IID}
& C0 & 5,000 & 18 & 1.3\% \\
& C1 & 5,000 & 42 & 1.3\% \\
& C2 & 5,000 & 60 & 1.3\% \\
& C3 & 5,000 & 10 & 1.6\% \\
& C4 & 5,000 & 19 & 1.4\% \\
& C5 & 5,000 & 34 & 1.3\% \\
& C6 & 5,000 & 48 & 1.2\% \\
& C7 & 5,000 & 96 & 1.3\% \\
& C8 & 5,000 & 13 & 1.3\% \\
& C9 & 5,000 & 64 & 1.5\% \\
\bottomrule
\end{tabular}
\end{table}

\subsection{Visual Analysis}
\begin{figure}[!htbp]   
    \centering
    \includegraphics[width=0.9\columnwidth, height=17cm, keepaspectratio]{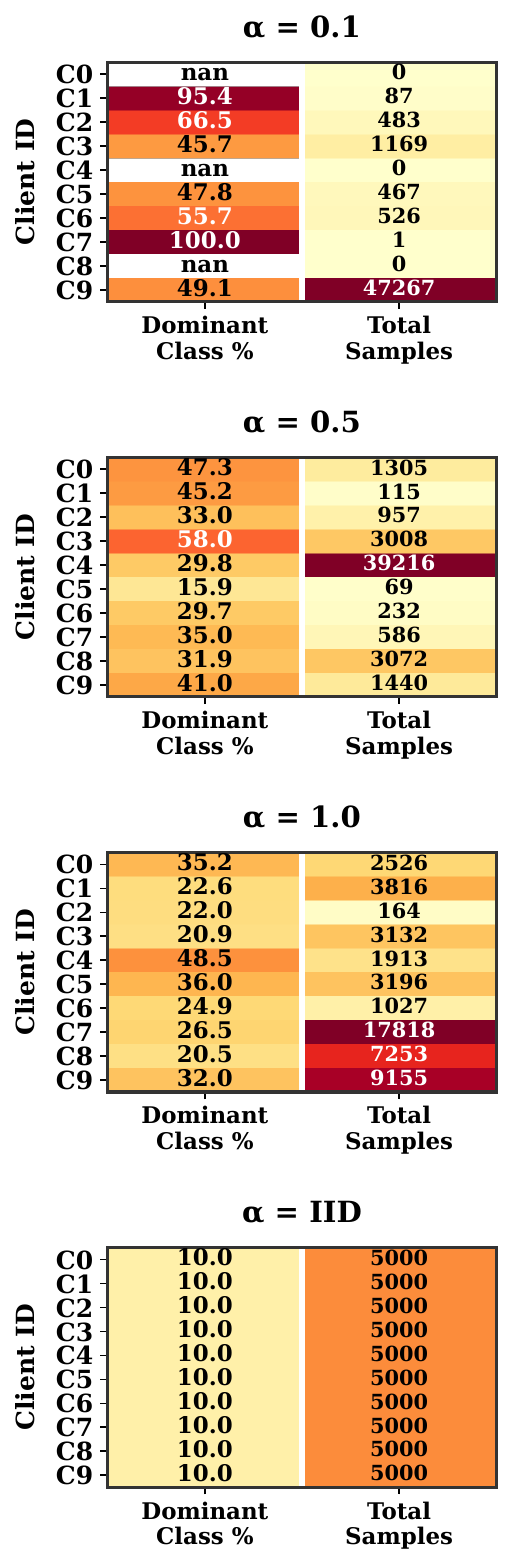}
    \caption{CIFAR-10 client distribution heatmaps for varying $\alpha$. Left: dominant class \% ; Right: sample count/100 (darker = higher). At $\alpha=0.1$, Clients 5 and 7 exceed 90\% single-class concentration (96.7\%, 93.6\%).}
    \label{fig:cifar10_heatmap}
\end{figure}

\begin{figure}[!htbp]
    \centering
    \includegraphics[width=0.9\columnwidth, height=17cm, keepaspectratio]{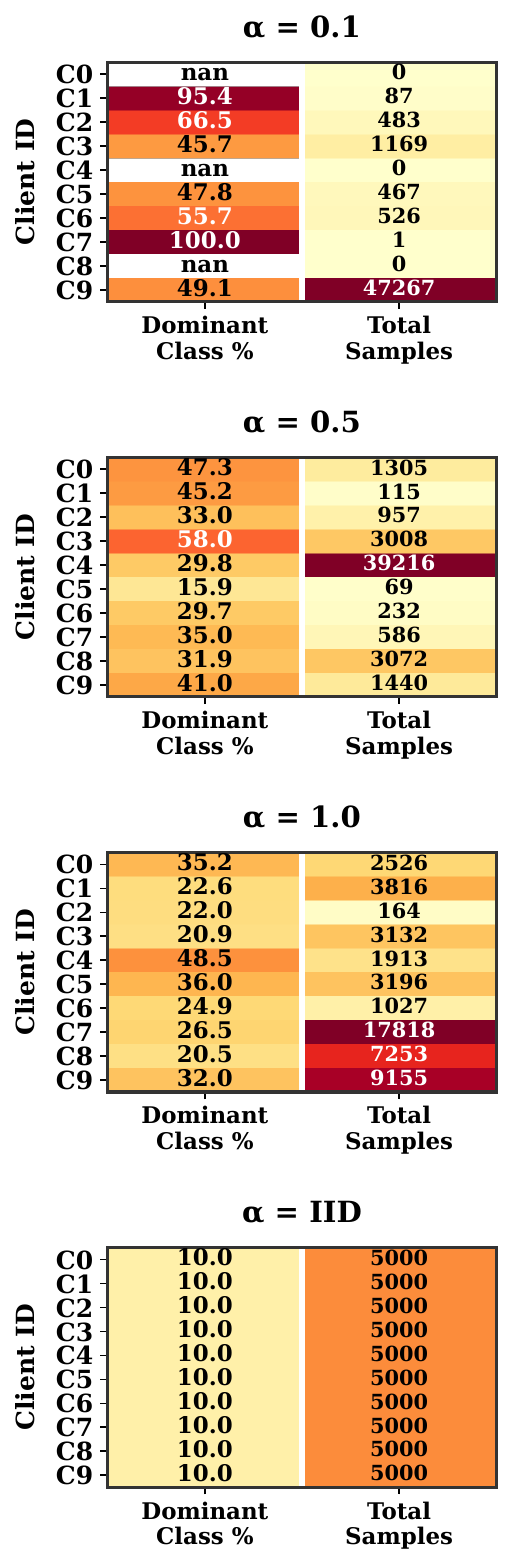}
   \caption{CIFAR-100 client distribution heatmaps for varying $\alpha$. Left: dominant class \% ; Right: sample count/100 (darker = higher). With 100 classes, $\alpha=0.1$ shows non-uniformity while IID is balanced. Dominant class percentages are diluted (e.g., Client 4: 18.4\%, 18$\times$ the 1\% baseline).}
    \label{fig:cifar100_heatmap}
\end{figure}

Figures \ref{fig:cifar10_heatmap} and \ref{fig:cifar100_heatmap} visualize data distributions via heatmaps. The left column shows dominant class percentage for each client, while the right column shows total sample count (scaled by 100 for visualization). Color intensity (yellow to red) represents magnitude: darker red indicates higher concentration or more samples.

The stark contrast between $\alpha=0.1$ and IID regimes is visually evident. Under extreme heterogeneity ($\alpha=0.1$), the heatmap exhibits highly non-uniform patterns with several clients showing deep red cells (90\%+ concentration for CIFAR-10), indicating pathological data skew. Conversely, IID partitioning produces homogeneous color patterns across all clients, confirming balanced distribution. Intermediate $\alpha$ values (0.5, 1.0) show gradual transitions from extreme to minimal heterogeneity, enabling systematic evaluation of optimizer robustness across the full spectrum of data heterogeneity.

% Figure: CIFAR-10 Heatmap

% Figure: CIFAR-100 Heatmap

\subsection{Implications for Optimizer Design}
These distribution characteristics have critical implications for quantized optimizer design:

\textbf{Momentum Stability:} Under extreme heterogeneity, clients receive highly skewed gradients (e.g., Client 5 seeing 96.7\% ship images). Quantized momentum estimates must maintain sufficient precision to capture these non-IID gradient patterns without catastrophic error accumulation across rounds.

\textbf{Variance Handling:} Variance terms in Adam-family optimizers exhibit exponential decay ($\beta_2=0.999$), leading to very small values after initial rounds. Linear INT8 quantization fails to represent these small values accurately (as shown in our Naive-INT8 baseline). Our log-space quantization addresses this by preserving relative precision across orders of magnitude.

\textbf{Asymmetric Update Budgets:} The 9.5$\times$ difference in sample counts (1,040 vs. 9,826) means different clients contribute vastly different gradient signal per round. Quantization error must not disproportionately harm small clients, which already have limited data for local optimization.

These challenges motivate our Q-LocalAdam design, which addresses heterogeneity via specialized quantizers while maintaining 4$\times$ memory efficiency.

\section{Detailed Description of Q-LocalAdam}
\label{app:algorithm_details}

This section describes the complete client-side procedure of Q-LocalAdam as implemented in our experiments, including its notation, state initialization, local updates, quantization operators, and memory representation. The central design principle is that the optimizer states are stored in compressed form between local optimizer steps but reconstructed in FP32 for each update. This reduces persistent optimizer-state storage without performing the moment updates or the model parameter updates in the quantized representation.

\subsection{Notation and State Representation}
\label{app:notation}

Let $t$ denote a communication round and let $\mathcal{S}_t$ denote the set of clients participating in that round. In our experiments all clients participate in every round. The global model broadcast by the server is denoted by $\theta_t$. Each participating client $k\in\mathcal{S}_t$ creates a local model $\theta_k$ initialized as $\theta_k=\theta_t$, together with a fresh optimizer, and trains on its local dataset $\mathcal{D}_k$. The model consists of several weight tensors (convolution and linear weights, biases, and normalization parameters), indexed by $i$.

For a local minibatch $\mathcal{B}\subseteq\mathcal{D}_k$, the client computes the gradient $g=\nabla_{\theta_k}L(\theta_k;\mathcal{B})$ of all tensors in FP32, where $L$ is the local training loss, and $g^{(i)}$ denotes the gradient of tensor $i$. The optimizer maintains a first moment, an exponentially weighted average of gradients, and a second moment, an exponentially weighted average of squared gradients.

We distinguish the stored quantized states from their reconstructed FP32 counterparts:
\begin{itemize}
    \item $m_k^{(i)}$ and $v_k^{(i)}$ denote the persistent quantized momentum and variance states of tensor $i$ on client $k$. Each is a collection of quantized blocks together with their block metadata, not a set of scalar values. Before the first update of a round they are empty, written $\varnothing$.
    \item $m$ and $v$ denote the transient FP32 momentum and variance tensors reconstructed and updated during a local step.
    \item $\hat m$ and $\hat v$ denote the bias-corrected FP32 moments used for the parameter update.
\end{itemize}

The remaining notation is as follows:
\begin{itemize}
    \item $\eta$ is the learning rate, and $\beta_1,\beta_2\in[0,1)$ are the exponential decay rates for the first and second moments.
    \item $\epsilon>0$ is the constant added to the denominator of the Adam parameter update.
    \item $\epsilon_{\log}>0$ is the additive constant used in the log-space variance quantizer.
    \item $\tau>0$ is the variance floor applied after dequantization.
    \item $B$ is the block size, which determines the number of values quantized together.
    \item $E$ is the number of local training epochs, and $s$ is the local minibatch-step counter used for bias correction.
\end{itemize}
In our experiments $\eta=10^{-3}$, $(\beta_1,\beta_2)=(0.9,0.999)$, $B=64$, and $\epsilon=\epsilon_{\log}=\tau=10^{-8}$.

The global round index $t$ and the local step counter $s$ serve different purposes. The round index identifies the model broadcast by the server, whereas $s$ counts the optimizer updates performed by a client during the current round. The counter is incremented once per minibatch, so every weight tensor uses the same value of $s$ in a given update. In the evaluated stateless-client setting, the optimizer states and the counter are reset at the beginning of each round.

\begin{algorithm}[t]
\caption{Q-LocalAdam: Detailed Client-Side Procedure}
\label{alg:Q-LocalAdam}
\small
\begin{algorithmic}[1]
\Require Global model $\theta_t$, learning rate $\eta$,
decay rates $\beta_1,\beta_2$, local epochs $E$,
block size $B$, variance floor $\tau$, constants $\epsilon,\epsilon_{\log}$

\State Server broadcasts $\theta_t$ to clients in $\mathcal{S}_t$

\For{each client $k\in\mathcal{S}_t$}
    \State $\theta_k\leftarrow\theta_t$; $s\leftarrow 0$
    \State $m_k^{(i)}\leftarrow\varnothing$, $v_k^{(i)}\leftarrow\varnothing$ for every weight tensor $i$
    \Comment{Fresh optimizer: no stored state}

    \For{epoch $e=1,\ldots,E$}
        \For{each minibatch $\mathcal{B}\subseteq\mathcal{D}_k$}
            \State $g\leftarrow\nabla_{\theta_k}L(\theta_k;\mathcal{B})$
            \Comment{FP32 gradient of all tensors}
            \State $s\leftarrow s+1$

            \For{each weight tensor $i$ of $\theta_k$}
                \State \textbf{Load previous states}
                \If{$m_k^{(i)}=\varnothing$}
                    \State $m\leftarrow 0$; $v\leftarrow 0$
                    \Comment{First update of the round}
                \Else
                    \State $m\leftarrow\operatorname{DequantizeLinear}(m_k^{(i)})$
                    \State $v\leftarrow\max(\operatorname{DequantizeLog}(v_k^{(i)}),\tau)$
                    \Comment{Variance floor, elementwise}
                \EndIf

                \State \textbf{Update moments in FP32}
                \State $m\leftarrow\beta_1m+(1-\beta_1)g^{(i)}$
                \State $v\leftarrow\beta_2v+(1-\beta_2)(g^{(i)})^2$

                \State \textbf{Store updated states}
                \State $m_k^{(i)}\leftarrow\operatorname{QuantizeLinear}(m)$
                \State $v_k^{(i)}\leftarrow\operatorname{QuantizeLog}(v)$
                \Comment{Used at the next local step}

                \State \textbf{Update parameters in FP32}
                \State $\hat m\leftarrow m/(1-\beta_1^s)$
                \State $\hat v\leftarrow v/(1-\beta_2^s)$
                \State $\theta_k^{(i)}\leftarrow
                \theta_k^{(i)}-\eta\hat m/(\sqrt{\hat v}+\epsilon)$
                \Comment{Uses the current FP32 moments}
            \EndFor
        \EndFor
    \EndFor

    \State Client sends $\theta_k$ to the server
\EndFor

\State Server computes
$\theta_{t+1}\leftarrow
\sum_{k\in\mathcal{S}_t}
\frac{|\mathcal{D}_k|}
{\sum_{j\in\mathcal{S}_t}|\mathcal{D}_j|}
\theta_k$
\end{algorithmic}
\end{algorithm}

\subsection{Initialization and Local Training}
\label{app:local_training}

At the beginning of round $t$, the server broadcasts $\theta_t$ to every participating client. Each client initializes its local model as $\theta_k=\theta_t$, creates a new optimizer, and sets $s=0$. No optimizer state exists at this point. The states are maintained locally, are discarded at the end of the round, and are never communicated to the server.

For each local minibatch, the client computes the gradient $g$ of all weight tensors in FP32 and increments the step counter, $s\leftarrow s+1$. The weight tensors are then processed one at a time. For each tensor $i$, the client loads the previously stored states, updates the moments, stores their quantized representations, and updates the weights.

\paragraph{Step 1: Loading the previous states.}
If no state is stored for tensor $i$, which is the case for the first update of every round, the moments are initialized to zero tensors, $m\leftarrow0$ and $v\leftarrow0$, and nothing is dequantized. Otherwise the optimizer reconstructs the states produced by the preceding local step:
\[
m \leftarrow \operatorname{DequantizeLinear}(m_k^{(i)}),
\]
\[
v \leftarrow
\max\!\left(
\operatorname{DequantizeLog}(v_k^{(i)}),\tau
\right).
\]
Here $\max(\cdot,\tau)$ is applied elementwise. The floor ensures that every reconstructed variance value is at least $\tau$ before the moment update. This safeguard is needed because quantization and reconstruction in log space can produce zero or negative values for sufficiently small variances. In our implementation the floor is the clamp applied inside the log dequantizer. It takes effect from the second update of a round onward, since the first update starts from exact zeros.

Dequantization occurs before the moment update because the stored blocks contain the previous optimizer state in compressed form, and Adam's moving-average equations need these values to combine them with the current gradient.

\paragraph{Step 2: FP32 moment update.}
Given the loaded states and the current gradient $g^{(i)}$ of the tensor, the client applies the exponential moving-average updates:
\[
m \leftarrow \beta_1m+(1-\beta_1)g^{(i)},
\]
\[
v \leftarrow \beta_2v+(1-\beta_2)(g^{(i)})^2.
\]
The square is elementwise. Both equations operate on FP32 tensors. The previous states entering these equations are reconstructed approximations of the states stored at the preceding step, so the resulting optimizer dynamics include the effect of quantization error in state storage.

\paragraph{Step 3: Requantization and persistent storage.}
After updating the moments, the client compresses the new states:
\[
m_k^{(i)}\leftarrow\operatorname{QuantizeLinear}(m),
\]
\[
v_k^{(i)}\leftarrow\operatorname{QuantizeLog}(v).
\]
The newly generated quantized blocks replace the previous blocks. These are the representations retained between local steps; the transient FP32 moment tensors are not kept as persistent optimizer state.

Requantization follows the moment update because the new states must first be computed from the previous states and the current gradient. It precedes the parameter update in the pseudocode, but it does not change the FP32 values used for that update: the current FP32 tensors $m$ and $v$ remain available as transient values.

\paragraph{Step 4: Bias correction and parameter update.}
The client computes the bias-corrected moments using the local step counter:
\[
\hat m=\frac{m}{1-\beta_1^s},
\qquad
\hat v=\frac{v}{1-\beta_2^s}.
\]
It then updates the weights of tensor $i$:
\[
\theta_k^{(i)}
\leftarrow
\theta_k^{(i)}
-\eta\,
\frac{\hat m}{\sqrt{\hat v}+\epsilon}.
\]
The square root, division, and update are elementwise. The current parameter update uses the newly updated FP32 moments, not their requantized reconstructions. The quantized copies affect only subsequent local steps, when they are loaded as the previous states. At the first update of a round, $s=1$, so $\hat m=g^{(i)}$ and $\hat v\approx(g^{(i)})^2$, and the update coincides with a standard first Adam step.

\subsection{Quantization Operators}
\label{app:quantizers}

Each moment tensor is flattened, padded to a multiple of $B$, and partitioned into $\lceil n/B\rceil$ contiguous blocks, where $n$ is its number of elements. Quantization is performed independently for each weight tensor and each block. The same block size is used for momentum and variance, but the two states use different quantization schemes. Quantized values are stored as unsigned 8-bit integers, so the largest quantization level is $L=255$. Padding entries are discarded after dequantization.

\paragraph{Blockwise linear quantization of momentum.}
The momentum tensor is padded with zeros. For a block $m^{(j)}$, define
\[
a^{(j)}=\min(m^{(j)}),\qquad
r^{(j)}=\max\!\big(\max(m^{(j)})-a^{(j)},\,10^{-8}\big).
\]
Each element is mapped to a quantization level:
\[
q^{(j)}
=
\operatorname{clip}\!\left(
\left\lfloor
\frac{m^{(j)}-a^{(j)}}{r^{(j)}}L
\right\rceil,\,0,\,L\right),
\]
and the reconstructed block is
\[
\tilde m^{(j)}
=
a^{(j)}+\frac{q^{(j)}}{L}r^{(j)}.
\]
The quantized values encode positions within the block's range, while the block minimum and range provide the metadata required for reconstruction. For a nondegenerate block, the maximum absolute reconstruction error is bounded by $r^{(j)}/(2L)$. The lower bound $10^{-8}$ on the range avoids division by zero for constant blocks.

\paragraph{Log-space quantization of variance.}
Variance values are nonnegative before quantization, and their magnitudes can span several orders. The variance tensor is padded with the value $\epsilon_{\log}$, and Q-LocalAdam quantizes the transformed values
\[
\ell^{(j)}=\log(v^{(j)}+\epsilon_{\log}),
\]
where non-finite entries are replaced by fixed values ($0$ for NaN, $10$ for $+\infty$, $-10$ for $-\infty$) as a numerical safeguard. Let
\[
a_\ell^{(j)}=\min(\ell^{(j)}),\qquad
r_\ell^{(j)}
=\max\!\big(\max(\ell^{(j)})-a_\ell^{(j)},\,10^{-4}\big).
\]
Quantization and reconstruction are then
\[
q_\ell^{(j)}
=
\operatorname{clip}\!\left(
\left\lfloor
\frac{\ell^{(j)}-a_\ell^{(j)}}{r_\ell^{(j)}}L
\right\rceil,\,0,\,L\right),
\]
\[
\tilde v^{(j)}
=
\max\!\left(
\exp\!\left(
a_\ell^{(j)}
+\frac{q_\ell^{(j)}}{L}r_\ell^{(j)}
\right)-\epsilon_{\log},\;\tau
\right).
\]
The logarithmic transform allocates quantization levels according to the scale of $v+\epsilon_{\log}$ rather than the absolute magnitude of $v$. The maximum reconstruction error in the log domain is at most $r_\ell^{(j)}/(2L)$, so before the subtraction of $\epsilon_{\log}$ the reconstructed value differs from $v+\epsilon_{\log}$ by a bounded multiplicative factor. This does not guarantee a bounded relative error in $v$ itself when $v$ is comparable to or smaller than $\epsilon_{\log}$, which is why the lower clamp at $\tau$ is applied after reconstruction. In our experiments $\tau=\epsilon_{\log}=10^{-8}$, so the clamp is implemented inside the log dequantizer. The minimum range $10^{-4}$ prevents precision collapse in blocks whose log-values are nearly constant.

\subsection{Persistent Memory}
\label{app:memory_layout}

Each quantized block stores $B$ one-byte values and FP32 metadata. In our GPU implementation the metadata consists of three FP32 scalars per block: the block minimum, the block maximum, and the block range. Only the minimum and the range are used for reconstruction, so the maximum is redundant. Ignoring allocator overhead, the persistent storage per block of $B$ values is therefore
\[
M_{\mathrm{block}}=B+12
\quad\text{bytes},
\]
which for $B=64$ is $76$ bytes compared with $256$ bytes for 64 FP32 values, a reduction of $256/76\approx3.37\times$. Storing only the minimum and the range gives $B+8=72$ bytes and a reduction of $256/72\approx3.56\times$, which is the layout used in our Raspberry Pi implementation.

Reported memory is obtained by summing the sizes of all stored state tensors (quantized values, block metadata, and padding) on client~0 after local training. In our GPU simulation the quantized blocks are held in host memory and moved to the accelerator when they are dequantized. These figures describe persistent moment storage, not the total memory consumed by training: gradients, model parameters, activations, and other workspaces are not reduced.

\paragraph{Relation to chunked execution.}
The description above updates one weight tensor at a time. On the Raspberry Pi we additionally split each tensor into chunks aligned with the block size $B$ and process one chunk at a time, so that only the current chunk is reconstructed in FP32. Because chunk boundaries coincide with block boundaries and quantization is blockwise, chunked execution computes the same updates as whole-tensor execution. It changes only the size of the transient FP32 workspace, not the persistent quantized representation.

\subsection{Baseline and Ablation Variants}
\label{app:variants}

All variants share the update of Algorithm~\ref{alg:Q-LocalAdam} and differ only in how the moments are stored.
\begin{itemize}
    \item \textbf{Vanilla-ClientAdam:} $m$ and $v$ are stored in FP32, without quantization and without a variance floor.
    \item \textbf{Naive INT8:} both $m$ and $v$ use the blockwise linear quantizer.
    \item \textbf{Momentum-only:} $m$ uses the linear quantizer and $v$ stays in FP32.
    \item \textbf{Variance-only:} $v$ uses the log-space quantizer and $m$ stays in FP32.
\end{itemize}

\subsection{Server Aggregation}
\label{app:aggregation}

After completing its $E$ local epochs, client $k$ transmits its updated model $\theta_k$ to the server. The server aggregates the participating models using data-size-weighted FedAvg:
\[
\theta_{t+1}
=
\sum_{k\in\mathcal{S}_t}
\frac{|\mathcal{D}_k|}
{\sum_{j\in\mathcal{S}_t}|\mathcal{D}_j|}
\theta_k,
\]
applied entry-wise to the full model state, including normalization statistics. Only model parameters are transmitted; the quantized optimizer states remain on the clients and are discarded at the end of the round. Consequently, Q-LocalAdam changes the client-side optimizer-state representation without changing the server's aggregation rule.

\section{GPU Results: Full Analysis}
\label{app:gpu-details}

This appendix provides the complete convergence, robustness, ablation, and
reproducibility analysis underlying the GPU results in
Section~\ref{sec:gpu}.

\subsection{Convergence Dynamics}
\label{app:gpu-convergence}

\begin{figure}[t]
\centering
\includegraphics[width=\columnwidth]{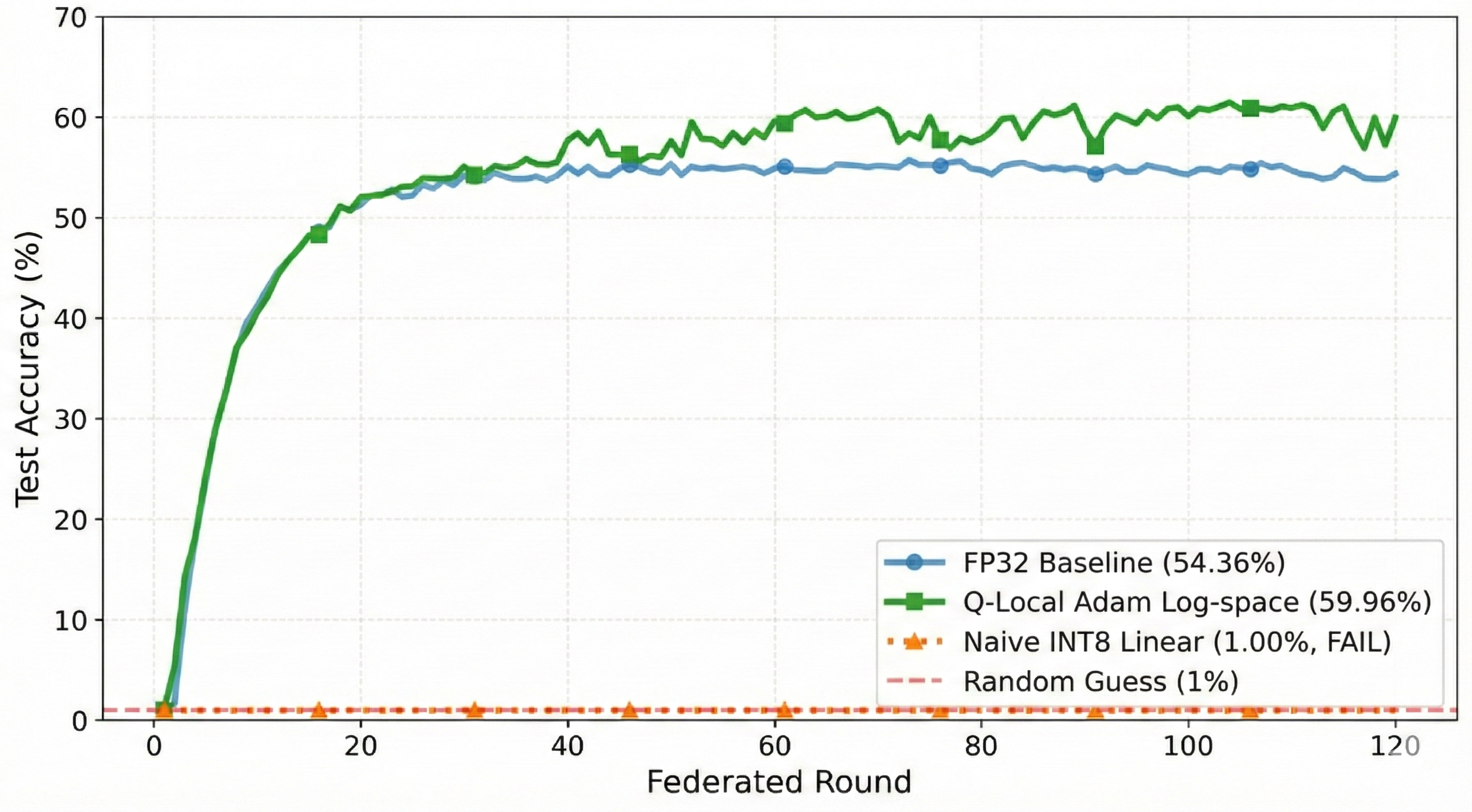}
\caption{\textbf{CIFAR-100 convergence analysis.} Under extreme heterogeneity
($\alpha{=}0.1$), Q-LocalAdam (green) pulls ahead of Vanilla-ClientAdam
(blue) to achieve a $+5.74$\,pp gain. Naive INT8 (orange dotted) flatlines at
1.00\% random guessing.}
\label{fig:convergence_cifar100}
\end{figure}

Figure~\ref{fig:convergence_cifar100} shows CIFAR-100 convergence under
extreme heterogeneity ($\alpha{=}0.1$). Q-LocalAdam tracks the FP32
convergence curve closely and pulls ahead after round 30, reaching a
$+5.74$\,pp final advantage. Naive INT8 remains flat at random guessing
throughout all 120 rounds. On CIFAR-10 ($\alpha{=}0.1$), Q-LocalAdam mirrors
the FP32 convergence curve, indicating that distribution-aware 8-bit
quantization introduces negligible optimization drag.

\subsection{Robustness Across Heterogeneity Levels}
\label{app:gpu-robustness}

Tables~\ref{tab:c10-het} and \ref{tab:c100-het} report accuracy and memory
across the full heterogeneity spectrum. Q-LocalAdam consistently matches or
exceeds FP32 performance at every level. On CIFAR-100, the accuracy gap
widens with data difficulty, from $+0.73$\,pp (IID) to $+2.15$\,pp
($\alpha{=}0.5$) and $+5.74$\,pp ($\alpha{=}0.1$).

\begin{table}[t]
\centering
\caption{CIFAR-10 robustness across heterogeneity levels.}
\label{tab:c10-het}
\small
\begin{tabular}{ccccc}
\toprule
\multirow{2}{*}{$\alpha$} & \multicolumn{2}{c}{Accuracy (\%)} & \multicolumn{2}{c}{Memory (MB)} \\
\cmidrule(lr){2-3} \cmidrule(lr){4-5}
 & FP32 & Q-LocalAdam & FP32 & Q-LocalAdam \\
\midrule
IID  & \textbf{89.09} & 88.78 & 50.35 & 14.95 \\
1.0  & 88.25 & \textbf{88.35} & 50.35 & 14.95 \\
0.5  & 87.33 & \textbf{87.44} & 50.35 & 14.95 \\
0.1  & 81.67 & \textbf{81.91} & 50.35 & 14.95 \\
\bottomrule
\end{tabular}
\end{table}

\begin{table}[t]
\centering
\caption{CIFAR-100 robustness across heterogeneity levels.}
\label{tab:c100-het}
\small
\setlength{\tabcolsep}{4pt}
\begin{tabular}{lcccc}
\toprule
$\alpha$ & \multicolumn{2}{c}{Accuracy (\%)} & \multicolumn{2}{c}{Memory (MB)} \\
\cmidrule(lr){2-3} \cmidrule(lr){4-5}
 & FP32 & Q-LocalAdam & FP32 & Q-LocalAdam \\
\midrule
IID & 62.24 & \textbf{62.97} & 89.82 & 26.67 \\
1.0 & 62.69 & \textbf{63.41} & 89.82 & 26.67 \\
0.5 & 60.95 & \textbf{63.10} & 89.82 & 26.67 \\
0.1 & 55.73 & \textbf{61.47} & 89.82 & 26.67 \\
\bottomrule
\end{tabular}
\end{table}

\subsection{Component Ablation}
\label{app:gpu-ablation}

We evaluate two partial variants on CIFAR-100 ($\alpha{=}0.1$):
momentum-only (8-bit $m$, FP32 $v$) and variance-only (FP32 $m$, 8-bit $v$).
Both yield only a $1.54\times$ memory reduction (58.24\,MB). Momentum-only
degrades accuracy to 52.49\%, while variance-only reaches 61.10\%. In
contrast, Q-LocalAdam achieves 61.47\% accuracy with a $3.37\times$ reduction
(26.67\,MB), showing that both asymmetric encodings are required for the best
accuracy--memory trade-off.

\subsection{Multi-Seed Reproducibility}
\label{app:gpu-seeds}

We repeat the evaluation with seeds 42, 123, and 456. On CIFAR-10
($\alpha{=}0.5$), Q-LocalAdam reaches $87.04{\pm}0.24\%$ versus
$86.93{\pm}0.73\%$ for FP32 ($p{=}0.78$), a statistically indistinguishable
difference. On CIFAR-100 ($\alpha{=}0.1$), Q-LocalAdam reaches
$61.40{\pm}0.82\%$ versus $54.99{\pm}1.28\%$ for FP32 ($p{=}0.005$), a
significant $+6.41$\,pp gain. Naive INT8 survives under milder skew
($85.86{\pm}1.08\%$ at $\alpha{=}0.5$) but collapses under extreme
heterogeneity ($12.74{\pm}4.75\%$), confirming that uniform quantization is
unstable specifically under severe non-IID conditions.

\subsection{Additional Sensitivity Analysis}
\label{app:ablations}
\subsubsection{Component Analysis: Detailed Analysis}
\label{app:component_ablation}

To validate that quantizing \emph{both} momentum and variance is necessary, we compare Q-LocalAdam against two variants on CIFAR-100 with $\alpha=0.1$: (1) \textbf{Momentum-only} quantizes $m$ to 8-bit while keeping $v$ in FP32, and (2) \textbf{Variance-only} quantizes $v$ to 8-bit while keeping $m$ in FP32 (Table \ref{tab:component_ablation}).

\begin{table}[!ht]
\centering
\caption{Component ablation at $\alpha=0.1$. Quantizing both momentum and variance is necessary for optimal accuracy-memory trade-offs.}
\label{tab:component_ablation}
\small
\begin{tabular}{lcrr}
\toprule
Dataset & Method & Acc (\%) & Mem (MB) \\
\midrule
\multirow{4}{*}{CIFAR-10} 
 & Vanilla-ClientAdam (FP32) & 81.67 & 50.35 \\
 & Q-LocalAdam (Both) & 81.91 & \textbf{14.95} \\
 & Momentum-only & \textbf{82.48} & 32.65 \\
 & Variance-only & 82.27 & 32.65 \\
\midrule
\multirow{4}{*}{CIFAR-100} 
 & Vanilla-ClientAdam (FP32) & 55.73 & 89.82 \\
 & Q-LocalAdam (Both) & \textbf{61.47} & \textbf{26.67} \\
 & Momentum-only & 52.49 & 58.24 \\
 & Variance-only & 61.10 & 58.24 \\
\bottomrule
\end{tabular}
\end{table}

\begin{figure*}[t]
\centering
\includegraphics[width=0.42\textwidth]{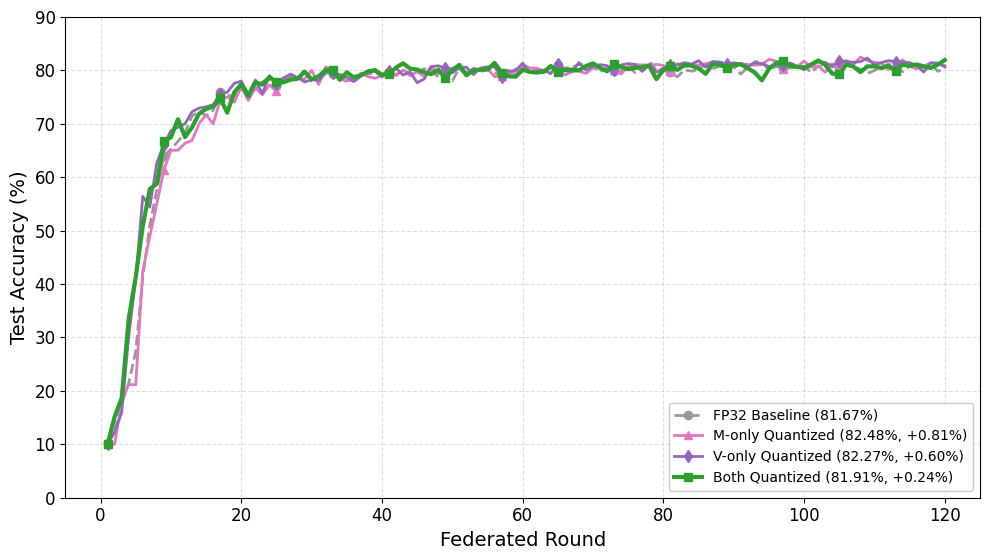}
\hfill
\includegraphics[width=0.42\textwidth]{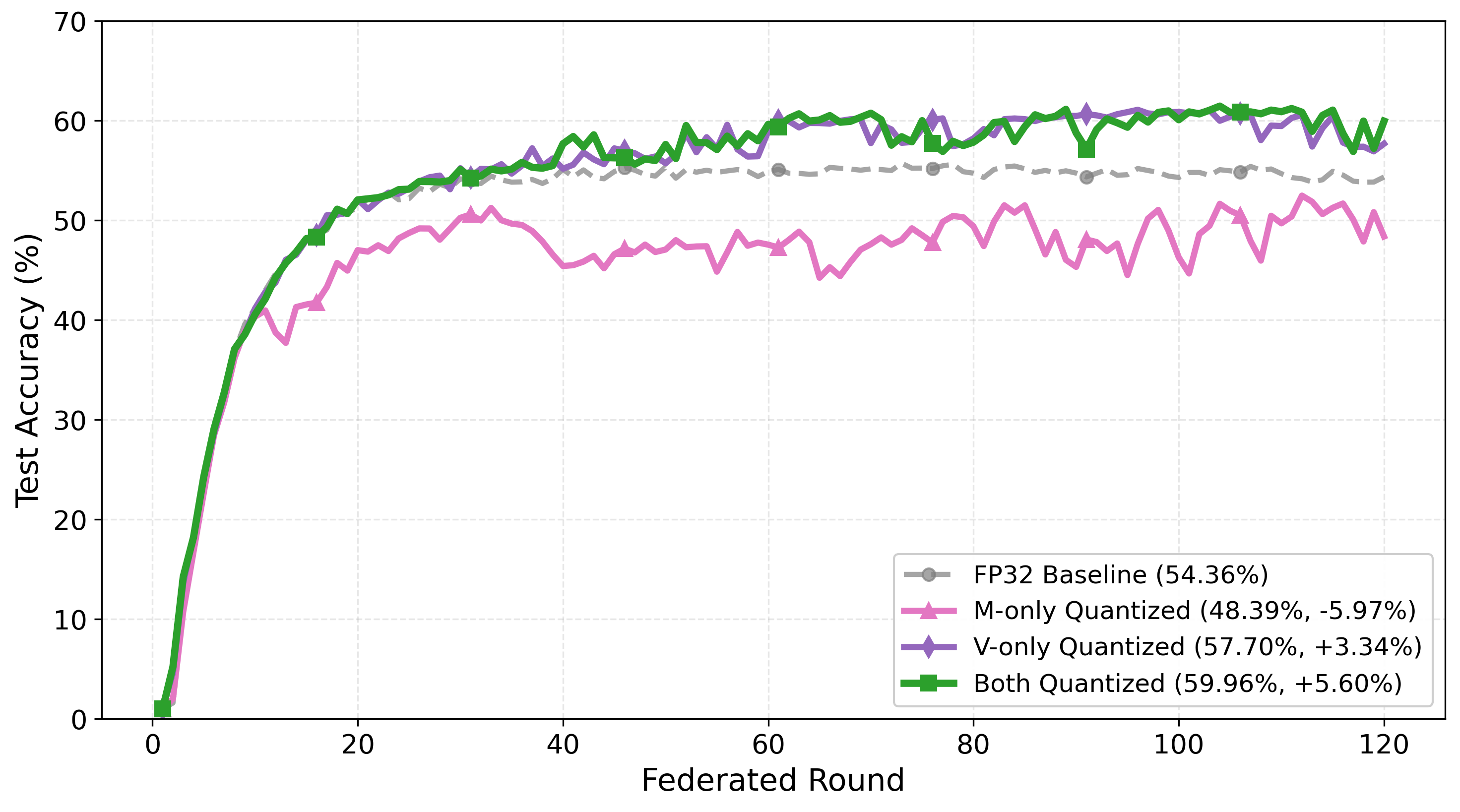}
\caption{\textbf{Component ablation at $\alpha=0.1$.} 
\textbf{Left:} CIFAR-10. Quantizing only momentum achieves competitive accuracy (82.48\%), 
but quantizing both states is essential for maximum memory reduction (3.37$\times$ vs. 1.54$\times$). 
\textbf{Right:} CIFAR-100. Quantizing only momentum or only variance provides partial benefits, 
but full Q-LocalAdam achieves the best accuracy with lowest memory.}
\Description{Two-panel line plot showing test accuracy over federated rounds for component ablation. Left panel shows CIFAR-10 with momentum-only, variance-only, and full quantization. Right panel shows CIFAR-100 with full Q-LocalAdam outperforming partial quantization.}
\label{fig:component_ablation}
\end{figure*}

\paragraph{CIFAR-100 Results.}
Figure \ref{fig:component_ablation} (right panel) shows convergence curves for all variants. Momentum-only achieves 52.49\% best accuracy with 58.24 MB optimizer memory (1.54$\times$ reduction). Variance-only reaches 61.10\% with 58.24 MB memory. Full Q-LocalAdam achieves 61.47\% accuracy with only 26.67 MB memory (3.37$\times$ reduction), demonstrating that quantizing both states with distribution-matched schemes is essential to reach the Pareto frontier. 

The results confirm our hypothesis from the empirical distribution analysis: momentum and variance have fundamentally different statistical structures and require different quantization schemes. Variance-only performs well because log-space quantization correctly handles the heavy-tailed distribution spanning 8 orders of magnitude, while linear quantization of momentum alone fails to address the variance precision bottleneck, leading to the 9pp accuracy drop (52.49\% vs. 61.47\%).

\paragraph{CIFAR-10 Results.}
On CIFAR-10 (Figure \ref{fig:component_ablation}, left panel), the pattern differs significantly: quantizing only momentum achieves 82.48\%, slightly exceeding the full Q-LocalAdam (81.91\%). This suggests that CIFAR-10's simpler 10-class structure is less sensitive to momentum quantization errors. The bounded, symmetric momentum distribution ($\pm 0.1$) combined with smaller gradient magnitudes in the easier task means that linear quantization preserves sufficient precision even when variance remains in full precision.

However, memory analysis reveals that partial quantization achieves only 1.54$\times$ compression (32.65 MB) versus Q-LocalAdam's 3.37$\times$ (14.95 MB), confirming that quantizing both states is essential for maximizing memory reduction regardless of task complexity.

\paragraph{Cross-Dataset Analysis.}
The divergent behavior of momentum-only quantization across datasets (82.48\% on CIFAR-10 vs. 52.49\% on CIFAR-100) highlights that task complexity significantly impacts quantization sensitivity. However, the consistent 3.37$\times$ memory advantage of full quantization (both $m$ and $v$) holds across all settings, making it the preferred choice for practical deployment.

This dataset-dependent sensitivity suggests that for extremely simple tasks, practitioners might consider momentum-only quantization as a lightweight alternative when memory constraints are less severe. However, for production deployments targeting diverse workloads, full quantization provides consistent benefits without task-specific tuning.

\subsubsection{Block Size Ablation: Complete Results}
\label{app:block_size}

We ablate block sizes $\{32, 64, 128\}$ for Q-LocalAdam under extreme heterogeneity ($\alpha = 0.1$) on both CIFAR-10 and CIFAR-100.

\paragraph{CIFAR-100 Results.}
Table \ref{tab:blocksize_ablation} shows that best test accuracy remains stable across block sizes (60.71\%--61.47\%), varying by less than 0.8pp. However, final accuracy at round 120 shows more sensitivity: $B = 128$ achieves only 55.97\% final accuracy compared to 59.96\% for $B = 64$ (4pp drop), suggesting that excessively large blocks may reduce quantization precision during later training stages when optimizer states have converged to smaller magnitudes. This late-stage divergence does not affect peak performance (best accuracy varies by only 0.54pp), but indicates potential instability in maintaining convergence.

Optimizer memory decreases from 30.88 MB ($B = 32$) to 24.56 MB ($B = 128$), following the expected trend of reduced per-block metadata overhead. Each block stores $B$ INT8 values (1 byte each) plus 2 FP32 scalars (min, max) for 8 bytes metadata. For $B=32$, metadata represents 25.0\% of quantized data; for $B=64$, 12.5\%; for $B=128$, 6.25\%.

\begin{figure*}[t]
\centering
\includegraphics[width=0.42\textwidth]{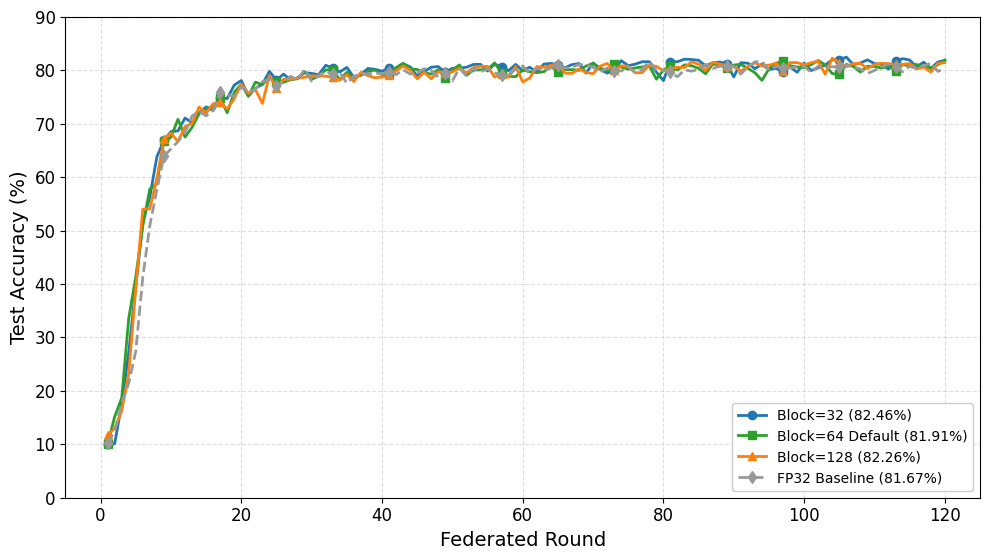}
\hfill
\includegraphics[width=0.42\textwidth]{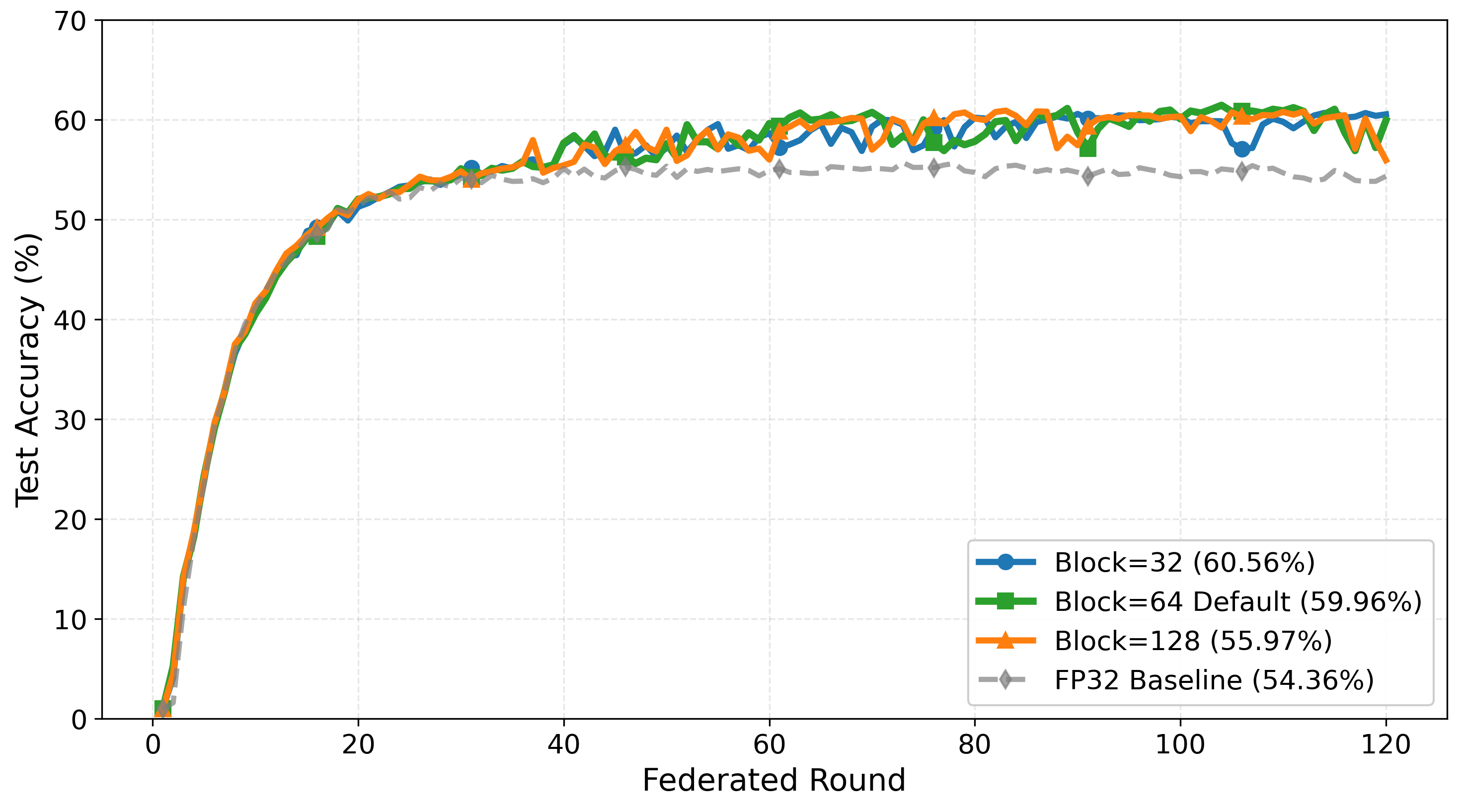}
\caption{\textbf{Block size ablation at $\alpha=0.1$.} 
\textbf{Left:} CIFAR-10. All three block sizes converge stably with best accuracy within 0.55pp 
(81.91\%--82.46\%), demonstrating robustness to quantization granularity on simpler tasks. 
\textbf{Right:} CIFAR-100. All block sizes reach similar best accuracy (0.76pp range), 
but $B=128$ exhibits late-stage instability with 4pp final accuracy drop.}
Two-panel line plot showing test accuracy over federated rounds for different block sizes. Left panel shows CIFAR-10 with three nearly overlapping curves. Right panel shows CIFAR-100 with block size 64 performing best.
\label{fig:block_size}
\end{figure*}

We select $B = 64$ as the default across all experiments, balancing peak accuracy (61.47\%, highest among all block sizes) and memory reduction (26.67 MB). Figure~\ref{fig:block_size}(right panel) visualizes convergence curves for all three block sizes on CIFAR-100.

\begin{table}[t]
\centering
\caption{Block size ablation on CIFAR-10 and CIFAR-100 with $\alpha=0.1$. Best accuracy varies by less than 0.8 percentage points across block sizes, while $B=64$ provides the best memory--accuracy trade-off.}
\label{tab:blocksize_ablation}

\footnotesize
\setlength{\tabcolsep}{3pt}
\renewcommand{\arraystretch}{1.05}

\begin{tabular}{lcccc}
\toprule
\textbf{Dataset} & \textbf{$B$} & \textbf{Best (\%)} & \textbf{Final (\%)} & \textbf{Mem (MB)} \\
\midrule
\multirow{3}{*}{CIFAR-10}
& 32  & \textbf{82.46} & 81.87 & 17.31 \\
& 64  & 81.91 & 81.91 & 14.95 \\
& 128 & 82.26 & 81.43 & 13.77 \\
\midrule
\multirow{3}{*}{CIFAR-100}
& 32  & 60.71 & 60.56 & 30.88 \\
& 64  & \textbf{61.47} & \textbf{59.96} & 26.67 \\
& 128 & 60.93 & 55.97 & \textbf{24.56} \\
\bottomrule
\end{tabular}

\end{table}

\paragraph{CIFAR-10 Results.}
Figure~\ref{fig:block_size}(left panel) shows convergence curves for all three block sizes on CIFAR-10. In contrast to CIFAR-100, all block sizes achieve nearly identical performance: best accuracy varies by only 0.55pp (81.91\% to 82.46\%), and final accuracy remains stable (0.48pp variation) without the 4pp final accuracy collapse observed at $B = 128$ on CIFAR-100. This demonstrates that task complexity significantly impacts quantization sensitivity: the simpler 10-class structure is more robust to coarser quantization granularity, while the 100-class task requires finer-grained blocks to maintain stable late-stage convergence.

Optimizer memory follows the expected trend: 17.31 MB ($B = 32$), 14.95 MB ($B = 64$), and 13.77 MB ($B = 128$), with metadata representing 25.0\%, 12.5\%, and 6.25\% of the quantized data size, respectively. Larger blocks amortize the fixed 8-byte per-block overhead more efficiently.

\paragraph{Design Principle: Block Size and Task Complexity.}
The contrast between CIFAR-10 (0.55pp best accuracy variation, stable final accuracy) and CIFAR-100 (0.76pp best accuracy variation, 4pp final accuracy drop at $B=128$) reveals that block size primarily affects convergence stability rather than peak performance. Both datasets achieve similar peak accuracies across block sizes, but CIFAR-100 exhibits late-stage divergence with large blocks. This suggests: \emph{block size should be chosen to balance peak performance and convergence stability}, with finer blocks preferred for complex tasks to avoid late-stage instability.

The mechanism behind this sensitivity likely relates to the number of distinct gradient patterns the optimizer must track. With 100 classes, variance values across different parameter groups exhibit more heterogeneity, requiring finer-grained quantization to preserve distinctions during late-stage refinement. With 10 classes, gradient patterns are more uniform, tolerating coarser quantization bins without stability issues.

\subsubsection{Learning Rate Sensitivity: Detailed Analysis}
\label{app:lr_sensitivity}

We evaluate Q-LocalAdam's sensitivity to learning rate on both datasets with $\alpha=0.1$.

\paragraph{CIFAR-100 Results.}
Figure \ref{tab:lr_sensitivity} (right panel) compares $\eta = 10^{-3}$ (default) and $\eta = 5 \times 10^{-4}$. Both settings achieve similar best accuracy (61.47\% vs. 61.82\%, 0.35pp difference), with the lower learning rate achieving slightly higher peak performance. Final accuracy shows more divergence (59.96\% vs. 58.09\%, 1.87pp), suggesting that the lower learning rate converges more slowly but reaches a comparable peak. This confirms that Q-LocalAdam does not require careful learning rate tuning and works robustly with standard hyperparameters from the federated learning literature.

\textit{Note on baseline hyperparameters:} We did not exhaustively tune Vanilla-ClientAdam (FP32) hyperparameters; it is possible that lower learning rates or adjusted $\beta_1, \beta_2$ could improve FP32 performance, though the +5.74pp gap at $\alpha=0.1$ suggests quantization provides substantial regularization benefits beyond hyperparameter effects.

\paragraph{CIFAR-10 Results.}
Figure \ref{tab:lr_sensitivity} (left panel) shows even stronger learning rate robustness on CIFAR-10. The two learning rates produce nearly identical convergence curves, with best accuracy differing by only 0.12pp (81.91\% vs. 81.79\%). This negligible gap demonstrates that Q-LocalAdam works reliably with standard FedAdam learning rates without task-specific tuning. The overlapping curves across all 120 rounds confirm stable convergence at both learning rates.

\begin{figure*}[t]
\centering
\includegraphics[width=0.42\textwidth]{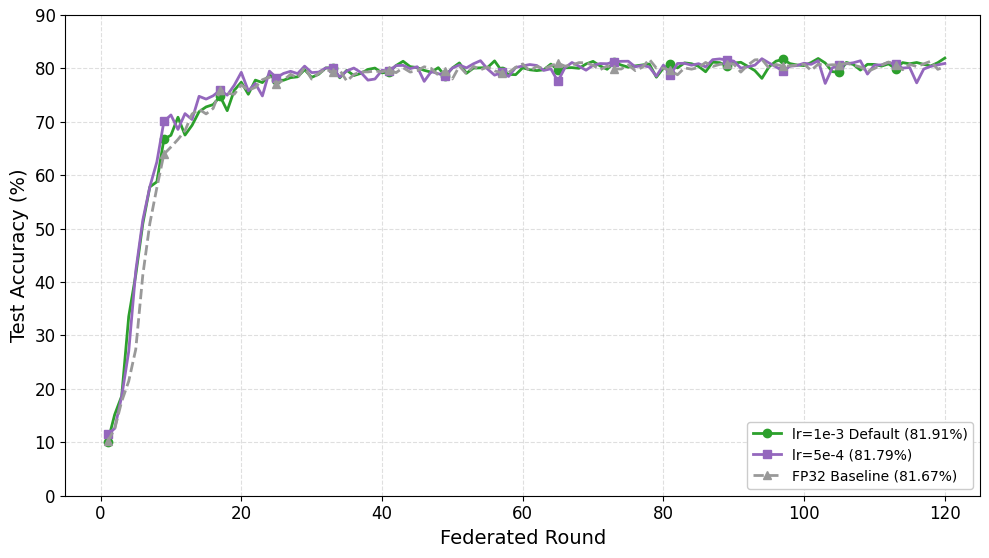}
\hfill
\includegraphics[width=0.42\textwidth]{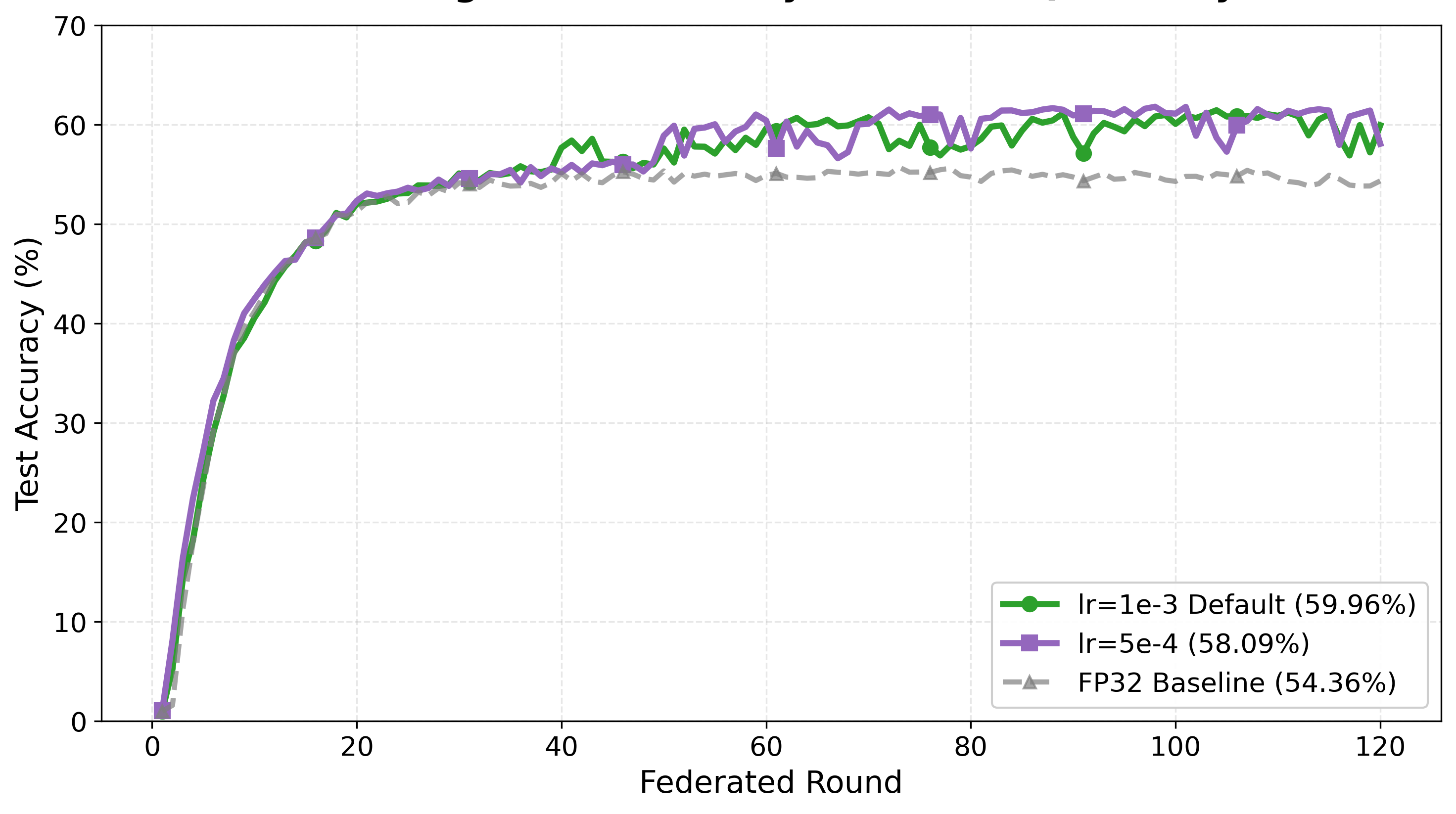}
\caption{\textbf{Learning rate sensitivity at $\alpha=0.1$.} 
\textbf{Left:} CIFAR-10. Q-LocalAdam achieves nearly identical best accuracy with 
lr=1e-3 (81.91\%) and lr=5e-4 (81.79\%), differing by only 0.12pp. 
\textbf{Right:} CIFAR-100. Both learning rates achieve similar best accuracy 
(61.47\% vs. 61.82\%), demonstrating robustness to hyperparameter choice.}
Two-panel line plot showing test accuracy over federated rounds comparing learning rates. Left panel shows CIFAR-10 with two nearly overlapping curves. Right panel shows CIFAR-100 with similar convergence patterns.\label{fig:lr_sensitivity}
\end{figure*}

\begin{table}[h]
\centering
\caption{Learning rate sensitivity comparison across datasets (best accuracies).}
\label{tab:lr_sensitivity}
\small
\begin{tabular}{lccc}
\toprule
\textbf{Dataset} & \textbf{Learning Rate} & \textbf{Best Acc (\%)} & \textbf{$\Delta$ Acc} \\
\midrule
\multirow{2}{*}{CIFAR-10}
  & $\eta = 10^{-3}$ (default) & 81.91 & -- \\
  & $\eta = 5 \times 10^{-4}$ & \textbf{81.79} & $-0.12$pp \\
\midrule
\multirow{2}{*}{CIFAR-100}
  & $\eta = 10^{-3}$ (default) & 61.47 & -- \\
  & $\eta = 5 \times 10^{-4}$ & \textbf{61.82} & $+0.35$pp \\
\bottomrule
\end{tabular}
\end{table}

\paragraph{Learning Rate Sensitivity and Task Complexity.}
Both datasets exhibit negligible learning rate sensitivity in terms of best accuracy: 0.12pp gap on CIFAR-10 and 0.35pp on CIFAR-100 (with the lower rate performing slightly better). This robustness suggests that Q-LocalAdam can be deployed with default settings (lr=$10^{-3}$, $B=64$) for typical federated vision tasks without extensive hyperparameter search. The similar peak performance across learning rates indicates that quantization does not amplify sensitivity to this critical hyperparameter.

\paragraph{Summary: Task Complexity and Quantization Robustness.}
Across all ablation studies, CIFAR-10 demonstrates substantially greater robustness to quantization hyperparameters than CIFAR-100:
\begin{itemize}
    \item Momentum-only quantization remains viable (82.48\% versus 52.49\%)
    \item Block size variation induces minimal change (0.55pp versus 0.76pp range in best accuracy, with stable final accuracy on CIFAR-10 but 4pp final accuracy drop on CIFAR-100 at $B=128$)
    \item Learning rate sensitivity is minimal on both datasets (0.12pp on CIFAR-10, 0.35pp on CIFAR-100)
\end{itemize}

Together, these results suggest a general principle: \emph{quantization hyperparameter selection should scale with task complexity}. Simpler tasks tolerate coarser settings (larger blocks, partial quantization), while harder tasks require more conservative choices (smaller blocks, full quantization) to maintain convergence stability. This observation motivates task-adaptive quantization strategies for future work, where block size and quantization scheme could be selected automatically based on dataset statistics or early-stage training metrics.

\section{Jetson Orin Nano: Extended Configuration Details}
\label{appendix:jetson_config}

\paragraph{Device specifications.}
All Jetson experiments run on an \textit{NVIDIA Jetson Orin Nano
Engineering Reference Developer Kit Super} (aarch64, 8\,GB LPDDR5
unified CPU/GPU memory, PyTorch~2.8.0). RAM usage is recorded
per-round via \texttt{/proc/meminfo}, GPU memory via
\texttt{torch.cuda.memory\_allocated()}, and CPU temperature via
\texttt{/sys/class/thermal/thermal\_zone0/temp}. Peak RAM across
120 rounds was $7{,}047.97$\,MB; no thermal throttling event
(CPU temperature $\geq 80^{\circ}$C) was observed.

\paragraph{Differences from GPU experimental setup.}
Table \ref{tab:jetson_vs_gpu} summarizes the key configuration
differences between the GPU (main paper) and Jetson experiments.
The primary changes are the input resolution, model width, and
batch size, all reduced to fit within the device's shared CPU/GPU
memory budget.

\begin{table}[h]
\centering
\caption{Configuration differences between GPU (main paper) and
Jetson Orin Nano experiments.}
\label{tab:jetson_vs_gpu}
\begin{tabular}{lcc}
\toprule
\textbf{Setting} & \textbf{GPU (Main)} & \textbf{Jetson Orin Nano} \\
\midrule
Input resolution        & $224\!\times\!224$  & $32\!\times\!32$ (native) \\
ResNet-18 width         & \texttt{width=64}   & \texttt{width=32}         \\
Batch size              & 64                  & 8                         \\
Optimizer mem.\ (FP32)  & 89.82\,MB           & 22.57\,MB                 \\
Memory reduction        & $3.37\times$        & $3.37\times$              \\
Platform                & x86\_64 / GPU       & aarch64 / Jetson          \\
PyTorch version         & 2.1                 & 2.8.0                     \\
\bottomrule
\end{tabular}
\end{table}

\paragraph{Memory compression ratio scaling.}
For a model with $N$ parameters and block size $B$, the
Q-LocalAdam optimizer memory footprint is:
\begin{equation}
\begin{aligned}
\text{Mem}_{\text{Q-LocalAdam}}
&= \underbrace{2N}_{\text{INT8 payload}}
 + \underbrace{\frac{16N}{B}}_{\text{block metadata}}
 \;\text{bytes} \\
&\approx 2.25N \;\text{bytes}
\quad (B = 64),
\end{aligned}
\label{eq:mem_complexity}
\end{equation}
where $2N$ bytes store the INT8-quantized $m$ and $v$ values,
and $16N/B$ bytes store per-block min/max FP32 scalars
(2 scalars $\times$ 4 bytes $\times$ 2 states $\times$ $N/B$
blocks). The full-precision baseline requires $8N$ bytes
(two FP32 buffers), giving a theoretical compression ratio of:
\begin{equation}
    \frac{8N}{2N + 16N/B} = \frac{8}{2 + 16/B} \;\approx\; 3.56\times \quad (B = 64).
    \label{eq:compression_ratio}
\end{equation}

Because the model size $N$ algebraically cancels out of the ratio, 
the theoretical compression efficiency is strictly constant across 
different model widths. Consequently, despite the Jetson experiments 
using a smaller ResNet-18 (\texttt{width=32}, ${\approx}2.7$\,M parameters) 
compared to the main GPU experiments (\texttt{width=64}, ${\approx}11$\,M parameters), 
the empirical memory reduction observed remains exactly $3.37\times$ on both platforms, 
maintaining consistency with the theoretical asymptote.

%-------------------REINTIALIZE------------------------

\section{Additional Analyses: Half-Precision Baselines, State Retention, and Scalability}
\label{app:additional-analyses}

This appendix provides the complete results for the half-precision baseline
comparison, the optimizer-state reinitialization ablation, and the
scalability analysis referenced in Section~\ref{sec:additional}.

\subsection{Comparison with Half-Precision Baselines}
\label{app:fp16_amp}

We benchmark Q-LocalAdam against AMPAdam~\cite{micikevicius2017mixed,wang2018training},
which uses FP16 gradients but keeps FP32 master weights and optimizer states,
and FP16Adam~\cite{kingma2014adam,micikevicius2017mixed}, which stores both
momentum and variance buffers in \texttt{float16} ($2.00\times$ memory
reduction), on CIFAR-10 and CIFAR-100 under extreme heterogeneity
($\alpha{=}0.1$, 5 clients, 120 rounds). Table~\ref{tab:fp16_amp_comparison}
reports the full comparison.

\begin{table}[t]
\centering
\caption{Memory-reduction comparison under extreme heterogeneity
($\alpha=0.1$). FP16Adam suffers from numerical instability,
whereas Q-LocalAdam retains competitive accuracy with
$3.37\times$ memory reduction.}
\label{tab:fp16_amp_comparison}
\scriptsize
\setlength{\tabcolsep}{3pt}
\renewcommand{\arraystretch}{1.1}
\begin{tabular}{@{}llcc@{}}
\toprule
\textbf{Dataset} & \textbf{Method} &
\textbf{Accuracy (\%)} & \textbf{Mem. (MB)} \\
& & \textbf{Best / Final} & \\
\midrule
\multirow{3}{*}{CIFAR-10}
& AMPAdam & 81.54 / 80.57 & 50.35 \\
& FP16Adam & 31.50 / 24.10 & 25.18 \\
& \textbf{Q-LocalAdam} &
\textbf{82.38 / 80.48} & \textbf{14.95} \\
\midrule
\multirow{3}{*}{CIFAR-100}
& AMPAdam & 55.55 / 54.35 & 89.82 \\
& FP16Adam & \multicolumn{2}{c}{\textit{Non-finite loss}} \\
& \textbf{Q-LocalAdam} &
\textbf{60.78 / 60.01} & \textbf{26.67} \\
\bottomrule
\end{tabular}

\vspace{2pt}
\begin{minipage}{\columnwidth}
\scriptsize
\textit{Reduction:} AMPAdam = $1.00\times$,
FP16Adam = $2.00\times$,
Q-LocalAdam = $3.37\times$.
\end{minipage}
\end{table}
```

\paragraph{FP16 variance underflow.}
FP16Adam fails catastrophically: on CIFAR-10 it peaks at 31.50\% with the
loss exploding by round 2, and on CIFAR-100 training diverges with non-finite
loss in the first round. This is a precision issue. Adam updates the variance
estimate as
\begin{equation}
v_t = \beta_2 v_{t-1} + (1-\beta_2)\,g_t^2.
\label{eq:adam_v}
\end{equation}
Under extreme non-IID data, gradients for missing classes can fall near
$10^{-6}$, so $g_t^2 \approx 10^{-12}$. Since \texttt{float16} cannot
represent such small values, these updates underflow to zero. As a result,
$v_t$ stalls near zero, the Adam denominator collapses to
$\sqrt{\hat{v}_t}+\epsilon \approx \epsilon$, and parameter updates become
unstable.

\paragraph{Efficiency trade-off.}
Q-LocalAdam avoids failure and improves accuracy over AMPAdam by $+0.84$\,pp
on CIFAR-10 and $+5.23$\,pp on CIFAR-100. The AUC over 120 rounds also
improves ($9{,}144$ vs.\ $9{,}065$ on CIFAR-10; $6{,}474$ vs.\ $6{,}112$ on
CIFAR-100). The cost is extra local quantize/dequantize overhead, which is
small relative to communication latency in cross-device FL.

\subsection{Effect of Optimizer-State Reinitialization}
\label{app:state-retention}

\begin{figure}[h]
\centering
\includegraphics[width=\linewidth]{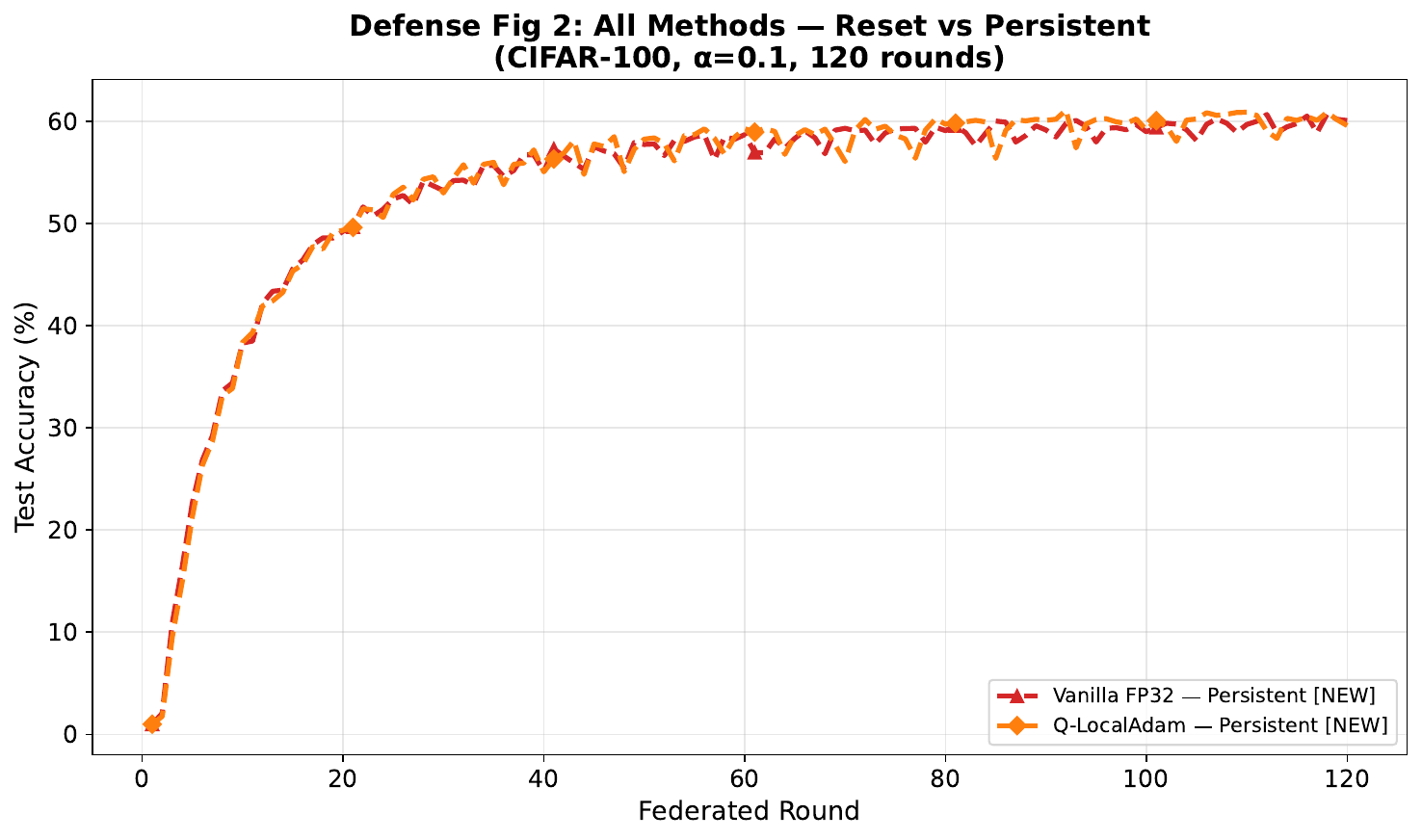}
\caption{All four configurations over 120 federated rounds (CIFAR-100,
$\alpha{=}0.1$). Solid lines: Reset (per-round reinitialization). Dashed
lines: Persistent (states carried across rounds). Q-LocalAdam-Persistent
achieves the highest best accuracy (61.07\%) across all configurations.}
\label{fig:defense_all}
\end{figure}

We compare two state-retention policies on CIFAR-100 ($\alpha{=}0.1$, 120
rounds, 5 clients, 2 local epochs, $\eta{=}10^{-3}$, seed 42,
$224{\times}224$ ResNet-18): \emph{Reset}, which reinitializes each client's
optimizer state at the start of every round, and \emph{Persistent}, which
retains it across rounds. Table~\ref{tab:defense_full} reports the full
results, and Figure~\ref{fig:defense_all} shows the convergence curves for
all four configurations.

\begin{table}[h]
\centering
\caption{All reinitialization experiments: CIFAR-100, $\alpha{=}0.1$, 120
rounds, 5 clients, 2 local epochs, $\eta{=}10^{-3}$, seed 42,
$224{\times}224$ ResNet-18.}
\label{tab:defense_full}
\footnotesize
\setlength{\tabcolsep}{3pt}
\renewcommand{\arraystretch}{1.1}
\begin{tabular}{p{0.34\columnwidth}lcc}
\toprule
\textbf{Method} &
\textbf{Config} &
\begin{tabular}[c]{@{}c@{}}\textbf{Best}\\\textbf{Acc (\%)}\end{tabular} &
\begin{tabular}[c]{@{}c@{}}\textbf{Final}\\\textbf{Acc (\%)}\end{tabular} \\
\midrule
Vanilla-ClientAdam (FP32) & Reset      & 55.73 & 54.36 \\
Vanilla-ClientAdam (FP32) & Persistent & 60.69 & 60.06 \\
Q-LocalAdam (Ours)        & Reset      & \textbf{61.47} & 59.96 \\
Q-LocalAdam (Ours)        & Persistent & 61.07 & 59.57 \\
\midrule
\multicolumn{4}{l}{\textit{Effect of reinitialization:}} \\
\quad Vanilla: Reset $-$ Persistent & & $-4.96$\,pp & \\
\quad Q-LocalAdam: Reset $-$ Persistent & & $+0.40$\,pp & \\
\midrule
\multicolumn{4}{l}{\textit{Q-LocalAdam advantage over Vanilla:}} \\
\quad Reset condition      & & $+5.74$\,pp & \\
\quad Persistent condition & & $+0.38$\,pp & \\
\bottomrule
\end{tabular}
\end{table}

\begin{figure}[h]
\centering
\includegraphics[width=\linewidth]{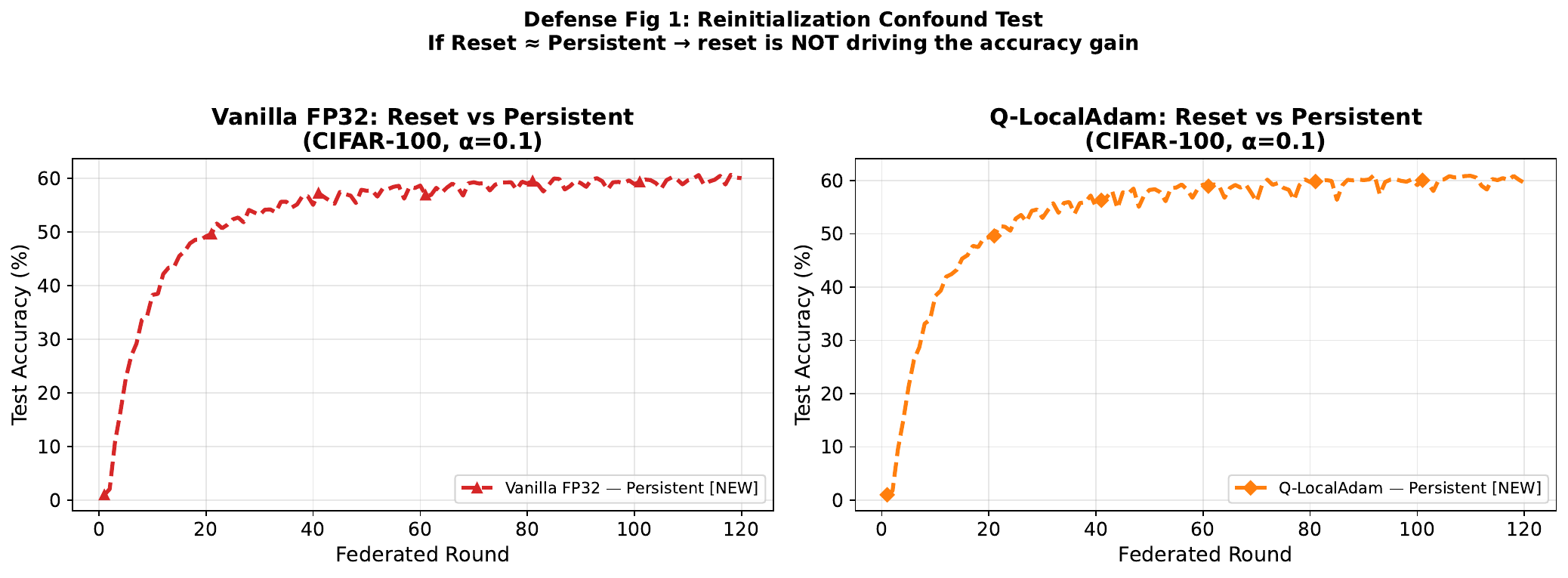}
\caption{Convergence under Reset and Persistent conditions for Vanilla FP32
and Q-LocalAdam.}
\label{fig:defense}
\end{figure}

The persistent variants converge more slowly in early rounds (1--30) due to
stale momentum conflicting with shifted local data distributions under
$\alpha{=}0.1$. Both persistent variants recover and surpass their reset
counterparts by round 50, after which momentum and variance estimates have
adapted sufficiently to the non-IID distribution. Retaining optimizer state
increases FP32 Adam's peak and final accuracies by 4.96 and 5.70 percentage
points, respectively, while Q-LocalAdam produces similar observed accuracies
under both policies (61.47\% peak / 59.96\% final under Reset vs.\ 61.07\% /
59.57\% under Persistent). These results show that Q-LocalAdam trains
successfully with either policy and does not require the Reset schedule.
They do not establish that quantization itself improves accuracy, since the
variance floor also affects optimization and attributing accuracy differences
to quantization requires matched floors.

\subsection{Scalability to Larger Client Pools}
\label{app:scalability}

We expand the extreme heterogeneity configuration (CIFAR-100,
$\alpha{=}0.1$) to pools of 50 and 100 clients under full participation,
using Q-LocalAdam with 120 total rounds, 2 local epochs, $\eta{=}10^{-3}$,
and $224{\times}224$ ResNet-18. Because Q-LocalAdam's block-wise memory
footprint is $2N + 16N/B$ bytes (asymptotically $2.25N$ at $B{=}64$, a
$3.56\times$ theoretical reduction), per-client optimizer memory remains
clamped at 26.67\,MB regardless of pool size. In a 100-client deployment,
aggregate optimizer memory falls from 8{,}982\,MB (FP32) to 2{,}667\,MB, a
reduction of over 6.3\,GB. Table~\ref{tab:scalability} reports the headline
result and Table~\ref{tab:scalability_rounds} reports per-round accuracy.

\begin{table}[h]
\centering
\caption{Q-LocalAdam scalability on CIFAR-100 ($\alpha{=}0.1$) over 120
rounds. Per-client memory is bounded at 26.67\,MB regardless of pool size.}
\label{tab:scalability}
\small
\begin{tabular}{lccc}
\toprule
\textbf{Clients} & \textbf{Clients/Round} & \textbf{Best Acc (\%)} & \textbf{Final Acc (\%)} \\
\midrule
50  & 50  & \textbf{55.85} & \textbf{55.85} \\
100 & 100 & 47.66 & 47.66 \\
\bottomrule
\end{tabular}
\end{table}

\begin{table}[h]
\centering
\caption{Q-LocalAdam test accuracy (\%) at selected rounds across client
pool sizes (CIFAR-100, $\alpha{=}0.1$, full participation).}
\label{tab:scalability_rounds}
\small
\begin{tabular}{ccc}
\toprule
\textbf{Round} & \textbf{50 Clients} & \textbf{100 Clients} \\
\midrule
1   & 1.00  & 1.00  \\
5   & 4.38  & 2.49  \\
10  & 9.76  & 8.01  \\
20  & 18.23 & 14.83 \\
30  & 25.79 & 20.23 \\
40  & 32.87 & 24.99 \\
50  & 38.96 & 28.74 \\
60  & 43.15 & 32.65 \\
70  & 46.50 & 36.48 \\
80  & 49.43 & 40.03 \\
90  & 51.42 & 42.04 \\
100 & 53.32 & 44.39 \\
110 & 54.44 & 46.43 \\
120 & \textbf{55.85} & 47.66 \\
\bottomrule
\end{tabular}
\end{table}

The 50-client setting converges more slowly in early rounds but reaches
55.85\% by round 120. The 100-client setting converges more slowly due to
more extreme per-client data skew under $\alpha{=}0.1$ across a larger pool,
reaching 47.66\% at round 120. The accuracy gap between the 50- and
100-client settings narrows consistently across rounds (from $\Delta{=}5.93$\,pp
at round 60 to $\Delta{=}8.19$\,pp at round 120), suggesting the 100-client
setting is still on an upward trajectory; extending training beyond 120
rounds is expected to close this gap and is left for future work.

\subsection{Asymptotic Complexity Derivation}
\label{app:complexity}

Standard FP32 adaptive optimizers require $8N$ bytes to maintain momentum and
variance buffers for $N$ parameters (4 bytes each for $m$ and $v$).
Q-LocalAdam replaces these with 1-byte INT8 buffers, partitioned into
contiguous blocks of size $B$, where each block additionally stores two FP32
metadata scalars (min/range or their log-space equivalents, 8 bytes total)
per state tensor. This gives a total memory footprint of
\begin{equation*}
\text{Memory}_{\text{Q-LocalAdam}} = 2N + \frac{16N}{B} \text{ bytes},
\end{equation*}
where the $2N$ term accounts for the two INT8 buffers ($m$ and $v$, 1 byte
each per parameter) and the $16N/B$ term accounts for the FP32 metadata
overhead (8 bytes per block, 2 tensors, $N/B$ blocks per tensor).

At $B{=}64$, this evaluates to an asymptotic footprint of $2.25N$ bytes,
compared to $8N$ bytes for FP32, yielding a theoretical compression factor of
\begin{equation*}
\frac{8N}{2.25N} = 3.56\times.
\end{equation*}
This factor is independent of $N$ and therefore holds at any model scale,
from small CNNs to billion-parameter architectures.
Table~\ref{tab:scaling} projects this scaling behavior across representative
parameter counts.

\begin{table}[h]
\centering
\caption{Projected optimizer memory scaling. The theoretical $3.56\times$
compression factor facilitates the deployment of massive architectures on
constrained edge hardware.}
\label{tab:scaling}
\small
\begin{tabular}{lccc}
\toprule
\textbf{Parameters} & \textbf{FP32 (MB)} & \textbf{Q-LocalAdam (MB)} & \textbf{Reduction} \\
\midrule
10M   & 76.3   & 21.5   & $3.56\times$ \\
100M  & 762.9  & 214.6  & $3.56\times$ \\
1B    & 7,629  & 2,146  & $3.56\times$ \\
10B   & 76,294 & 21,458 & $3.56\times$ \\
\bottomrule
\end{tabular}
\end{table}

\paragraph{Execution and network scalability.}
Block-wise quantization isolates parameters into contiguous partitions with
zero cross-parameter dependencies, avoiding serialization bottlenecks by
mapping directly to standard parallel hardware primitives (e.g., vectorized
min/max reductions and elementwise exponentiation on GPU/TPU backends).
Furthermore, because the memory footprint is entirely dictated by $N$,
per-client overhead remains clamped at $2.25N$ bytes regardless of the number
of participating clients, ensuring the optimizer's memory requirements remain
fully decoupled from the size of the federated network. This property is
empirically confirmed in Appendix~\ref{app:scalability}, where per-client
memory stays fixed at 26.67\,MB across pools of 5, 50, and 100 clients.

%=----------------------------VIT------------------------

\section{ViT-Tiny Generalization and Collapse Diagnostics}
\label{app:vit_failure}

This appendix provides the complete ViT-Tiny results referenced in
Section~\ref{sec:transformer}, including the per-method breakdown and the
numerical diagnostics behind the naive INT8 collapse.

\subsection{Full ViT-Tiny Results}
\label{app:vit-results}

Table~\ref{tab:transformer} reports the full ViT-Tiny comparison on CIFAR-10
($\alpha{=}0.1$, 100 rounds). Q-LocalAdam compresses optimizer memory from
43.93\,MB to 13.04\,MB ($3.37\times$) and attains peak accuracy of 63.30\%,
$+0.49$\,pp above FP32. Naive INT8 reaches the same memory footprint but
collapses to 10.00\% final accuracy.

\begin{table*}[t]
\centering
\caption{ViT-Tiny generalization (CIFAR-10, $\alpha{=}0.1$, 100 rounds).}
\label{tab:transformer}
\small
\begin{tabular}{lcccc}
\toprule
\textbf{Method}
  & \textbf{Best Acc.\ (\%)}
  & \textbf{Final Acc.\ (\%)}
  & \textbf{Opt.\ Mem.\ (MB)}
  & \textbf{Reduction} \\
\midrule
Vanilla-ClientAdam (FP32)   & 62.81 & 61.65 & 43.93 & -- \\
\textbf{Q-LocalAdam (Ours)} & \textbf{63.30} & \textbf{62.63} & \textbf{13.04} & $\mathbf{3.37\times}$ \\
Naive INT8                  & 38.32 & 10.00 & 13.04 & $3.37\times$ \\
\bottomrule
\end{tabular}
\end{table*}

\subsection{Collapse Diagnostics}
\label{app:vit-diagnostics}

\begin{figure}[h]
\centering
\includegraphics[width=0.9\linewidth,height=0.32\textheight,keepaspectratio]{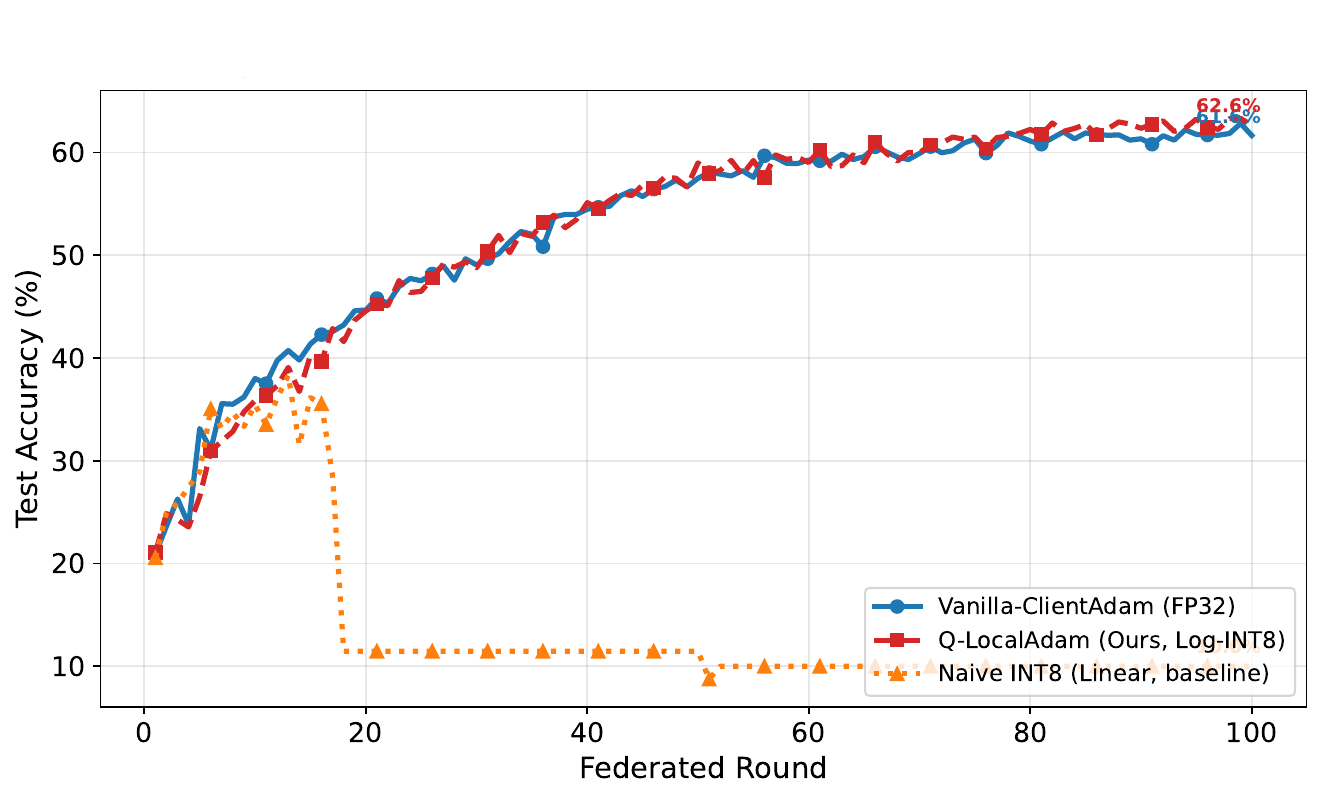}
\caption{ViT-Tiny convergence (CIFAR-10, $\alpha{=}0.1$). Naive INT8
collapses to 10\% random guessing by round 55; Q-LocalAdam tracks and
slightly exceeds FP32.}
\label{fig:vit_failure}
\end{figure}

Figure~\ref{fig:vit_failure} shows the ViT-Tiny convergence trajectory. Naive
INT8 initially learns but collapses to 10\% random guessing by round 55,
while Q-LocalAdam tracks and slightly exceeds the FP32 baseline throughout.
Per-round optimizer-state norms show that the naive INT8 variance buffer
degrades sharply before collapse, and the post-collapse optimizer-memory
trace confirms that the optimizer state drops to zero, consistent with
variance underflow destroying the Adam denominator. These traces replicate
the analysis summarized in Section~\ref{sec:transformer}.

\section{Additional Analysis on federated text data}
\label{sec:nlp_appendix}

\begin{figure}[t]
  \centering
  \includegraphics[width=\linewidth]{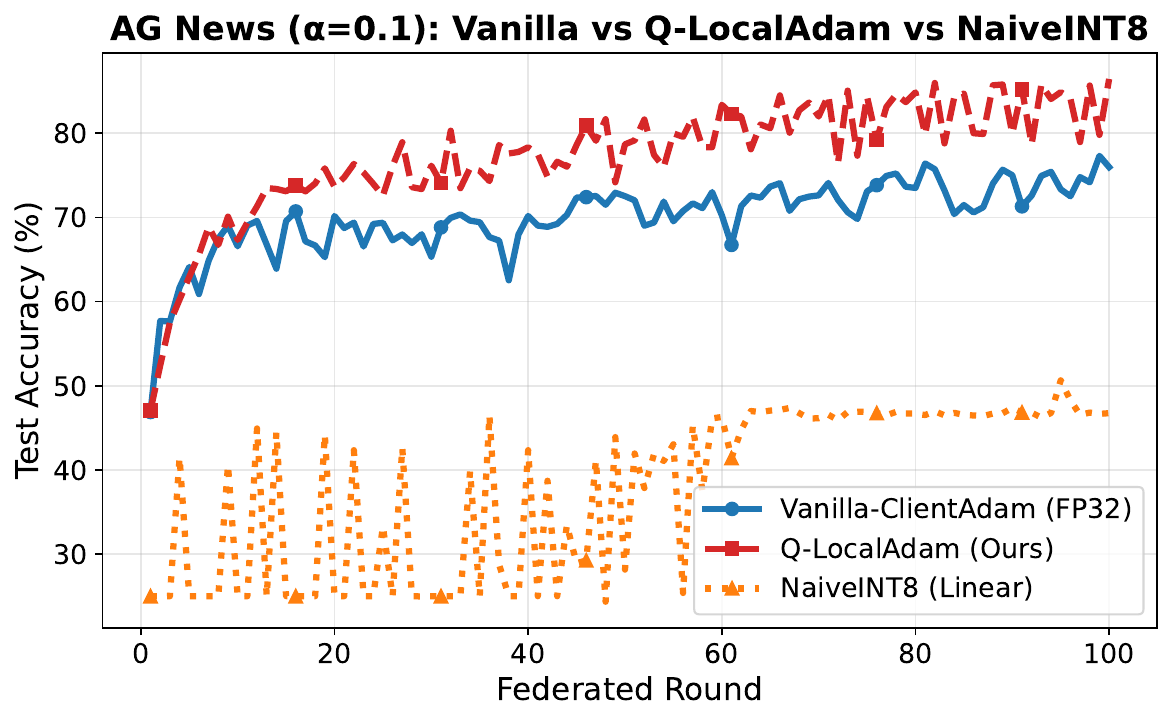}
  \caption{AG~News ($\alpha=0.1$, 5 clients, 100 rounds): Q-LocalAdam outperforms Vanilla-ClientAdam~FP32 and Naive~INT8.}
  \label{fig:agnews_main}
\end{figure}

\begin{figure}[t]
  \centering
  \includegraphics[width=\linewidth]{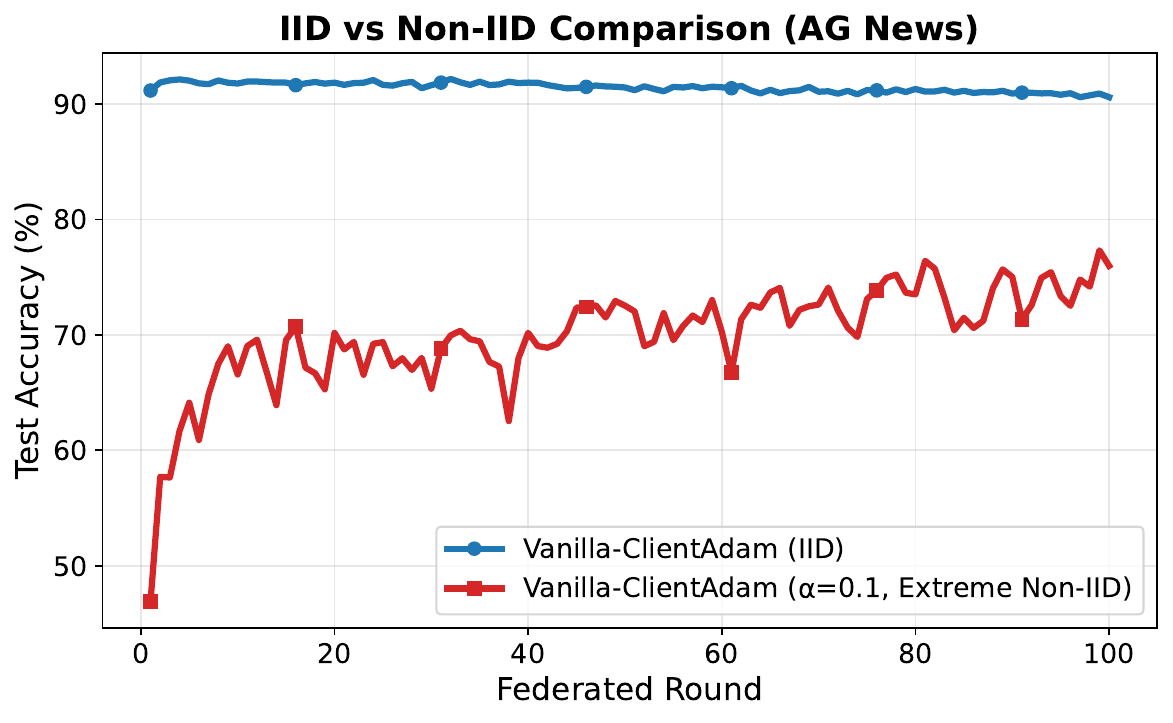}
  \caption{IID vs.\ non-IID convergence on AG~News (Vanilla-ClientAdam).}
  \label{fig:agnews_iid_noniid}
\end{figure}

\paragraph{Additional interpretation.}
The $+9.13$\,pp gain on AG~News exceeds the gains observed on CIFAR-100 and ViT-Tiny, suggesting that log-space quantization can be especially beneficial under severe heterogeneity. This pattern supports the view that multiplicative control of variance is important when client drift produces highly uneven token-level updates(Figures \ref{fig:agnews_main},\ref{fig:agnews_iid_noniid}).

\begin{table}[t]
\centering
\caption{Q-LocalAdam across tasks, architectures, and modalities ($\alpha=0.1$).}
\label{tab:cross_domain}
\footnotesize
\begin{tabular}{llccc}
\toprule
Task & Backbone & FP32 & $\Delta$Acc & Red. \\
\midrule
C10   & ResNet-18       & 81.67 & +0.24 & 3.37$\times$ \\
C100  & ResNet-18       & 55.73 & +5.74 & 3.37$\times$ \\
C10   & ViT-Tiny        & 62.81 & +0.49 & 3.37$\times$ \\
AGNews & TextTransformer & 77.32 & +9.13 & 3.37$\times$ \\
\bottomrule
\end{tabular}
\end{table}

\

\section{Analysis of Optimizer Dynamics}

While the primary contribution of Q-LocalAdam is its substantial reduction in client-side optimizer memory, the results on highly heterogeneous data also show a consistent improvement in model accuracy. To better understand the source of this improvement, we conducted a controlled diagnostic analysis comparing Vanilla-ClientAdam and Q-LocalAdam under identical training conditions. Both optimizers were initialized with the same model parameters and trained using identical client data, minibatch ordering, thereby isolating differences arising solely from optimizer-state representation.

This analysis is motivated by the update rule of Adam-style optimizers, where parameter updates are inversely proportional to the square root of the second-moment estimate. If this estimate becomes excessively small for infrequently updated parameters, the corresponding adaptive learning rates can become disproportionately large, leading to unstable optimization. We therefore investigate whether the log-space quantization employed by Q-LocalAdam alters the evolution of the second-moment estimates in a manner that is particularly beneficial under severe non-IID data.

The diagnostic consists of two complementary analyses. First, a single client was trained for 200 local optimization steps while logging statistics of the classifier head (\texttt{fc.weight}), including the mean and minimum second-moment estimates, update norms, and total model weight norm. Second, the classifier head was analyzed at the row level. Since the CIFAR-100 classifier contains one row per class, rows were grouped according to the number of samples of that class present on the selected client: zero samples, rare (1--5 samples), medium (6--50 samples), and majority (more than 50 samples). The resulting statistics are summarized in Table~\ref{tab:supp_diagnostics}..

The diagnostics reveal a consistent pattern. The smallest second-moment estimates occur in the minority-class rows, precisely where client heterogeneity produces the sparsest parameter updates. Compared with Vanilla-ClientAdam, Q-LocalAdam consistently maintains a slightly higher lower bound on the second-moment estimate for zero- and low-frequency classes, while producing comparable update norms and slightly smaller overall model weight norms. These observations are consistent with the hypothesis that log-space variance encoding prevents the adaptive statistics of infrequently updated parameters from collapsing excessively, thereby stabilizing optimization for minority-class parameters.

This interpretation is consistent with the primary experimental results obtained under the same CIFAR-100 setting ($\alpha=0.1$). Q-LocalAdam improves the final test accuracy by 5.10 percentage points while simultaneously reducing optimizer-state memory by a factor of $3.37\times$. Taken together, these diagnostics suggest that the performance gain is not solely a consequence of memory compression. Rather, the quantized representation of the second-moment state appears to modify the optimization dynamics in a manner that is beneficial under severe client heterogeneity.

Although these diagnostics do not establish a formal causal mechanism, they provide empirical evidence that log-space variance encoding acts as an implicit stabilizing regularizer for adaptive optimization under extreme non-IID data, particularly for minority-class parameters whose variance estimates are most susceptible to collapse.

\begin{table}[t]
\centering
\caption{Supplementary optimizer diagnostics on CIFAR-100 ($\alpha=0.1$). Higher minimum second-moment values indicate a stronger variance floor.}
\label{tab:supp_diagnostics}
\small
\setlength{\tabcolsep}{4pt}
\begin{tabular}{lcc}
\hline
\textbf{Overall Metrics} & \textbf{Vanilla} & \textbf{Q-LocalAdam} \\
\hline
Accuracy (\%) & 55.55 & \textbf{60.65} \\
Optimizer memory (MB) & 89.82 & \textbf{26.67} \\
Mean $v$ (\texttt{fc.weight}) & $3.629{\times}10^{-5}$ & $\mathbf{3.870{\times}10^{-5}}$ \\
Min $v$ (\texttt{fc.weight}) & $3.265{\times}10^{-9}$ & $\mathbf{9.990{\times}10^{-9}}$ \\
Update norm & 35.3977 & 35.4452 \\
Weight norm & 84.787 & \textbf{82.473} \\
\hline
\multicolumn{3}{c}{\textbf{Mean Minimum Second Moment ($v$)}}\\
\hline
Zero samples & $6.44{\times}10^{-8}$ & $\mathbf{6.57{\times}10^{-8}}$ \\
Rare (1--5) & $4.33{\times}10^{-8}$ & $\mathbf{4.41{\times}10^{-8}}$ \\
Medium (6--50) & $\mathbf{4.59{\times}10^{-7}}$ & $3.96{\times}10^{-7}$ \\
Majority ($>$50) & $\mathbf{7.86{\times}10^{-6}}$ & $7.82{\times}10^{-6}$ \\
\hline
\multicolumn{3}{c}{\textbf{Mean Update Norm}}\\
\hline
Zero samples & 0.493 & 0.510 \\
Rare (1--5) & 2.503 & 2.636 \\
Medium (6--50) & 3.879 & 3.834 \\
Majority ($>$50) & 4.432 & 4.463 \\
\hline
\end{tabular}
\end{table}

\section{Extending Q-LocalAdam to 4-Bit Optimizer States: Details}
\label{app:4bit}

This section gives the design, protocol and full results behind
Section~\ref{sec:fourbit}.

\paragraph{Design and memory layout.}
The 4-bit variant retains the quantization schemes of the 8-bit method (linear
for momentum, log-space for variance) but uses 16 levels ($0$--$15$) instead of
256. Two codes are packed into each byte, so a 64-element block stores 32 bytes
of codes plus two FP32 metadata values (block minimum and range, 8 bytes): 40
bytes versus 256 bytes in FP32, a $6.4\times$ reduction per optimizer state. The
block maximum is not stored because it is recoverable from the minimum and
range. The Adam update, the per-step dequantize--update--requantize cycle,
per-round state reinitialization and FedAvg aggregation are unchanged.

\paragraph{Protocol and metrics.}
All 4-bit experiments follow the main protocol: CIFAR-100, ResNet-18 with
$224{\times}224$ inputs, 5 clients, 2 local epochs, 120 rounds, batch size 64,
$\eta{=}10^{-3}$, and Dirichlet partitions with $\alpha\in\{0.1,0.5,1.0\}$. We
report peak accuracy (maximum over 120 rounds), final accuracy (round 120) and,
where noted, the mean over the last 10 rounds, because peak accuracy is a
maximum over noisy rounds and is biased upward. Optimizer memory is measured
from the stored tensors on client~0, including metadata and padding.

\paragraph{Relation to prior 4-bit optimizers.}
Li et al.~\cite{li2023memory} previously pushed Adam optimizer states to 4 bits,
identifying a zero-point problem in which mapping small positive second-moment
values to zero produces excessively large updates. We observe the same failure
under non-IID federated training: Naive INT4, which uses a linear quantizer for
both moments, remains at chance accuracy in all 120 rounds. We address the
problem differently: instead of excluding zero from a linear grid, we quantize
the variance in log space, which keeps every reconstructed value strictly
positive after the variance floor $\tau$ and bounds the reconstruction error
multiplicatively across the full dynamic range.

\paragraph{Memory.}
At $B{=}64$, the 4-bit variant reduces client optimizer memory from 89.82~MB
(FP32) to 14.03~MB, a $6.40\times$ reduction and $1.9\times$ smaller than the
8-bit variant (26.67~MB). This agrees with the analytical estimate: each element
costs 4 bits plus $64/64=1$ bit of amortized metadata, i.e., 5 bits per element
per state, so for the 11.2M-parameter ResNet-18 two states occupy
$2\times 11.2\mathrm{M}\times 5/8\approx 14.0$~MB. Wall-clock time per
experiment increased from 3.95~h (FP32) to 4.40~h (4-bit, $\alpha{=}0.1$), about
11\%, because the $224{\times}224$ forward/backward passes dominate the total
time.

\paragraph{Multi-seed validation of the headline result.}
We repeat the main 4-bit comparison on CIFAR-100 ($\alpha{=}0.1$,
$\eta{=}10^{-3}$) over three seeds (42, 123, 456), which change the Dirichlet
partition, the initialization and the data order. 4-bit Q-LocalAdam reaches
$60.25\pm0.33\%$ final accuracy ($60.27\pm0.52\%$ averaged over the last 10
rounds, $60.99\pm0.85\%$ peak) compared with $53.95\pm0.83\%$
($53.79\pm0.96\%$, $54.58\pm0.96\%$) for FP32 Adam, a gain of $+6.3$~pp (final)
and $+6.5$~pp (last-10 mean) with $6.4\times$ less optimizer memory (Welch
$p\approx0.002$, $n{=}3$). The gain is present in every seed: the final
accuracies of 4-bit Q-LocalAdam (60.60, 59.94, 60.21) all exceed the best FP32
seed (54.90) by at least 5.0~pp, and its cross-seed standard deviation is lower
than that of FP32. As Section~\ref{sec:floor} shows, FP32 Adam also benefits
from a variance floor, so we do not attribute this gap to quantization itself.

\paragraph{Matched learning rate.}
A natural concern is that $\eta{=}10^{-3}$ is simply too high for FP32 Adam. At
$\eta{=}5{\times}10^{-4}$ (seed 42) FP32 Adam improves from 54.90\% to 56.53\%
final accuracy, and 4-bit Q-LocalAdam improves from 60.60\% to 61.72\%. The gap
is essentially unchanged ($+5.70$~pp at $10^{-3}$, $+5.19$~pp at
$5{\times}10^{-4}$), so the advantage is not an artifact of an untuned FP32
baseline.

\begin{figure}[t]
    \centering
    \includegraphics[width=\columnwidth]{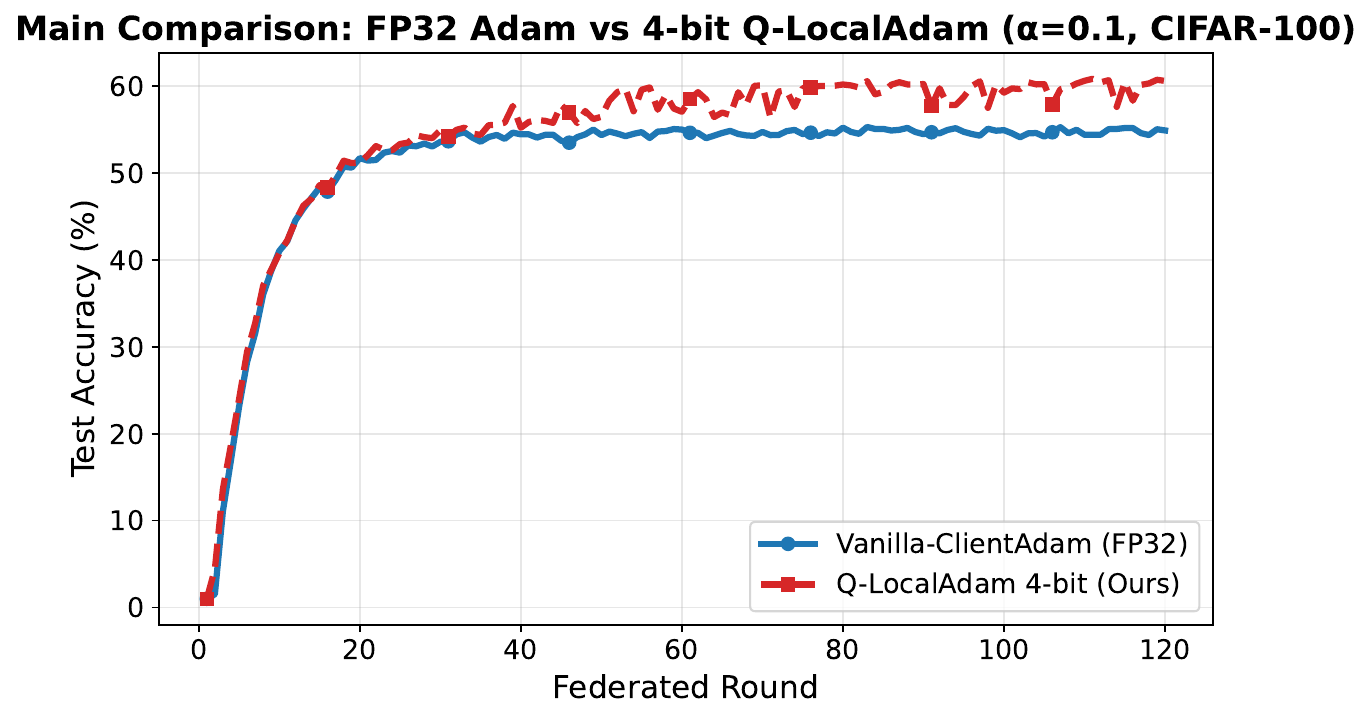}
    \caption{CIFAR-100 ($\alpha{=}0.1$): FP32 Adam vs.\ 4-bit Q-LocalAdam over 120 rounds (seed 42). The 4-bit variant tracks FP32 Adam early and leads after round 30.}
    \label{fig:4bit_conv}
\end{figure}

\paragraph{Accuracy and convergence under extreme heterogeneity.}
At $\alpha{=}0.1$ (seed 42), 4-bit Q-LocalAdam reaches 60.84\% peak and 60.60\%
final accuracy, compared with 61.47\% / 59.96\% for the 8-bit variant and
55.73\% / 54.36\% for FP32 Adam. The 4-bit variant is $0.63$~pp below 8-bit in
peak accuracy and $0.64$~pp above it in final accuracy, both smaller than the
cross-seed standard deviation of the 8-bit method ($\pm0.82$~pp,
Table~\ref{tab:multiseed}). It exceeds FP32 Adam by $+5.11$~pp (peak) and
$+6.24$~pp (final). An independent same-seed rerun of FP32 Adam under the 4-bit
codebase gives 55.30\% peak / 54.90\% final, a $+5.54$~pp peak gain,
confirming that the gap does not depend on a single baseline run.
Figure~\ref{fig:4bit_conv} shows test accuracy over rounds. In the first 20
rounds the 4-bit variant tracks FP32 Adam closely (51.20\% vs.\ 51.72\% at round
20). After round 30, FP32 Adam plateaus at roughly 54--55\%, while 4-bit
Q-LocalAdam continues to improve (55.21\% at round 40, 57.05\% at round 60,
60.07\% at round 70) and settles at 59--61\%. This late-stage separation under
strong skew matches the 8-bit method, so reducing the precision of the stored
state does not change the convergence profile.

\paragraph{Robustness across heterogeneity levels.}
Table~\ref{tab:4bit_main} shows that the 4-bit variant stays within $0.90$~pp of
the 8-bit variant at every heterogeneity level and above FP32 Adam: 62.20\% vs.\
60.95\% at $\alpha{=}0.5$ and 63.65\% vs.\ 62.69\% at $\alpha{=}1.0$ (peak).
Final-round accuracies are 60.60\%, 62.20\% and 62.84\% for $\alpha{=}0.1$, 0.5
and 1.0, so accuracy increases monotonically as the partition becomes more
balanced, as for the other methods. The gain over FP32 grows with heterogeneity
($+0.96$, $+1.25$ and $+5.11$~pp for $\alpha{=}1.0$, 0.5 and 0.1), the same
trend as at 8 bits. At $\alpha{=}1.0$ the 4-bit variant slightly exceeds the
8-bit variant ($+0.24$~pp), which we regard as noise.

\begin{table}[t]
\centering\small
\caption{4-bit Q-LocalAdam across heterogeneity levels (CIFAR-100, ResNet-18, seed 42). Peak / final accuracy (\%).}
\label{tab:4bit_main}
\begin{tabular}{lccc}
\toprule
$\alpha$ & FP32 Adam & 8-bit Q-LocalAdam & 4-bit Q-LocalAdam \\
\midrule
1.0 & 62.69 / 62.69 & 63.41 / 63.41 & 63.65 / 62.84 \\
0.5 & 60.95 / 60.95 & 63.10 / 63.10 & 62.20 / 62.20 \\
0.1 & 55.73 / 54.36 & 61.47 / 59.96 & 60.84 / 60.60 \\
\bottomrule
\end{tabular}
\end{table}

\paragraph{Why linear quantization fails at 4 bits.}
Naive INT4 stays at 1.00\% accuracy in every one of the 120 rounds
(Table~\ref{tab:4bit_abl}). This is the 4-bit analogue of the Naive INT8
failure, but more severe: with 16 quantized values per block, every
second-moment entry smaller than about $r/30$ (for block range $r$) is mapped to
the block minimum, so $\hat v\approx 0$ and the Adam step
$\hat m/(\sqrt{\hat v}+\epsilon)$ becomes extremely large.
Log-space coding avoids this because every code, including the lowest, decodes
to a strictly positive value no smaller than the floor $\tau$, and the
reconstruction error is multiplicative rather than absolute.

\begin{table}[t]
\centering
\caption{Component ablation, CIFAR-100 ($\alpha{=}0.1$). Accuracy is peak / final (\%); ``--'' means not reported.}
\label{tab:4bit_abl}
\resizebox{\columnwidth}{!}{%
\begin{tabular}{@{}lcccc@{}}
\toprule
 & \multicolumn{2}{c}{Mem.\ (MB, reduction)} & \multicolumn{2}{c}{Acc.\ (\%, peak / final)} \\
\cmidrule(lr){2-3}\cmidrule(l){4-5}
Variant & 8-bit & 4-bit & 8-bit & 4-bit \\
\midrule
Naive (linear $m$, $v$) & 26.67 (3.37$\times$) & 14.03 (6.40$\times$) & 1.00 / 1.00 & 1.00 / 1.00 \\
Momentum-only (FP32 $v$) & 58.24 (1.54$\times$) & 51.93 (1.73$\times$) & 52.49 / 52.49 & 17.28 / 1.00 \\
Variance-only (FP32 $m$) & 58.24 (1.54$\times$) & 51.93 (1.73$\times$) & 61.10 / 61.10 & 60.70 / 60.24 \\
Full Q-LocalAdam & 26.67 (3.37$\times$) & 14.03 (6.40$\times$) & 61.47 / 59.96 & 60.84 / 60.60 \\
\bottomrule
\end{tabular}%
}
\end{table}

\paragraph{Component ablation: variance is robust, momentum is the bottleneck.}
Table~\ref{tab:4bit_abl} isolates each component. Variance-only quantization
(FP32 momentum, log-space 4-bit variance) reaches 60.70\% peak / 60.24\% final,
within $0.14$~pp of the full 4-bit method and only $0.40$~pp below its 8-bit
counterpart (61.10\%), so the log-space variance quantizer is essentially
insensitive to the reduction from 8 to 4 bits. In contrast, momentum-only
quantization (linear 4-bit momentum, FP32 variance) is unstable: it reaches
17.28\% by round 4 and then collapses to near-chance accuracy (final 1.00\%),
whereas its 8-bit counterpart trains to 52.49\%. Quantizing momentum is
therefore not neutral at 4 bits when variance is stored exactly. We hypothesize
that a linear 4-bit momentum quantizer, which has resolution $r/15$ and cannot
represent zero exactly, adds noise of order $r/30$ to every element; an
accurate and small FP32 denominator amplifies this noise into large normalized
steps, whereas the coarse, log-quantized variance yields a more conservative
effective step size that absorbs it. We have not isolated this mechanism
experimentally.

\paragraph{Comparison with zero-free linear variance quantization.}
~\cite{li2023memory} address the zero-point problem with a linear
quantizer whose codebook excludes zero (and, in their full recipe, rank-1
normalization and small blocks for the first moment). We therefore compare
three variance quantizers in our federated protocol, keeping our linear
momentum quantizer fixed (Table~\ref{tab:vquant}). Linear min/max quantization
of $v$ (Naive INT4) collapses to chance (1.00\%). Replacing it with a
zero-excluded block-wise linear quantizer removes the collapse (51.59\% last-10
mean), which confirms that quantization of small variances to zero causes the
failure. Log-space quantization, however, is a further $8.7$~pp better
(60.27\%) and exceeds FP32 Adam, whereas the zero-free linear variant remains
$2.2$~pp below it. Avoiding the zero point is thus necessary but not sufficient
in this setting. We also implemented the full recipe of Li et al.\ (block size
128 with a dynamic-exponent map for $m$, rank-1 normalization with a
zero-excluded linear map for $v$, and tensors with fewer than 4096 elements kept
in FP32) inside our framework. It trains stably and uses the least memory
(11.68~MB, $7.69\times$) but reaches 52.94\%, $0.85$~pp below FP32 Adam and
$7.3$~pp below 4-bit Q-LocalAdam. The zero-free quantizer stores one scalar per
block rather than two, which accounts for its slightly smaller footprint
($13.33$ vs.\ $14.03$~MB) and does not change the conclusion.

\begin{table}[t]
\centering\small
\caption{Variance quantizer ablation, CIFAR-100, $\alpha{=}0.1$, 4-bit states, linear momentum in all rows. Last-10 mean accuracy (\%); mean $\pm$ std over seeds ($n$) where $n{>}1$.}
\label{tab:vquant}
\begin{tabular}{lccc}
\toprule
Variance quantizer & $n$ & Mem (MB) & Acc.\ (\%) \\
\midrule
Linear min/max (Naive INT4)          & 1 & 14.03 & 1.00 \\
Zero-free linear (block-wise)        & 1 & 13.33 & 51.59 \\
Log-space (ours)                     & 3 & 14.03 & \textbf{60.27 $\pm$ 0.52} \\
\midrule
FP32 Adam (reference)                & 3 & 89.82 & 53.79 $\pm$ 0.96 \\
Li et al.-style recipe (re-impl.)    & 1 & 11.68 & 52.94 \\
\bottomrule
\end{tabular}
\end{table}

\paragraph{Block size and learning rate.}
Table~\ref{tab:4bit_block} shows that increasing $B$ from 32 to 64 and 128
reduces memory from 16.84 to 14.03 and 12.63~MB ($5.33\times$, $6.40\times$ and
$7.11\times$), because the fixed metadata cost is spread over more codes (the
effective storage falls from 6 to 5 and 4.5 bits per state element). Block size
has little effect on final accuracy (60.60\%, 60.60\% and 60.77\% for $B{=}32$,
64 and 128). Metadata is 33\% of storage at $B{=}32$, 20\% at $B{=}64$, and
about 11\% for the 8-bit variant at $B{=}64$. $B{=}128$ attains the best peak
accuracy (61.60\%) and the lowest memory, but the difference is within our
single-seed resolution; larger blocks are preferable at 4 bits within the range
tested, though they may provide coarser quantization resolution. A lower
learning rate improves the 4-bit method: $\eta{=}5{\times}10^{-4}$ gives 61.89\%
peak / 61.72\% final versus 60.84\% / 60.60\% at $\eta{=}10^{-3}$, indicating
that the default learning rate is not optimal for 4-bit states.

\begin{table}[t]
\centering\small
\caption{Block size and learning rate for 4-bit Q-LocalAdam (CIFAR-100, $\alpha{=}0.1$).}
\label{tab:4bit_block}
\begin{tabular}{lcccc}
\toprule
Setting & Eff.\ bits & Mem (MB) & Reduction & Peak / final Acc (\%) \\
\midrule
$B{=}32$, $\eta{=}10^{-3}$ & 6.0 & 16.84 & 5.33$\times$ & 61.05 / 60.60 \\
$B{=}64$, $\eta{=}10^{-3}$ & 5.0 & 14.03 & 6.40$\times$ & 60.84 / 60.60 \\
$B{=}128$, $\eta{=}10^{-3}$ & 4.5 & 12.63 & 7.11$\times$ & 61.60 / 60.77 \\
$B{=}64$, $\eta{=}5{\times}10^{-4}$ & 5.0 & 14.03 & 6.40$\times$ & 61.89 / 61.72 \\
\bottomrule
\end{tabular}
\end{table}

\paragraph{Limitations of the 4-bit evaluation.}
The heterogeneity sweep, ablations and block-size study use a single seed (42);
only the headline comparison is repeated over three seeds. All experiments use
one dataset (CIFAR-100) and one architecture (ResNet-18), and differences below
$\sim$1~pp (block sizes, $\alpha{=}0.5$ vs.\ $1.0$, 4-bit vs.\ 8-bit) should not
be considered significant. We did not evaluate the 4-bit variant on ViT-Tiny,
AG News or the Jetson Orin Nano. FP32 and 8-bit numbers in
Table~\ref{tab:4bit_main} are taken from separate runs reported earlier in this
paper. Reported memory is persistent optimizer state and excludes the transient
full-precision buffers used while updating one tensor. Because optimizer states
are reinitialized every round, quantization error accumulates over only
$\sim$300 local steps; behavior under persistent state is not evaluated for 4
bits.
%\end{comment}

\end{document}